\documentclass[10pt,twocolumn,letterpaper]{article}

\usepackage{cvpr}
\usepackage{times}
\usepackage{epsfig}
\usepackage{graphicx}
\usepackage{amsmath}
\usepackage{amssymb}
\usepackage{enumerate}
\usepackage{multirow}
\usepackage{threeparttable}
\usepackage[normalem]{ulem}
\usepackage{appendix}

\newcommand{\vB}{\mathbf{B}}

\newcommand{\vL}{\mathbf{L}}

\newcommand{\vE}{\mathbf{E}}
\newcommand{\vM}{\mathbf{M}}

\newcommand{\calT}{\mathcal{T}}
\newcommand{\calE}{\mathcal{S}}

\usepackage[pagebackref=true,breaklinks=true,letterpaper=true,colorlinks,bookmarks=false]{hyperref}

\cvprfinalcopy 

\def\cvprPaperID{2605} 

\ifcvprfinal\fi
\begin{document}

\title{Bringing a Blurry Frame Alive at High Frame-Rate with an Event Camera}

\author{Liyuan Pan $^{1,2}$, Cedric Scheerlinck$^{1,2}$, Xin Yu$^{1,2}$, Richard Hartley$^{1,2}$, Miaomiao Liu$^{1,2}$, and Yuchao Dai$^{3}$\\ 
$^{1}$ Research School of Engineering, Australian National University, Canberra, Australia \\ 
$^{2}$ Australia Centre for Robotic Vision \\
$^{3}$ School of Electronics and Information, Northwestern Polytechnical University, Xi'an, China \\
\tt\small{liyuan.pan}@anu.edu.au
}
\maketitle


\begin{abstract}
Event-based cameras can measure intensity changes (called `{\it events}') with microsecond accuracy under high-speed motion and challenging lighting conditions. With the active pixel sensor (APS), the event camera allows simultaneous output of the intensity frames. However, the output images are captured at a relatively low frame-rate and often suffer from motion blur.
A blurry image can be regarded as the integral of a sequence of latent images, while the events indicate the changes between the latent images. Therefore, we are able to model the blur-generation process by associating event data to a latent image. 
In this paper, we propose a simple and effective approach, the \textbf{Event-based Double Integral (EDI)} model, to reconstruct a high frame-rate, sharp video from a single blurry frame and its event data.
The video generation is based on solving a simple non-convex optimization problem in a single scalar variable. Experimental results on both synthetic and real images demonstrate the superiority of our EDI model and optimization method in comparison to the state-of-the-art.


\end{abstract}


\section{Introduction}

\begin{figure*}[ht]
\begin{center}
\resizebox{\textwidth}{!}{
\begin{tabular}{cccc}
\hspace{-0.10cm}
\includegraphics[width=0.215\textwidth]{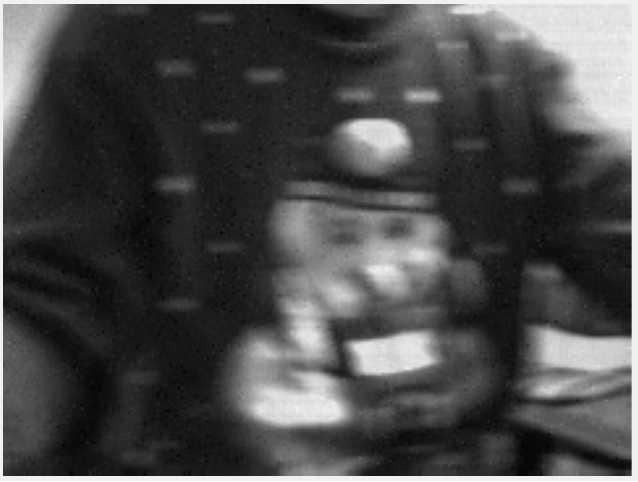}
&\includegraphics[width=0.215\textwidth]{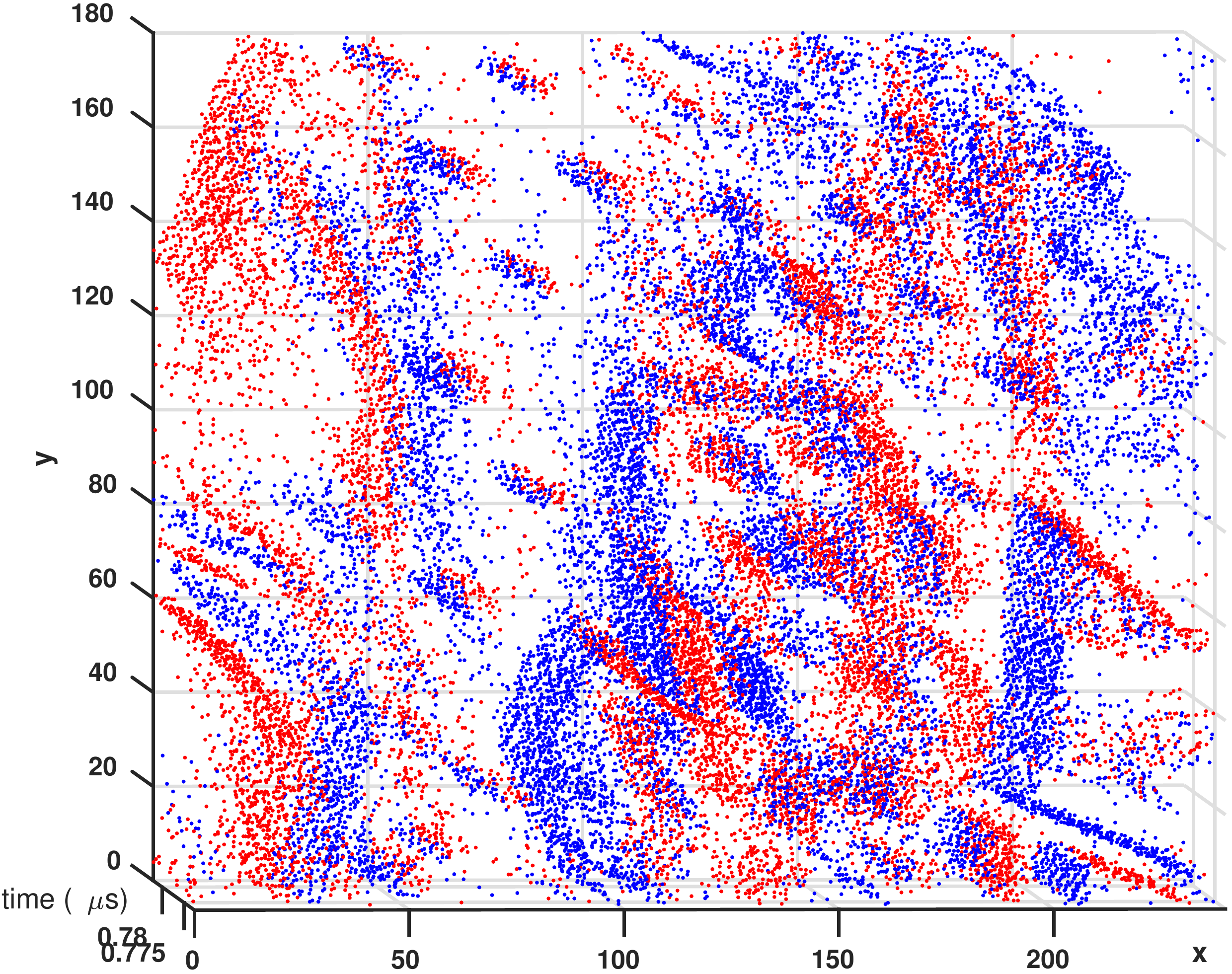}
&\includegraphics[width=0.215\textwidth]{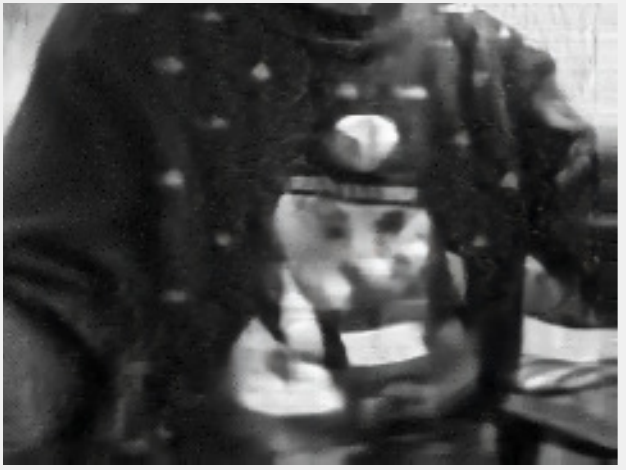}
&\includegraphics[width=0.215\textwidth]{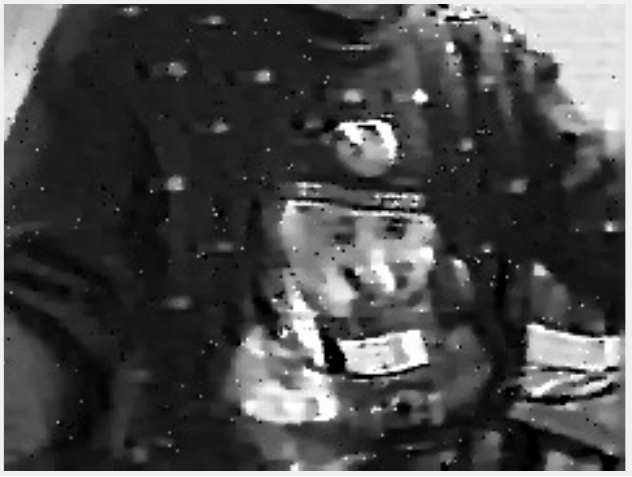}\\
\hspace{-0.20cm}
(a) The Blurry Image  
&(b) The Events
&(c) Tao \etal \cite{Tao_2018_CVPR}  
&(d) Pan \etal \cite{pan2017deblurring}\\
\hspace{-0.10cm}
\includegraphics[width=0.215\textwidth]{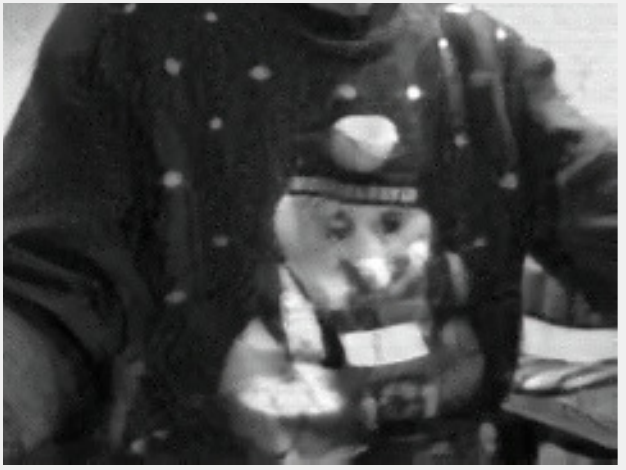}
&\includegraphics[width=0.215\textwidth]{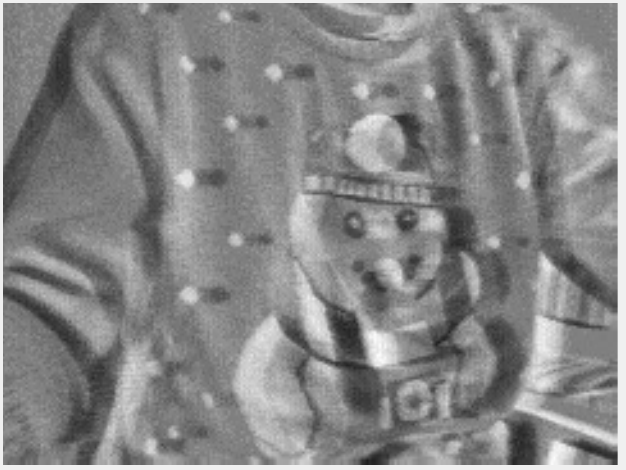}
&\includegraphics[width=0.215\textwidth]{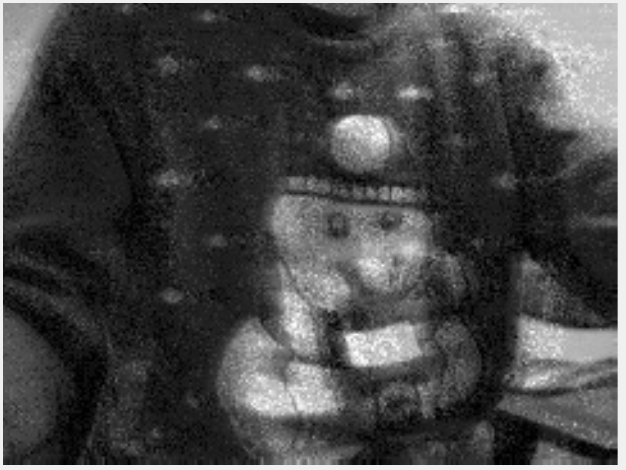}
&\includegraphics[width=0.215\textwidth]{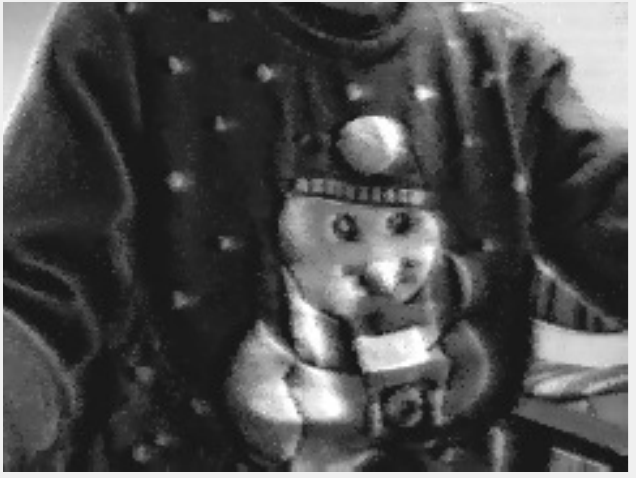}\\
\hspace{-0.10cm}
(e) Jin \etal \cite{Jin_2018_CVPR}
&(f) {\small\begin{tabular}[c]{@{}c@{}}Scheerlinck \etal \cite{Scheerlinck18arxiv}\\ (events only)\end{tabular}}
&(g) Scheerlinck \etal \cite{Scheerlinck18arxiv} 
&(h) Ours\\
\end{tabular}
}
\end{center}
\vspace{-3mm}
 \caption{\em \label{fig:eventdeblur}
Deblurring and reconstruction results of our method compared with the state-of-the-art methods on our real {\it blurry event dataset}. 
(a) The input blurry image. 
(b) The corresponding event data. 
(c) Deblurring result of Tao \etal \cite{Tao_2018_CVPR}. 
(d) Deblurring result of Pan \etal \cite{pan2017deblurring}.
(e) Deblurring result of Jin \etal \cite{Jin_2018_CVPR}. Jin uses video as training data to train a supervised model to perform deblur, where the video can also be considered as similar information as the event data. 
(f)-(g) Reconstruction results of Scheerlinck \etal \cite{Scheerlinck18arxiv}, (f) from only events, (g) from combining events and frames. 
(h) Our reconstruction result. (Best viewed on screen).
}
\vspace{-2mm}
\end{figure*}

Event cameras (such as the Dynamic Vision Sensor (DVS) \cite{lichtsteiner2008128} and the Dynamic and Active-pixel Vision Sensor (DAVIS) \cite{brandli2014240}) are sensors that asynchronously measure the intensity changes at each pixel independently with microsecond temporal resolution\footnote{If nothing moves in the scene, no events are triggered.}.
The event stream encodes the motion information by measuring the precise pixel-by-pixel intensity changes. Event cameras are more robust to low lighting and highly dynamic scenes than traditional cameras since they are not affected by under/over exposure or motion blur associated with a synchronous shutter. 

Due to the inherent differences between event cameras and standard cameras, existing computer vision algorithms designed for standard cameras cannot be applied to event cameras directly. Although the DAVIS \cite{brandli2014240} can provide the simultaneous output of the intensity frames and the event stream, there still exist major limitations with current event cameras:
\begin{itemize}
    \item {\bf Low frame-rate intensity images:} In contrast to the high temporal resolution of event data ($\geq 3\mu s$ latency), the current event cameras only output low frame-rate intensity images ($\geq 5ms$ latency). 
    \item {\bf Inherent blurry effects:} When recording highly dynamic scenes, motion blur is a common issue due to the relative motion between the camera and the scene. The output of the intensity image from the APS tends to be blurry.  
\end{itemize}

To address these above challenges, various methods have been proposed by reconstructing high frame-rate videos. The existing methods can be in general categorized as \textbf{1}) Event data only solutions \cite{Bardow16cvpr, Reinbacher16bmvc, Barua16wcav}, where the results tend to lack the texture and consistency of natural videos, as they fail to use the complementary information contained in the low frame-rate intensity image;
\textbf{2}) Low frame-rate intensity-image-only solutions \cite{Jin_2018_CVPR}, where an end-to-end learning framework has been proposed to learn regression between a single blurry image and a video sequence, whereas the rich event data are not used; and \textbf{3}) Jointly exploiting event data and intensity images \cite{Scheerlinck18arxiv, Shedligeri18arxiv, Brandli14iscas}, building upon the interaction between both sources of information. However, these methods fail to address the blur issue associated with the captured image frame. Therefore, the reconstructed high frame-rate videos can be degraded by blur.


Although blurry frames cause undesired image degradation, they also encode the relative motion between the camera and the observed scene. Taking full advantage of the encoded motion information would benefit the reconstruction of high frame-rate videos.

To this end, we propose an {\textbf{Event-based Double Integral (EDI)}} model to resolve the above problems by reconstructing a high frame-rate video from a single image (even blur) and its event sequence, where the blur effects have been reduced in each reconstructed frame. Our EDI model naturally relates the desired high frame-rate sharp video, the captured intensity frame and event data. Based on the EDI model, high frame-rate video generation is as simple as solving a non-convex optimization problem in a single scalar variable.

Our main contributions are summarized as follows.
\begin{enumerate}[1)] 
\item We propose a simple and effective model, named the Event-based Double Integral (EDI) model, to restore a high frame-rate sharp video from a single image (even blur) and its corresponding event data.
\vspace{-2mm}
\item Using our proposed formulation of EDI, we propose a stable and general method to generate a sharp video under various types of blur by solving a single variable non-convex optimization problem, especially in low lighting and complex dynamic conditions.
\vspace{-2mm}
\item The frame rate of our reconstructed video can theoretically be as high as the event rate (200 times greater than the original frame rate in our experiments).
\end{enumerate}

 
\section{Related Work}

Event cameras such as the DAVIS and DVS \cite{brandli2014240,lichtsteiner2008128} report log intensity changes, inspired by human vision. Although several works try to explore
the advantages of the high temporal resolution provided by event cameras \cite{zhou2018semi, Kim16eccv, Rebecq17ral, Zhu17cvpr,Zhu-RSS-18,Gehrig18eccv,Kueng16iros}, how to make the best use of the event camera has not yet been fully investigated. 

{\noindent{\bf Event-based image reconstruction.}} 
Kim \etal \cite{Kim14bmvc} reconstruct high-quality images from an event camera under a strong assumption that the only movement is pure camera rotation, and later extend their work to handle 6-degree-of-freedom motion and depth estimation \cite{Kim16eccv}.
Bardow \etal \cite{Bardow16cvpr} aim to simultaneously recover optical flow and intensity images.
Reinbacher \etal \cite{Reinbacher16bmvc} restore intensity images via manifold regularization.
Barua \etal \cite{Barua16wcav} generate image gradients by dictionary learning and obtain a logarithmic intensity image via Poisson reconstruction.
However, the intensity images reconstructed by the previous approaches suffer from obvious artifacts as well as lack of texture 
due to the spatial sparsity of event data.

To achieve more image detail in the reconstructed images, several methods trying to combine events with intensity images have been proposed.
The DAVIS \cite{brandli2014240} uses a shared photo-sensor array to simultaneously output events (DVS) and intensity images (APS).
Scheerlinck \etal \cite{Scheerlinck18arxiv} propose an asynchronous event-driven complementary filter to combine APS intensity images with events, and obtain continuous-time image intensities.
Brandli \etal \cite{Brandli14iscas} directly integrate events from a starting APS image frame, and each new frame resets the integration.
Shedligeri \etal \cite{Shedligeri18arxiv} first exploit two intensity images to estimate depth. Then, they use the event data only to reconstruct a pseudo-intensity sequence (using \cite{Reinbacher16bmvc}) between the two intensity images and use the pseudo-intensity sequence to estimate ego-motion using visual odometry. Using the estimated 6-DOF pose and depth, they directly warp the intensity image to the intermediate location.
Liu \etal \cite{Liu17vc} assume a scene should have static background. Thus, their method needs an extra sharp static foreground image as input and the event data are used to align the foreground with the background.

\vspace{1mm}

{\noindent{\bf Image deblurring.}}
Traditional deblurring methods usually make assumptions on the scenes (such as a static scene) or exploit multiple images (such as stereo, or video) to solve the deblurring problem. 
Significant progress has been made in the field of single image deblurring. Methods using gradient based regularizers, such as Gaussian scale mixture \cite{fergus2006removing}, $l_1 \backslash l_2$ norm \cite{krishnan2011blind}, edge-based patch priors \cite{sun2013edge} and ${l}_0$-norm regularizer \cite{xu2013unnatural}, have been proposed.  
Non-gradient-based priors such as the color line based prior \cite{lai2015blur}, and the extreme channel (dark/bright channel) prior \cite{pan2017deblurring, yan2017image} have also been explored. Another family of image deblurring methods tends to use multiple images \cite{cho2012video, hyun2015generalized, sellent2016stereo, Pan_2017_CVPR, pan2018depth}. 

Driven by the success of deep neural networks, Sun \etal \cite{sun2015learning} propose a convolutional neural network (CNN) to estimate locally linear blur kernels. Gong \etal \cite{gong2017motion} learn optical flow from a single blurry image through a fully-convolutional deep neural network. The blur kernel is then obtained from the estimated optical flow to restore the sharp image. 
Nah \etal~\cite{Nah_2017_CVPR} propose a multi-scale CNN that restores latent images in an end-to-end learning manner without assuming any restricted blur kernel model. Tao \etal \cite{Tao_2018_CVPR} propose a light and compact network, SRN-DeblurNet, to deblur the image.
However, deep deblurring methods generally need a large dataset to train the model and usually require sharp images provided as supervision. In practice, blurry images do not always have corresponding ground-truth sharp images.

\vspace{1mm}
{\noindent{\bf Blurry image to sharp video.}}
Recently, two deep learning based methods \cite{Jin_2018_CVPR,purohit2018bringing} propose to restore a video from a single blurry image with a fixed sequence length.
However, their reconstructed videos do not obey the 3D geometry of the scene and camera motion.
Although deep-learning based methods achieve impressive  performance in various scenarios, their success heavily depend on the consistency between the training datasets and the testing datasets, thus hinder the generalization ability for real-world applications. 


\section{Formulation}
In this section, we develop an {\bf EDI} model of the relationships between the events, the latent image and the blurry image. 
Our goal is to reconstruct a high frame-rate, sharp video from a single image and its corresponding events.
This model can tackle various blur types and work stably in highly dynamic contexts and low lighting conditions. 

\subsection{Event Camera Model}

Event cameras are bio-inspired sensors that asynchronously report logarithmic intensity changes \cite{brandli2014240,lichtsteiner2008128}. Unlike conventional cameras that produce the full image at a fixed frame-rate, event cameras trigger events whenever the change in intensity at a given pixel exceeds a preset threshold. Event cameras do not suffer from the limited dynamic ranges typical of sensors with synchronous exposure time, and are able to capture high-speed motion with microsecond accuracy.


Inherent in the theory of event cameras is the concept of the latent
image $L_{xy}(t)$, denoting the instantaneous intensity at pixel 
$(x, y)$ at time $t$, related to the rate of photon arrival at that pixel.
The latent image $L_{xy}(t)$ is not directly output by the camera.
Instead, the camera outputs a sequence of {\em events},
denoted by $(x,y,t,\sigma)$, which record
changes in the intensity of the latent image.
Here, $(x, y)$ are image coordinates, $t$ is the time the event takes place, and polarity $\sigma = \pm 1 $ denotes the direction (increase or decrease) of the intensity change at
that pixel and time. Polarity is given by,
{\small
\begin{align} \label{eq:log}
\sigma = \calT\left( \log \Big( \frac{\vL_{xy}(t)}{\vL_{xy}(t_\text{ref})} \Big) , c \right),
\end{align}
}
where $\calT(\cdot,\cdot)$ is a truncation function,
{\small
\begin{equation} \nonumber \label{eq:pkmodel}
\calT(d,c) =  
\begin{cases}
+1, & d \geq c,\\
0, & d\in(-c,c),\\
-1, & d \leq -c.
\end{cases}
\end{equation}
}
Here, $c$ is a threshold parameter determining whether an event should be recorded or not, $\vL_{xy}(t)$ is the intensity at pixel $(x,y)$ in the image $\vL(t)$
and $t_\text{ref}$ denotes the timestamp of the previous event. When an event is triggered, $\vL_{xy}(t_\text{ref})$ at that pixel is updated to a new intensity level.

\begin{figure}[t]
\begin{center}
\begin{tabular}{cccc}
\includegraphics[width=0.215\textwidth]{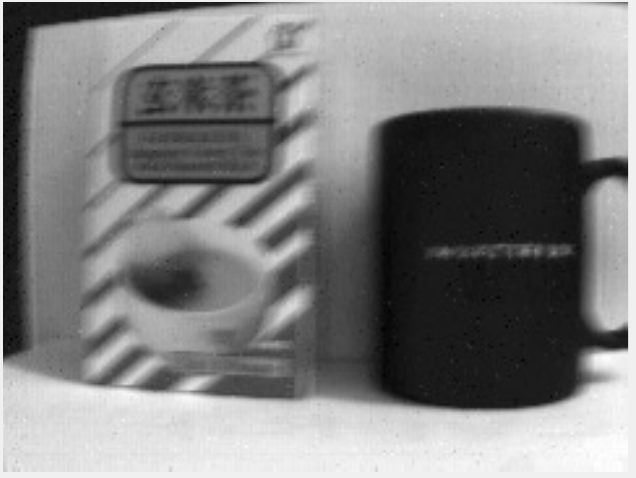}
&\includegraphics[width=0.215\textwidth]{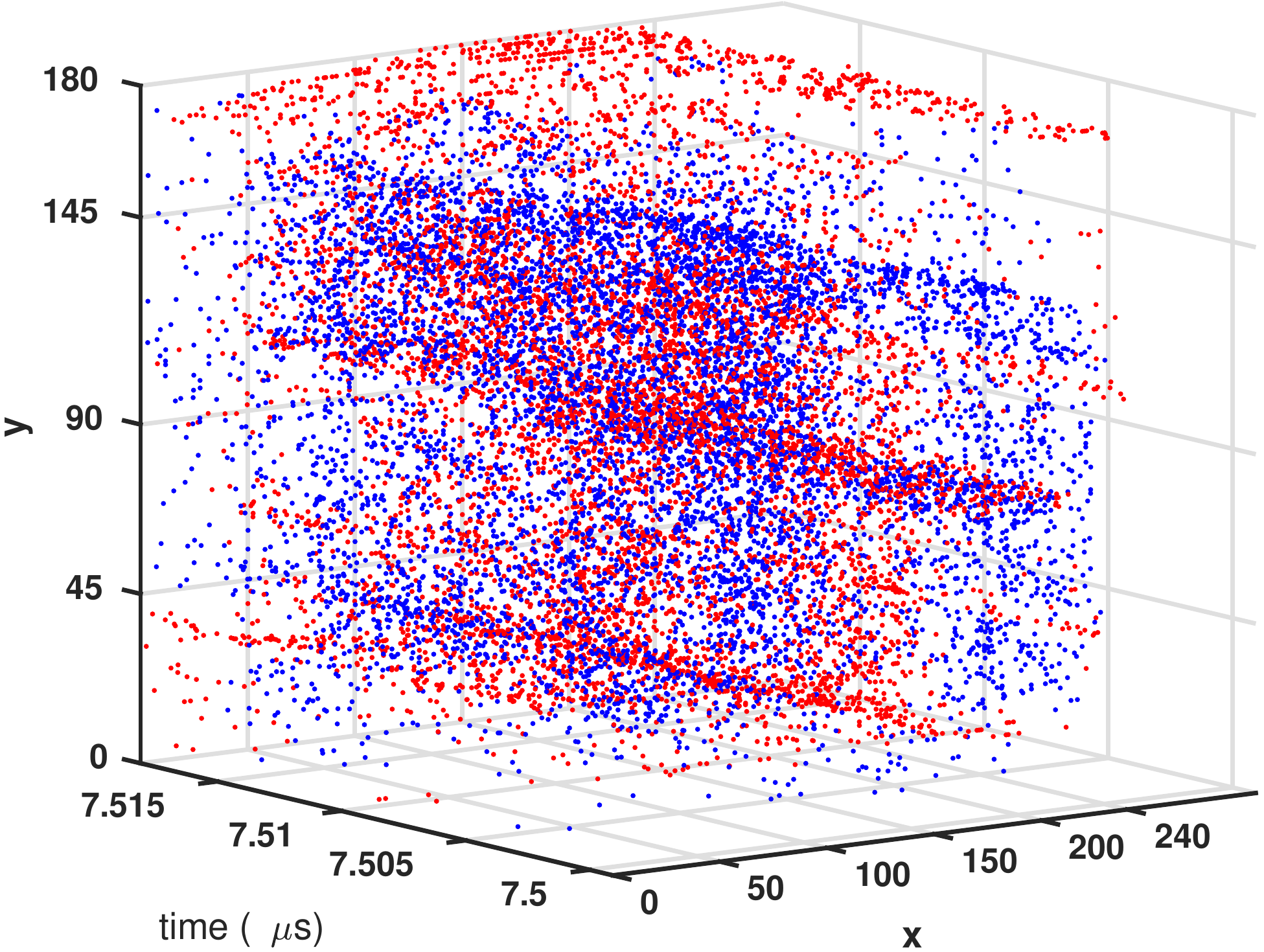}\\
(a) The Blurry Image  
&(b) The Events\\
\includegraphics[width=0.215\textwidth]{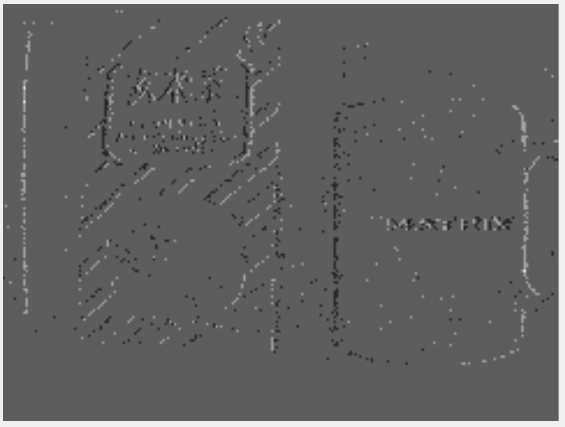}
&\includegraphics[width=0.215\textwidth]{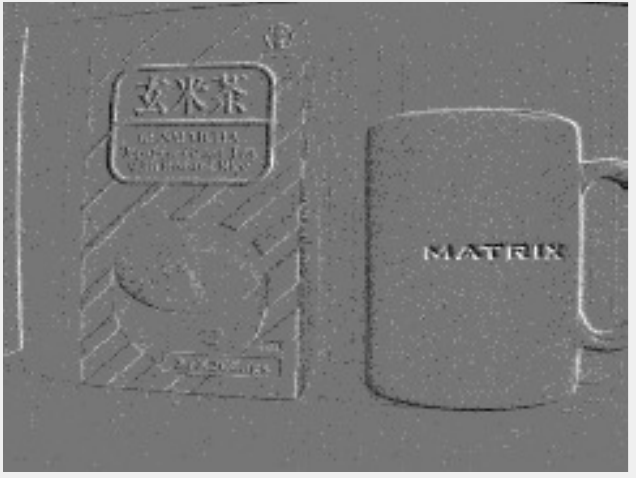}\\
(c) $E(t) = \int e(t) \, dt$  
&(d) $\frac{1}{T}\int \exp(c\,\vE(t)) dt$\\
\includegraphics[width=0.215\textwidth]{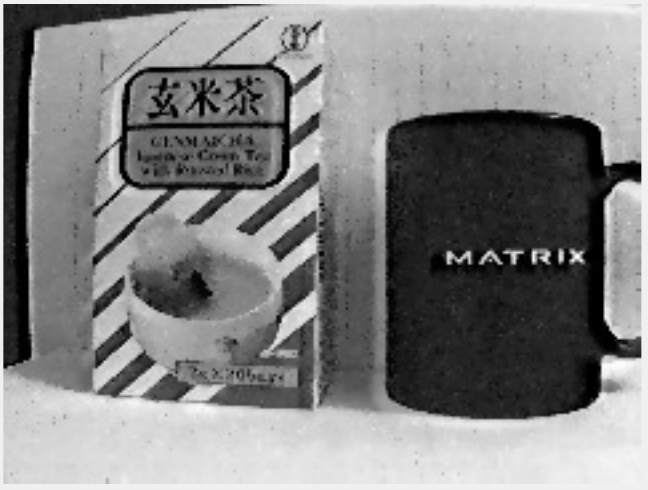}
&\includegraphics[width=0.215\textwidth]{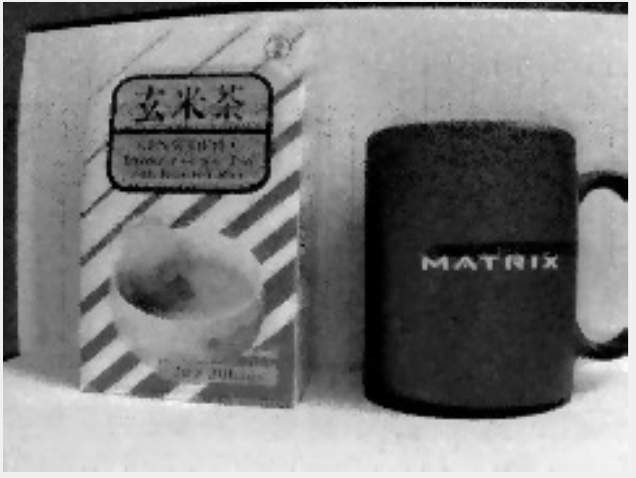}\\
\multicolumn{2}{c}{(e) Sample Frames of Our Reconstructed Video}\\
\end{tabular}
\end{center}
\vspace{-3mm}
\caption{\label{fig:eventsample} \em The event data and our reconstructed result, where (a) and (b) are the input of our method. (a) The intensity image from the event camera.
(b) Events from the event camera plotted in 3D space-time $(x, y, t)$ (blue: positive event; red: negative event).
(c) The first integral of several events during a small time interval. 
(d) The second integral of events during the exposure time. 
(e)Samples from our reconstructed video from $\vL(0)$ to $\vL(200)$. 
}
\end{figure}


\subsection{Intensity Image Formation}

In addition to the event sequence, event cameras can provide a full-frame
grey-scale intensity image, at a much slower rate than the event sequence.
The grey-scale images may suffer from motion blur due to their long exposure time. A general model  of image formation is given by,
%
{\small
\begin{equation} \label{eq:avgmodel}
\vB =  \frac{1}{T}\int_{f-T/2}^{f+T/2} \vL(t) \, dt,
\end{equation}
}
where $\vB$ is a blurry image, equal to the average value of
the latent image during the exposure time $[f-T/2, f+T/2]$.
This equation applies to each pixel $(x, y)$ independently, and
subscripts $x,y$ denoting pixel location are often omitted henceforth.


\subsection{Event-based Double Integral Model}
We aim to recover a sequence of latent intensity images by exploiting both the blur model and the event model. 
%
We define $e_{xy}(t)$ as a function of continuous time $t$ such that
\vspace{-2mm}
\begin{equation} \nonumber \label{eq:et}
e_{xy}(t) = \sigma \,\, \delta_{t_0}(t).
\vspace{-1mm}
\end{equation}
whenever there is an event $(x, y, t_0, \sigma)$.
Here, $\delta_{t_0}(t)$ is an impulse function, with unit integral, at time $t_0$, and the sequence of events is turned into a continuous time signal, consisting of a sequence of impulses.
There is such a function $e_{xy}(t)$ for every point $(x,y)$ in the image.
Since each pixel can be treated separately, we omit the subscripts $x, y$.

During an exposure period $[f-T/2, f+T/2]$, we define $\vE(t)$ as the sum of events between time $f$ and $t$ at a given pixel,
\begin{equation} \nonumber \label{eq:Etwithet} 
\begin{split}
\vE(t) &= \int_{f}^t e(s) ds ,\\
\end{split}
\end{equation}
which represents the proportional change in intensity between time $f$ and $t$.
%
Except under extreme conditions, such as glare and no-light conditions,
the latent image sequence $\vL(t)$ is expressed as,
\begin{equation}\label{eq:Ltsigma}
\begin{split}
\vL(t)\ & = \vL(f) \, \exp( c\,  \vE(t)) = \vL(f) \, \exp(c)^{\vE(t)} ~.
\end{split}
\end{equation}
%
In particular, an event $(x,y,t,\sigma)$ is triggered when the intensity of a pixel $(x,y)$ increases or decreases by an amount $c$ at time $t$.
We put a tilde on top of things to denote logarithm, \eg $\widetilde{\vL}(t)=\log({\vL}(t))$. 
%
\begin{equation}\label{eq:logLtwithEt}
\begin{split}
\widetilde{\vL}(t)\ & = \widetilde{\vL}(f) + c \, \vE(t).\\
\end{split}
\end{equation}

Given a sharp frame, we can reconstruct a high frame-rate video from the sharp starting point $\vL(f)$ by using Eq.~\eqref{eq:logLtwithEt}. When the input image is blurry, a trivial solution would be to first deblur the image with an existing deblurring method and then to reconstruct the video using Eq.~\eqref{eq:logLtwithEt} (see Fig.\ref{fig:baseline} for details). 
However, in this way, the event data between intensity images is not fully exploited, thus resulting in inferior performance.
Instead, we propose to reconstruct the video by exploiting the inherent connection between event and blur, and present the following model.

As for the blurred image,
{\small
\begin{equation}\label{eq:blurevent}
\begin{split}
\vB &=  \frac{1}{T} \int_{f-T/2}^{f+T/2} \vL(t) dt\\
& = \frac{\vL(f)}{T}\int_{f-T/2}^{f+T/2}  \exp \Big (c \int_{f}^{t} e(s) ds\Big)\ dt  ~.
\end{split}
\end{equation}
}
In this manner, we construct the relation between the captured blurry image $\vB$ and the latent image $\vL(f)$ through the double integral of the event. We name Eq.\eqref{eq:blurevent} the \textbf{Event-based Double Integral (EDI)} model. 


Taking the logarithm on both sides of Eq.~\eqref{eq:blurevent} and rearranging, yields
{\small
\begin{equation}\label{eq:logEDIM}
\widetilde{\vL}(f) = \widetilde{\vB} - \log \left(\frac{1}{T}\int_{f-T/2}^{f+T/2} \exp(c \,\vE(t)) dt\right), 
\end{equation}
}
which shows a linear relation between the blurry image, the latent image and the integral of the events in the log space.


\subsection{High Frame-Rate Video Generation}
The right-hand side of Eq.~\eqref{eq:logEDIM} is known, apart from perhaps
the value of the contrast threshold $c$, the first term from
the grey-scale image, the second term from the event sequence, 
it is possible to compute $\widetilde\vL(f)$, and hence $\vL(f)$ by
exponentiation.  Subsequently, from 
Eq.~\eqref{eq:logLtwithEt} the latent image $\vL(t)$ at any time may be computed.

To avoid accumulated errors of constructing a video from many frames
of a blurred video, it is more suitable to construct each frame $\vL(t)$ using the closest blurred frame.



Theoretically, we could generate a video with frame-rate as high as the DVS's eps (events per second). However, as each event carries little information and is subject to noise, several events must be processed together to yield a reasonable image. We generate a reconstructed frame every 50-100 events, so for our experiments, the frame-rate of the reconstructed video is usually 200 times greater than the input low frame-rate video. 
Furthermore, as indicated by Eq.~\eqref{eq:logEDIM}, the challenging blind motion deblurring problem has been reduced to a single variable optimization problem of how to find the best value of the contrast threshold $c$. In the following section, we use $\vL(c,t)$ to present the latent sharp image $\vL(t)$ with different $c$.

\begin{figure}[t]
\begin{center}
\begin{tabular}{cc}
\includegraphics[width=0.215\textwidth]{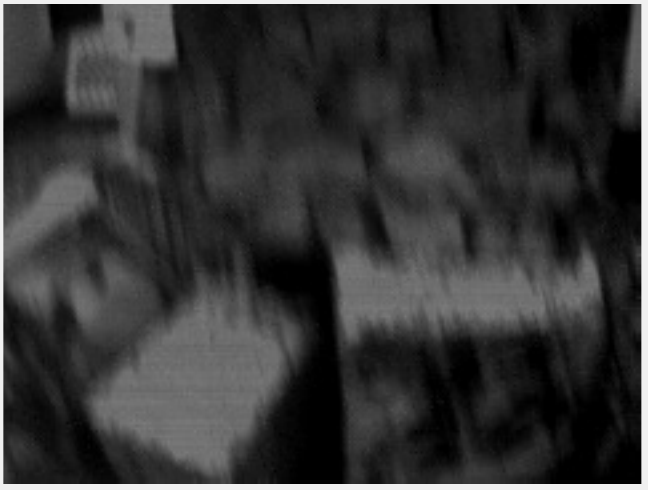}  
& \includegraphics[width=0.215\textwidth]{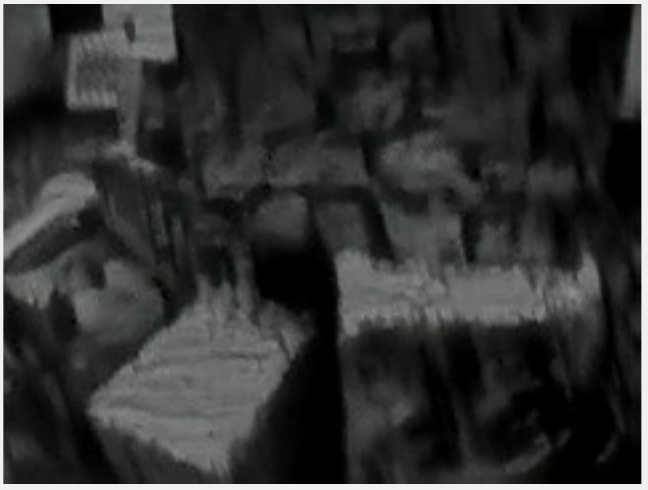}  \\
(a) The blurry image
&(b) Tao \etal \cite{Tao_2018_CVPR}\\
\includegraphics[width=0.215\textwidth]{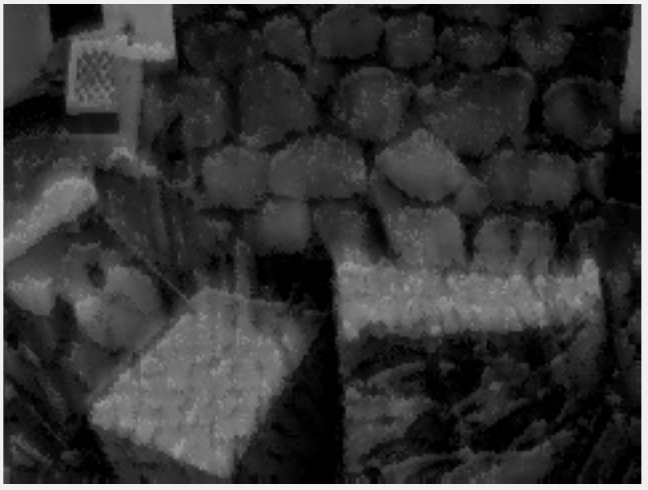}  
&\includegraphics[width=0.215\textwidth]{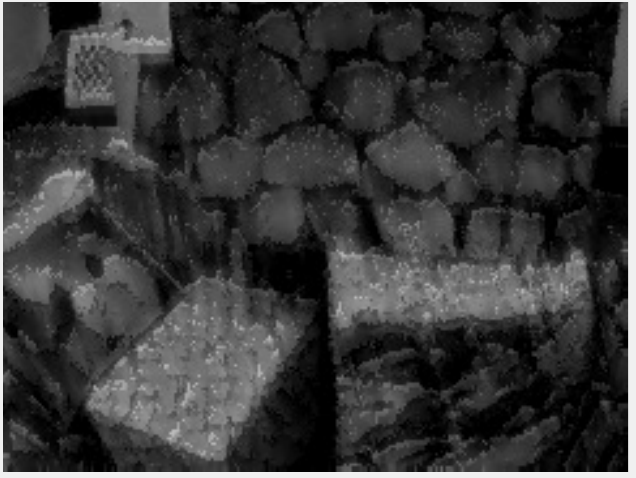} \\
(c) By human observation
&(d) By energy minimization\\
\end{tabular}
\end{center}
\vspace{-2mm}
\caption{\label{fig:real23}
An example of our reconstruction result using different methods to estimate $c$, from the real dataset \cite{mueggler2017event}. (a) The blurry image. (b) Deblurring result of \cite{Tao_2018_CVPR} (c) Our result where $c$ is chosen by manual inspection. (d) Our result where $c$ is computed automatically by our proposed energy minimization \eqref{eq:solveK}.}
\label{fig:manually}
\end{figure}
\vspace{-1 mm}
 \begin{figure}[t]
 \begin{center}
 \includegraphics[width=0.405\textwidth]{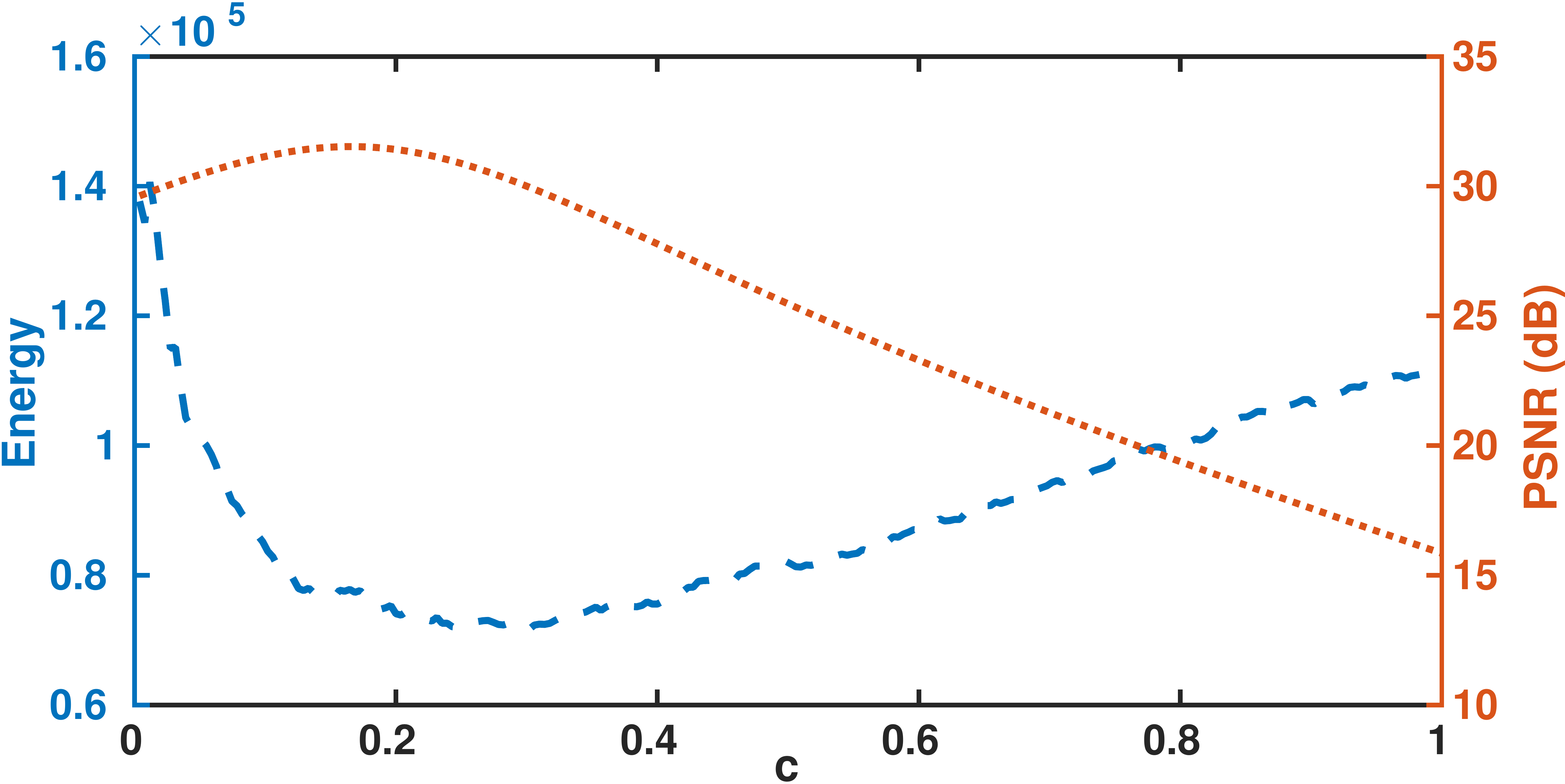}
 \end{center}
 \vspace{-3mm}
 \caption{\em \label{fig:deltaVsblur} The figure plot deblurring performance against the value of $c$. The image is clearer with higher PSNR value.}
 \vspace{-2mm}
 \end{figure}

\section{Optimization}

The unknown contrast threshold $c$ represents the minimum change in log intensity required to trigger an event.
By choosing an appropriate $c$ in Eq.~\eqref{eq:blurevent}, we can generate a sequence of sharper images.
To this end, we first need to evaluate the sharpness of the reconstructed images. Here, we propose two different methods to estimate the unknown variable $c$: manually chosen and automatically optimized.

\begin{figure*}[ht]
\begin{center}
\resizebox{\textwidth}{!}{
\begin{tabular}{cccc}
\includegraphics[width=0.213\textwidth]{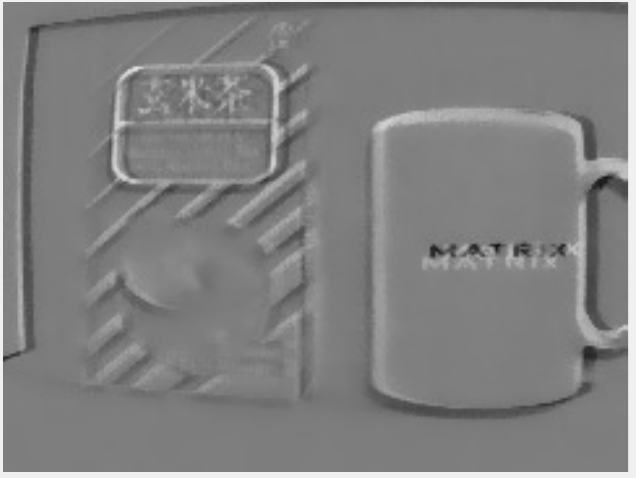}
&\includegraphics[width=0.213\textwidth]{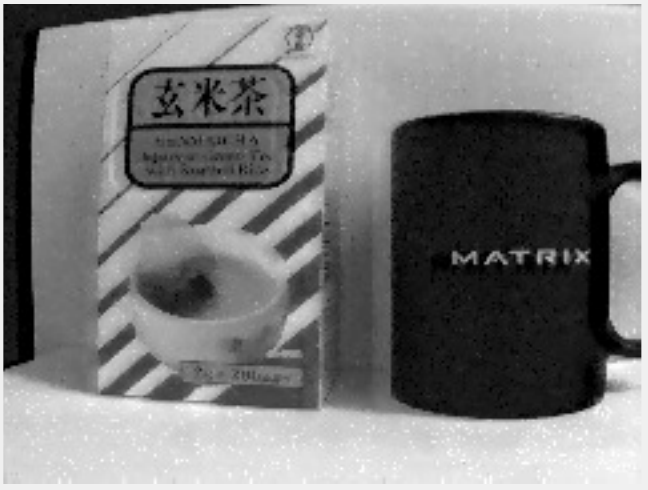}
&\includegraphics[width=0.213\textwidth]{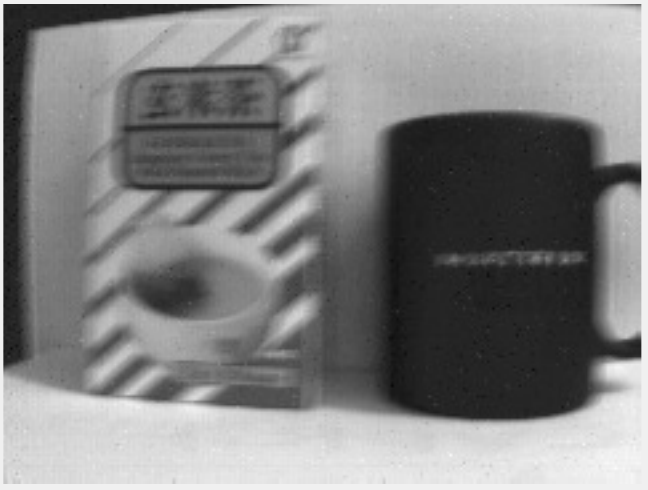}
&\includegraphics[width=0.213\textwidth]{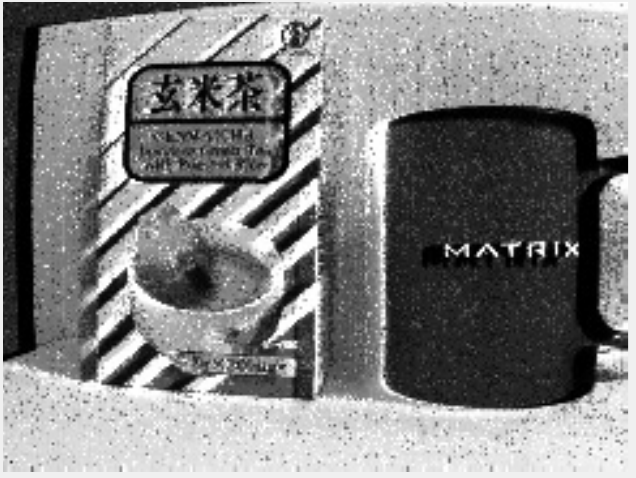}\\[0.1in]
\includegraphics[width=0.213\textwidth]{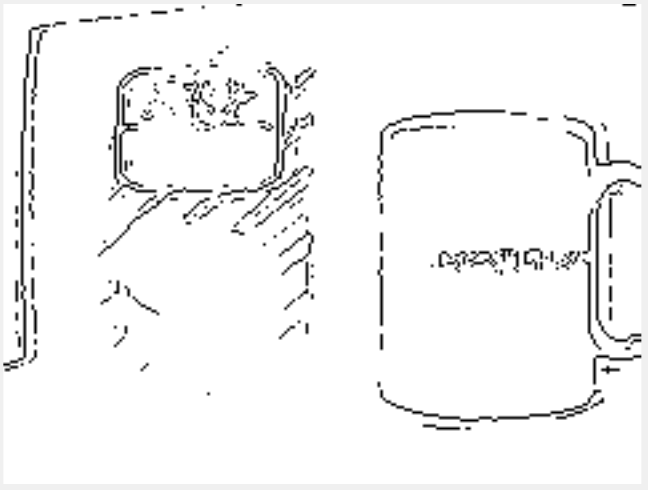}
&\includegraphics[width=0.213\textwidth]{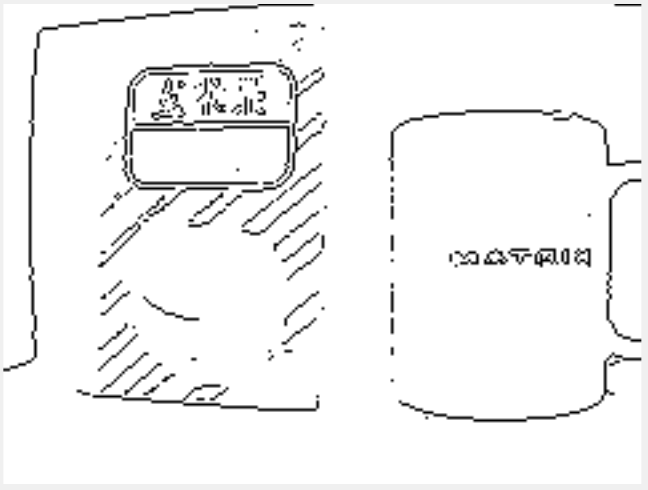}
&\includegraphics[width=0.213\textwidth]{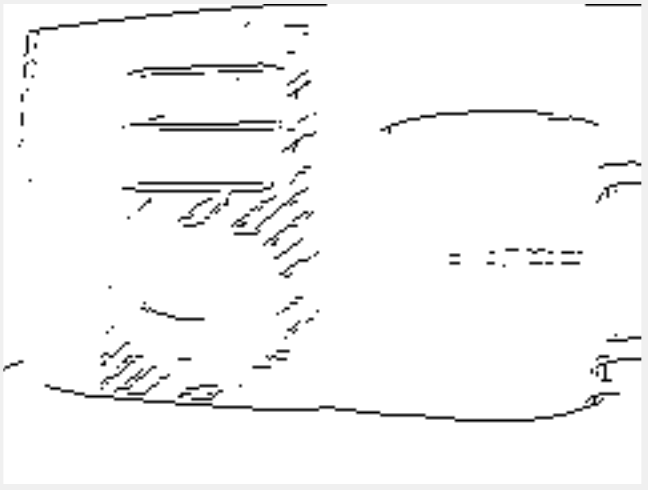}
&\includegraphics[width=0.213\textwidth]{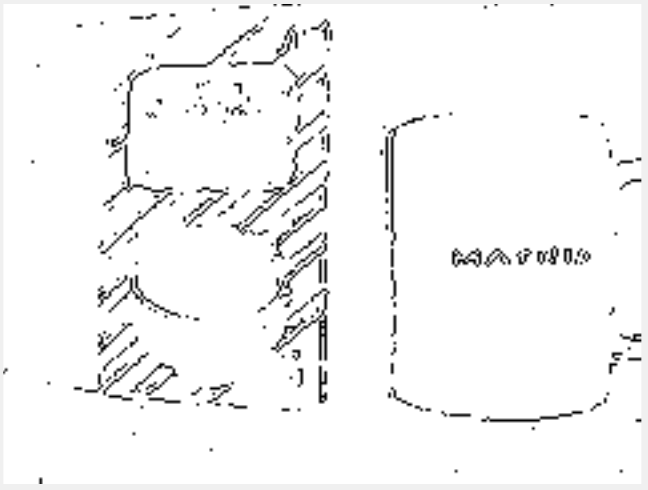}
\\
\end{tabular}
}
\end{center}
\vspace{-2.5 mm}
\caption{\label{fig:csample} \em At left, the edge image $M(f)$
and below, its Sobel edge map.  To the right are $3$ reconstructed
latent images using different values of $c$,
low 0.03, middle 0.11 and high 0.55.  Above, the reconstructed images, below, their Sobel
edge maps.  The optimal value of the threshold $c$ is found by computing the cross-correlation of such images with the edge map at the left. (Best viewed on screen).
}
\end{figure*}

\begin{figure*}[ht]
\begin{center}
\resizebox{\textwidth}{!}{
\begin{tabular}{cccc}
\includegraphics[width=0.213\textwidth]{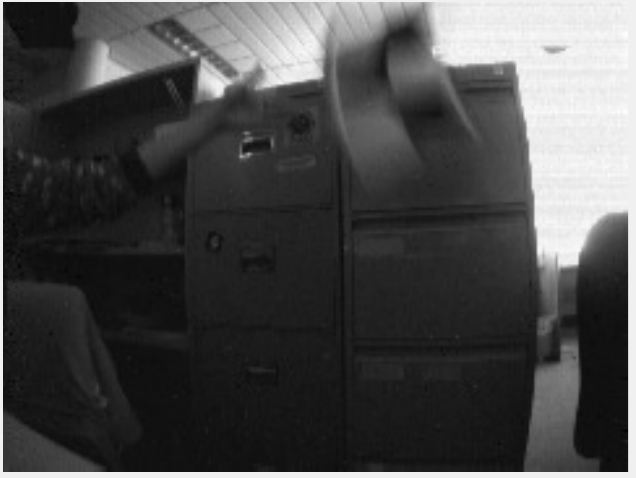}
&\includegraphics[width=0.213\textwidth]{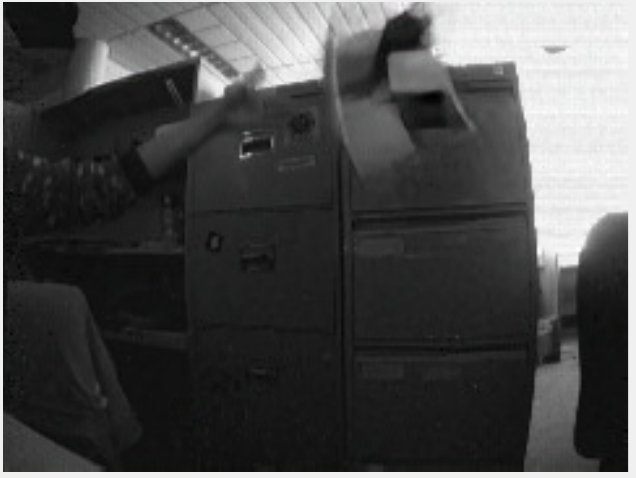}
&\includegraphics[width=0.213\textwidth]{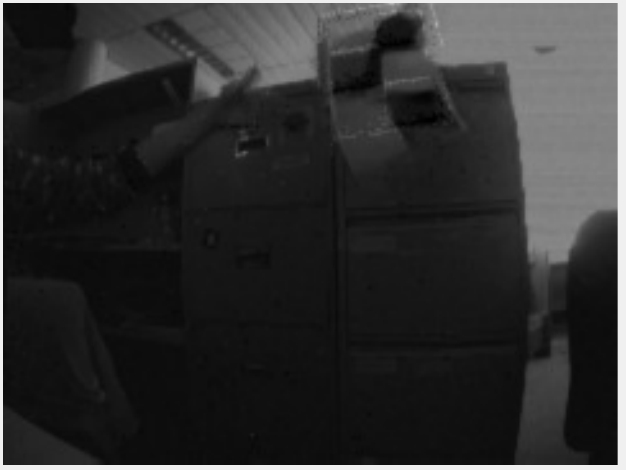}
&\includegraphics[width=0.213\textwidth]{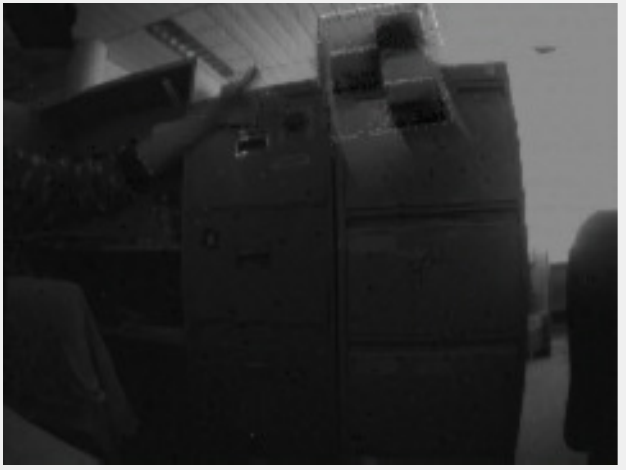}\\
(a) The Blur Image 
&(b) Jin \etal \cite{Jin_2018_CVPR}  
&(c) Baseline 1
&(d) Baseline 2\\
\includegraphics[width=0.213\textwidth]{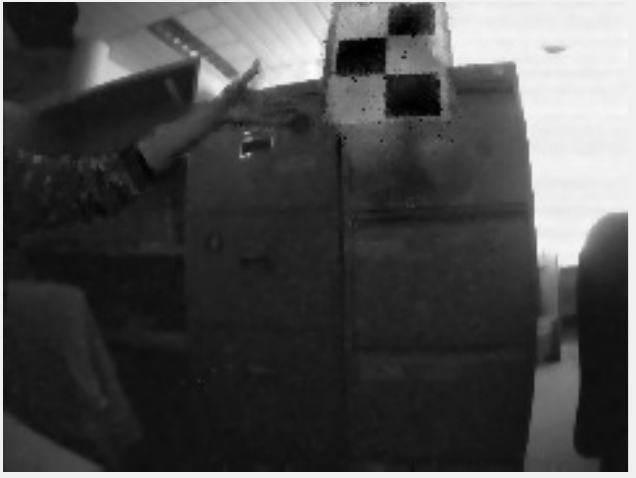}
&\includegraphics[width=0.213\textwidth]{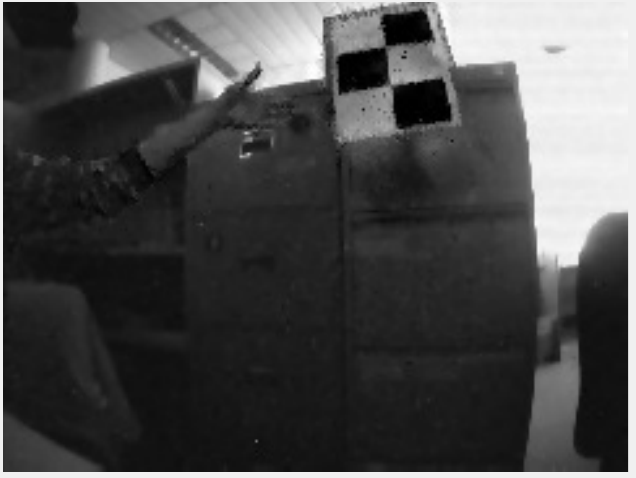}
&\includegraphics[width=0.213\textwidth]{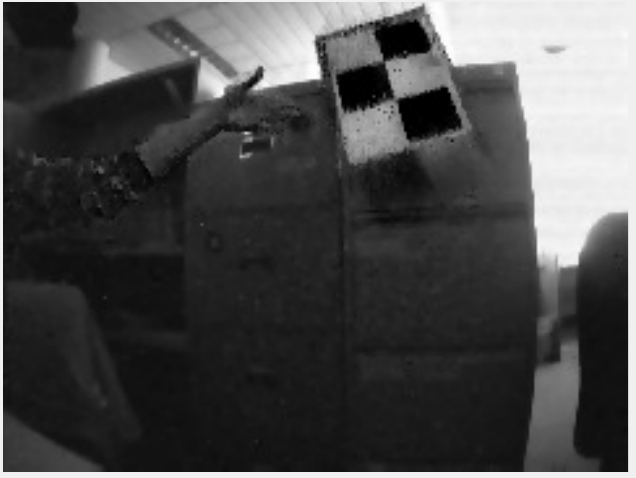}
&\includegraphics[width=0.213\textwidth]{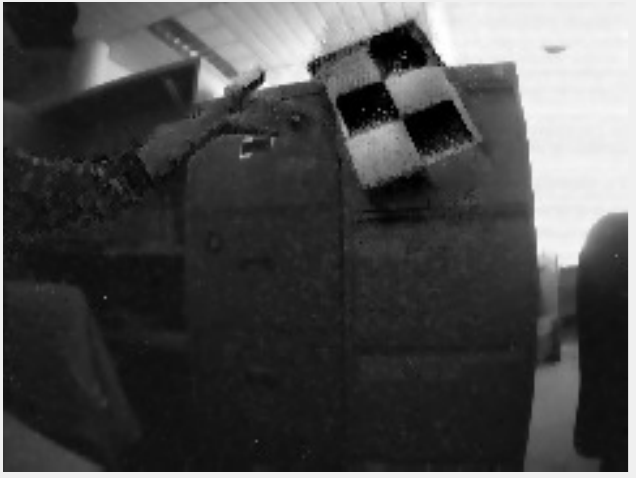}\\
\multicolumn{4}{c}{(f) Samples of Our Reconstructed Video}
\end{tabular}
}
\end{center}
\vspace{-2.5 mm}
 \caption{\em \label{fig:baseline} Deblurring and reconstruction results on our real {\it blurry event dataset}. 
(a) Input blurry images. 
(b) Deblurring result of \cite{Jin_2018_CVPR}. 
(c) Baseline 1 for our method. We first use the state-of-the-art video-based deblurring method \cite{Jin_2018_CVPR} to recover a sharp image. Then use the sharp image as input to a state-of-the-art reconstruction method \cite{Scheerlinck18arxiv} to get the intensity image.
(d) Baseline 2 for our method. We first use method \cite{Scheerlinck18arxiv} to reconstruct an intensity image. Then use a deblurring method \cite{Jin_2018_CVPR} to recover a sharp image.
(e) Samples from our reconstructed video from $\vL(0)$ to $\vL(150)$.
(Best viewed on screen).
}
\end{figure*}

\subsection{Manually Chosen $c$}
According to our {\bf EDI} model in Eq.~\eqref{eq:blurevent}, given a value for $c$, we can obtain a sharp image.
Therefore, we develop a method for deblurring by manually inspecting the visual effect of the deblurred image. 
In this way, we incorporate human perception into the reconstruction loop and the deblurred images should satisfy human observation.
In Fig.~\ref{fig:manually}, we give an example for manually chosen and automatically optimized results on dataset from \cite{mueggler2017event}.

\subsection{Automatically Chosen $c$}

To automatically find the best $c$, we need to build an evaluation 
metric (energy function) that can evaluate the quality of the
deblurred image $\vL(c,t)$. Specifically, we propose to exploit 
different prior knowledge for sharp images and the event data. 
\vspace{-1 mm}
\subsubsection{Edge Constraint for Event Data}

As mentioned before, when a proper $c$ is given, our reconstructed image $\vL(c,t)$ will contain much sharper edges compared with the original input intensity image. Furthermore, event cameras inherently yield responses at moving intensity boundaries, 
so edges in the latent image may be located where (and when) events occur.
This allows us to find latent image edges.
An edge at time $t$ corresponds to an event (at the pixel in question)
during some time interval around $t$ so we convolve the event sequence
with an exponentially decaying window, to obtain an edge map,
%
\begin{equation} \nonumber \label{Mt}
\begin{split}
\vM(t) &= \int_{0}^{T} \exp(-( \alpha |t-s|))  \, e(t) \, ds,\\
\end{split}
\end{equation}
%
where $ \alpha$ is a weight parameter for time attenuation, which we
set to $1$. 
As $\vM(t)$ mainly reports the intensity changes but not the intensity value itself, we need to change $\vM(t)$ to an edge map $\calE(\vM(t))$ using the
Sobel filter, which is also applied to $\vL(c,t)$. (See Fig. \ref{fig:csample} for details). 

Here, we use cross-correlation between $\calE(\vL(c,t))$ and $\calE(\vM(t))$ to evaluate the sharpness of $\vL(c,t)$.
%
{\small
\begin{equation}
\phi_{\rm{edge}}(\vL(c,t)) = \sum_{x, y} 
\calE(\vL(c,t))(x, y) \cdot \calE(\vM(t))(x, y) ~.
\end{equation}
}
\vspace{-4 mm}
\subsubsection{Regularizing the Intensity Image}
In our model, total variation is used to suppress noise in the latent image while preserving edges, and penalize the spatial fluctuations\cite{rudin1992nonlinear}. Therefore, we use conventional total variation (TV) based regularization
\begin{equation} \label{TVnorm}
\phi_{\rm{TV}}(\vL(c,t)) = |\nabla \vL(c,t)|_{1},
\end{equation}
%
where $\nabla$ represents the gradient operators.
\vspace{-1 mm}
\subsubsection{Energy Minimization}

The optimal $c$ can be estimate by solving Eq.~\eqref{eq:solveK},
\begin{equation}\label{eq:solveK}
\min_{c} \phi_{\rm{TV}}(\vL(c,t)) + \lambda \phi_{\rm{edge}}(\vL(c,t)) ,
\end{equation}
where $\lambda$ is a trade-off parameter. The response of cross-correlation reflect the matching rate of $\vL(c,t)$ and $\vM(t)$ which makes $\lambda<0$.  
This single-variable minimization problem can be solved by the nonlinear least-squares method \cite{more1978levenberg}, Scatter-search\cite{ugray2007scatter} or Fibonacci search \cite{dunlap1997golden}. 

In Fig.~\ref{fig:deltaVsblur}, we illustrate the clearness of the reconstructed image against the value of $c$. Meanwhile, we also provide the PSNR of the corresponding reconstructed image. As demonstrated in the figure, our proposed reconstruction metric could locate/identify the best deblurred image with peak PSNR properly. 

{\small
\begin{table*}\footnotesize
\begin{center}
\caption{\em  {\em Quantitative comparisons on the Synthetic dataset \cite{Nah_2017_CVPR}.} This dataset provides videos can be used to generate not only blurry images but also event data. All methods are tested under the same blurry condition, where methods \cite{Nah_2017_CVPR,Jin_2018_CVPR,Tao_2018_CVPR,Zhang_2018_CVPR} use GoPro dataset \cite{Nah_2017_CVPR} to train their models. Jin \cite{Jin_2018_CVPR} achieves their best performance when the image is down-sampled to 45\% mentioned in their paper.}
\label{all_all}
\begin{tabular}{c|c|c|c|c|c|c|c|c}
\hline
\multicolumn{9}{c}{Average result of the deblurred images on dataset\cite{Nah_2017_CVPR}}  \\ \hline
            & Pan \etal \cite{pan2017deblurring} & Sun \etal \cite{sun2015learning}  & Gong \etal \cite{gong2017motion}  & Jin \etal \cite{Jin_2018_CVPR} & Tao \etal \cite{Tao_2018_CVPR} & Zhang \etal \cite{Zhang_2018_CVPR} & Nah \etal \cite{Nah_2017_CVPR}  & Ours    \\ \hline
PSNR(dB)  & 23.50  & 25.30  & 26.05   & 26.98    & {\bf30.26}     & 29.18   & 29.08    & 29.06   \\ \hline
SSIM      & 0.8336 & 0.8511 & 0.8632  & 0.8922   & 0.9342    & 0.9306  & 0.9135   & {\bf{0.9430}}    \\ \hline
\multicolumn{9}{c}{Average result of the reconstructed videos on dataset\cite{Nah_2017_CVPR}}  \\ \hline
\multicolumn{1}{c|}{}     & \multicolumn{2}{c|}{Baseline 1 \cite{Tao_2018_CVPR} + \cite{Scheerlinck18arxiv}} & \multicolumn{2}{c|}{Baseline 2 \cite{Scheerlinck18arxiv} + \cite{Tao_2018_CVPR}} & \multicolumn{2}{c|}{Scheerlinck \etal \cite{Scheerlinck18arxiv}} & \multicolumn{1}{c|}{Jin \etal \cite{Jin_2018_CVPR}}    & Ours   \\ \hline
\multicolumn{1}{c|}{PSNR(dB)} & \multicolumn{2}{c|}{25.52}          & \multicolumn{2}{c|}{26.34}          & \multicolumn{2}{c|}{25.84}           & \multicolumn{1}{c|}{25.62}  & {\bf{28.49 }}  \\ \hline
\multicolumn{1}{c|}{SSIM} & \multicolumn{2}{c|}{0.7685}          & \multicolumn{2}{c|}{0.8090}          & \multicolumn{2}{c|}{0.7904}           & \multicolumn{1}{c|}{0.8556} & {\bf{0.9199 }} \\ \hline
\end{tabular}
\end{center}
\vspace{-3mm}
\end{table*}
}

\begin{figure*}[ht]
\begin{center}
\begin{tabular}{cccccc}
\multicolumn{2}{c}{
\hspace{-0.3 cm}
\begin{tabular}{cc}
\includegraphics[height=0.130\textwidth]{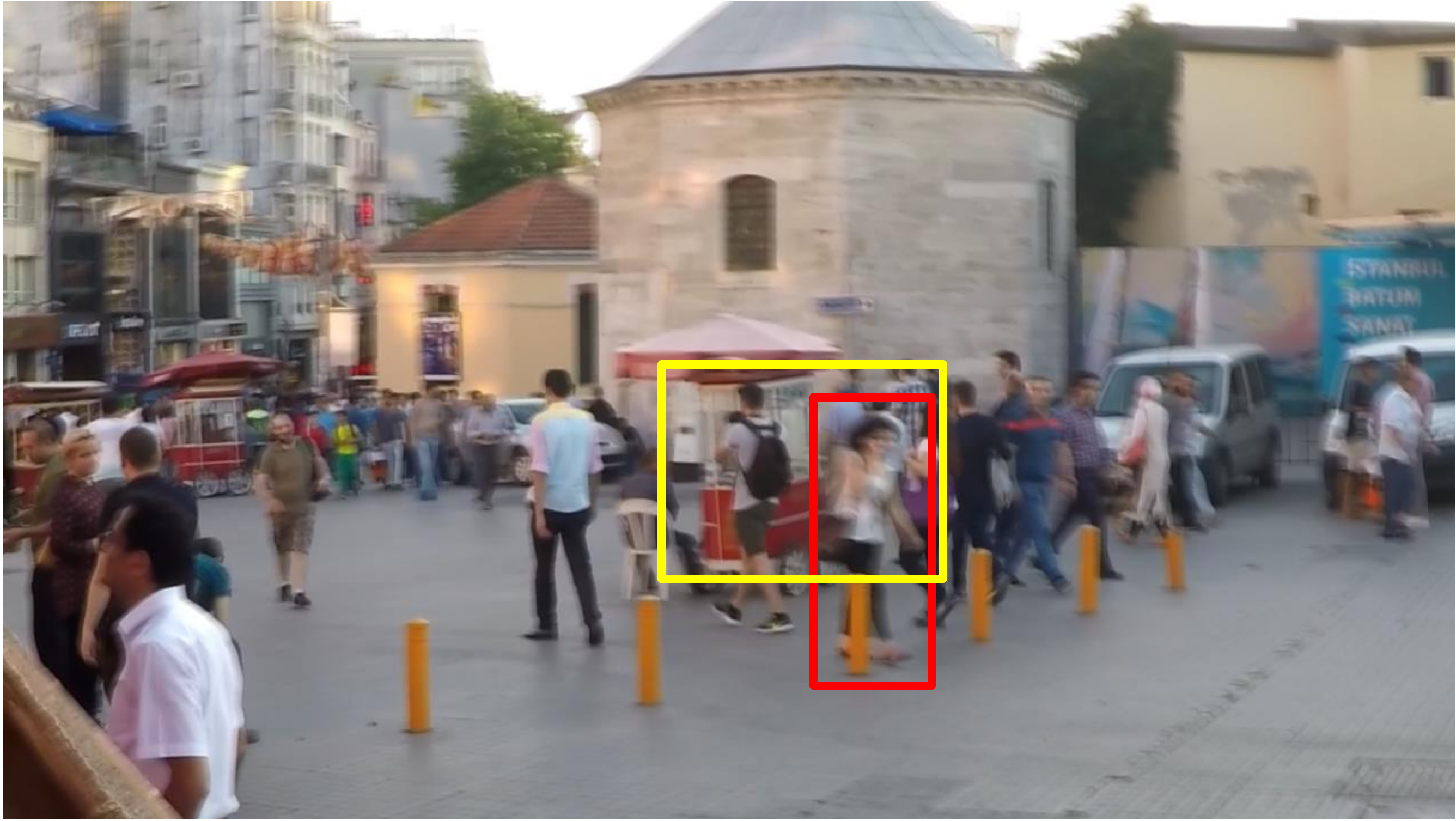}
\hspace{-0.4 cm}
&\includegraphics[height=0.130\textwidth]{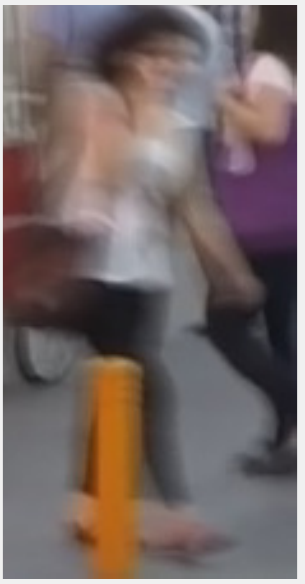}\\
\multicolumn{2}{c}{(a) The Blurry Image}
\end{tabular}
}
&\multicolumn{2}{c}{
\hspace{-0.6 cm}
\begin{tabular}{cc}
\includegraphics[height=0.130\textwidth]{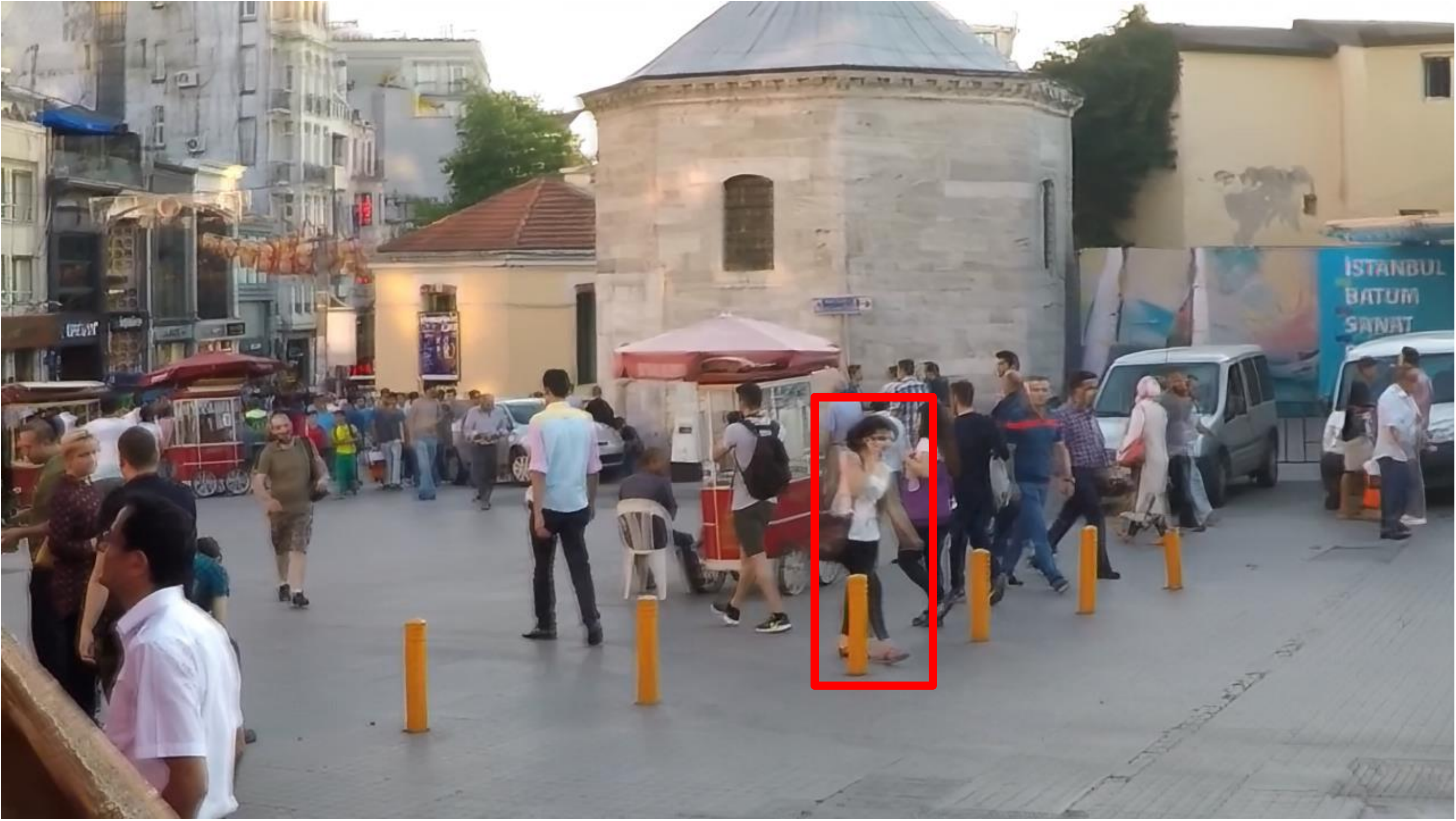}
\hspace{-0.4 cm}
&\includegraphics[height=0.130\textwidth]{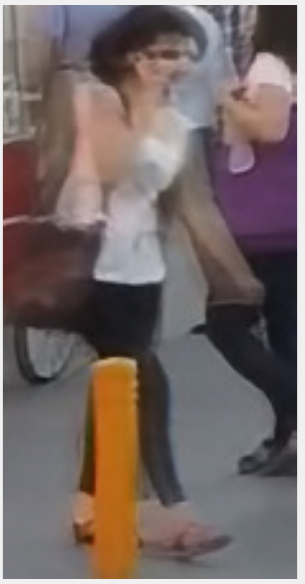}\\
\multicolumn{2}{c}{(b) Jin \etal \cite{Jin_2018_CVPR}}
\end{tabular}
}
&\multicolumn{2}{c}{
\hspace{-0.6 cm}
\begin{tabular}{cc}
\includegraphics[height=0.130\textwidth]{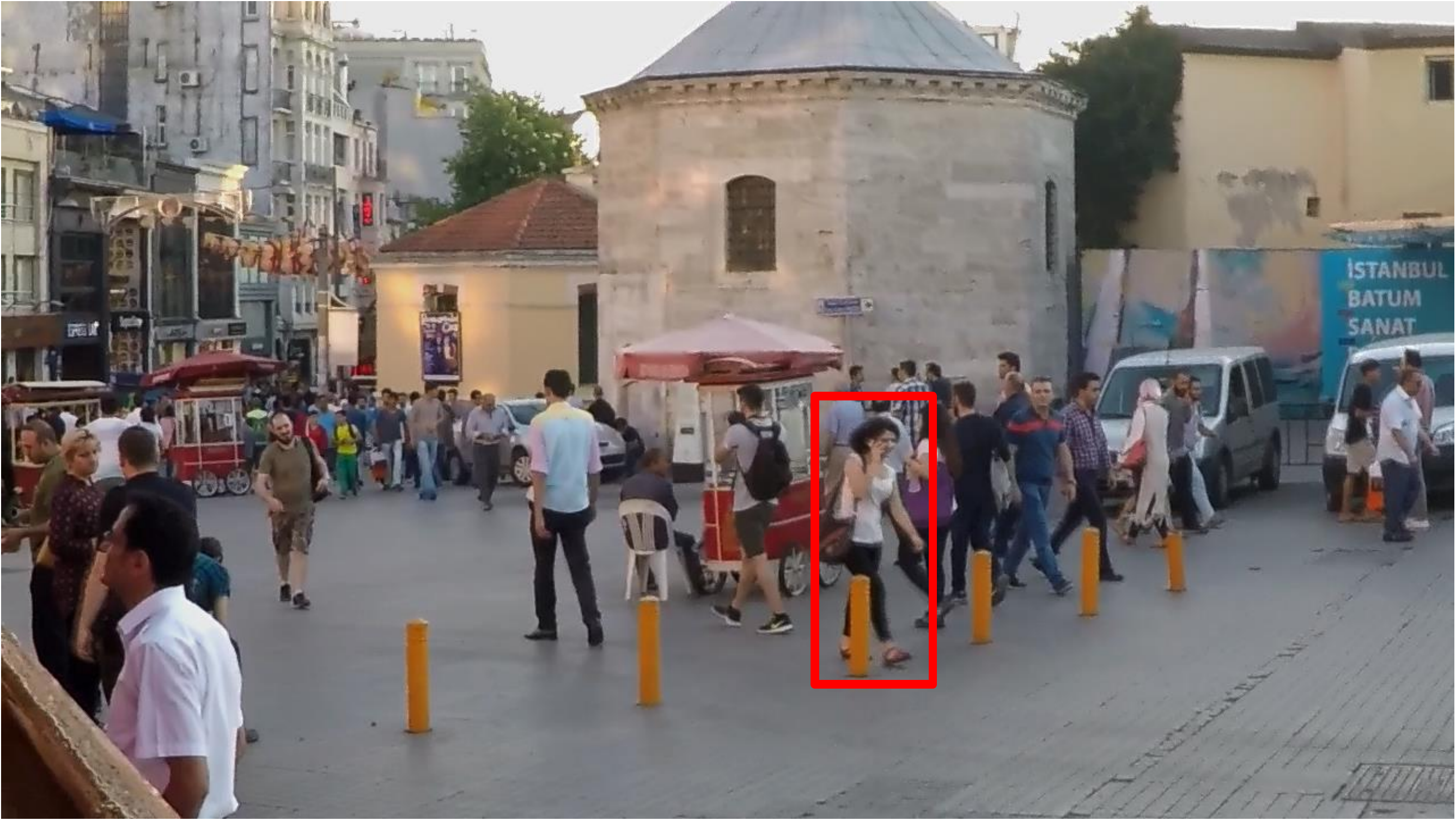}
\hspace{-0.4 cm}
&\includegraphics[height=0.130\textwidth]{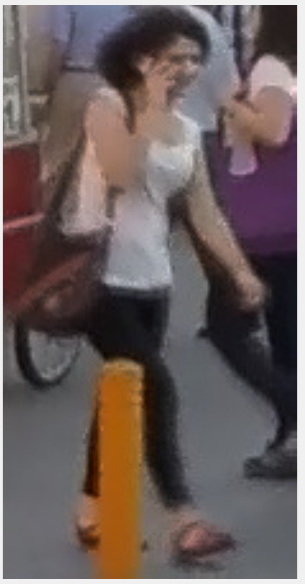}\\
\multicolumn{2}{c}{(c) Ours}
\end{tabular}
}\\
\multicolumn{6}{l}{
\hspace{-0.3 cm}
\begin{tabular}{ccccccc}
\includegraphics[height=0.130\textwidth]{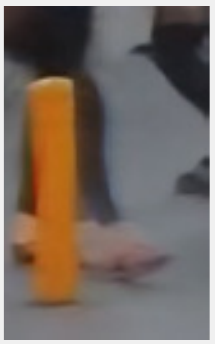}
\hspace{ 0.630 cm}
&\includegraphics[height=0.130\textwidth]{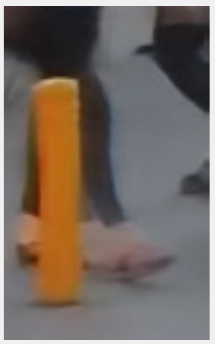}
\hspace{ 0.630 cm}
&\includegraphics[height=0.130\textwidth]{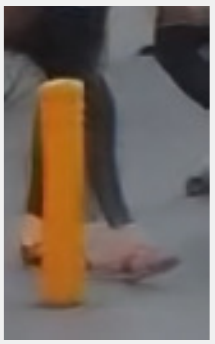}
\hspace{ 0.630 cm}
&\includegraphics[height=0.130\textwidth]{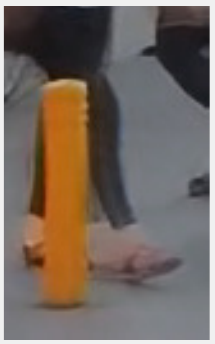}
\hspace{ 0.630 cm}
&\includegraphics[height=0.130\textwidth]{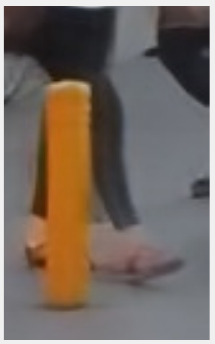}
\hspace{ 0.630 cm}
&\includegraphics[height=0.130\textwidth]{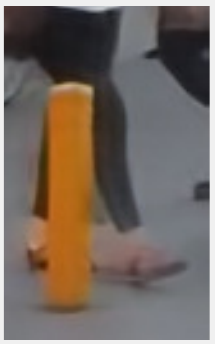}
\hspace{ 0.630 cm}
&\includegraphics[height=0.130\textwidth]{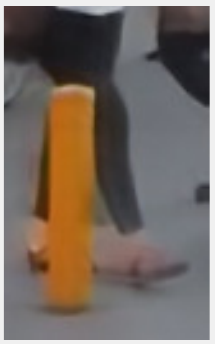}\\
\multicolumn{7}{c}{(d) The Reconstructed Video of \cite{Jin_2018_CVPR}}\\
\end{tabular}
}\\
\multicolumn{6}{l}{
\hspace{-0.3 cm}
\begin{tabular}{ccccccccccc}
\includegraphics[height=0.130\textwidth]{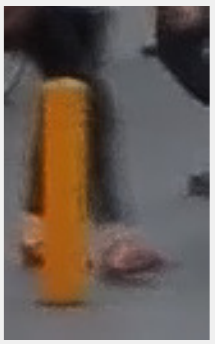}
\hspace{-0.39 cm}
&\includegraphics[height=0.130\textwidth]{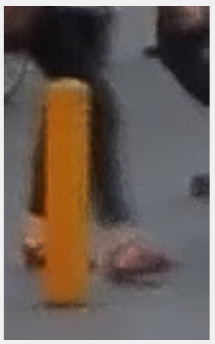}
\hspace{-0.39 cm}
&\includegraphics[height=0.130\textwidth]{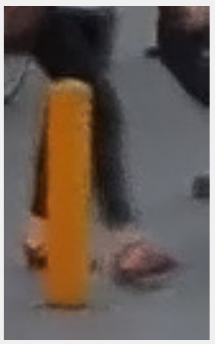}
\hspace{-0.39 cm}
&\includegraphics[height=0.130\textwidth]{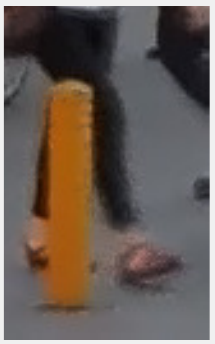}
\hspace{-0.39 cm}
&\includegraphics[height=0.130\textwidth]{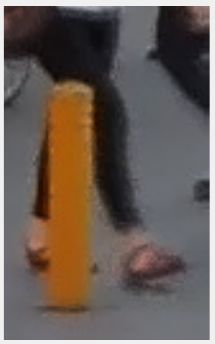}
\hspace{-0.39 cm}
&\includegraphics[height=0.130\textwidth]{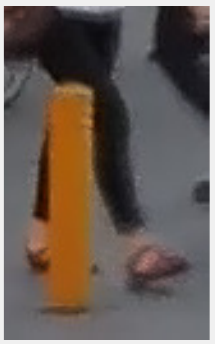}
\hspace{-0.39 cm}
&\includegraphics[height=0.130\textwidth]{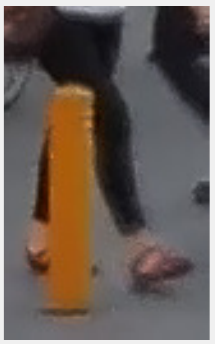}
\hspace{-0.39 cm}
&\includegraphics[height=0.130\textwidth]{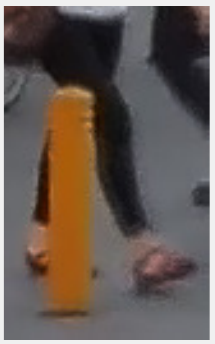}
\hspace{-0.39 cm}
&\includegraphics[height=0.130\textwidth]{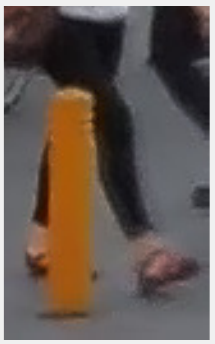}
\hspace{-0.39 cm}
&\includegraphics[height=0.130\textwidth]{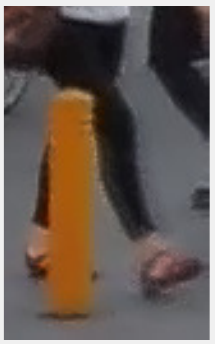}
\hspace{-0.39 cm}
&\includegraphics[height=0.130\textwidth]{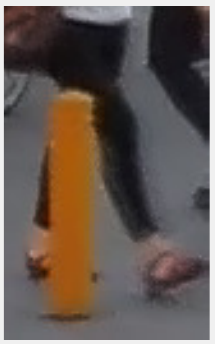}\\
\multicolumn{11}{c}{(e) The Reconstructed Video of Our Method}\\
\end{tabular}
}\\
\multicolumn{3}{c}{ 
\hspace{-0.3 cm}
\begin{tabular}{ccc}
\includegraphics[height=0.112\textwidth]{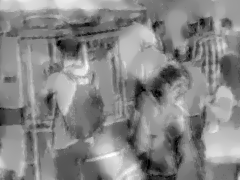}
\hspace{-0.30 cm}
&\includegraphics[height=0.112\textwidth]{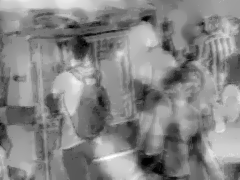}
\hspace{-0.30 cm}
&\includegraphics[height=0.112\textwidth]{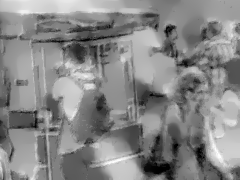}\\
\multicolumn{3}{c}{ (f) Reinbacher \etal \cite{Reinbacher16bmvc}}\\
\end{tabular}
}
&\multicolumn{3}{c}{ 
\hspace{-0.55 cm}
\begin{tabular}{ccc}
\includegraphics[height=0.112\textwidth]{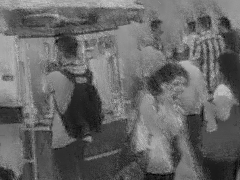}
\hspace{-0.30 cm}
&\includegraphics[height=0.112\textwidth]{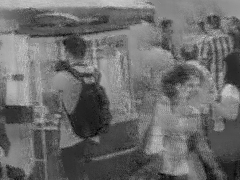}
\hspace{-0.30 cm}
&\includegraphics[height=0.112\textwidth]{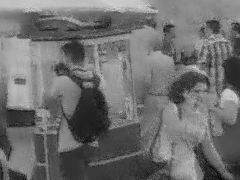}\\
\multicolumn{3}{c}{ (g) Scheerlinck \etal \cite{Scheerlinck18arxiv}}\\
\end{tabular}
}\\
\end{tabular}
\end{center}
\vspace{-2.5 mm}
\caption{\em \label{fig:VideoGoPro} An example of the reconstructed result on our synthetic event dataset based on the GoPro dataset~\cite{Nah_2017_CVPR}. \cite{Nah_2017_CVPR} provides videos to generate the blurry images and event data. 
(a) The blurry image. The red close-up frame is for (b)-(e), the yellow close-up frame is for (f)-(g). 
(b) The deblurring result of Jin \etal \cite{Jin_2018_CVPR}. 
(c) Our deblurring result. 
(d) The crop of their reconstructed images and the frame number is fixed at 7. Jin \etal \cite{Jin_2018_CVPR} uses the GoPro dataset added with 20 scenes as training data and  their model is supervised by 7 consecutive sharp frames.
(e) The crop of our reconstructed images. 
(f) The crop of Reinbacher \cite{Reinbacher16bmvc} reconstructed images from only events.
(g) The crop of Scheerlinck \cite{Scheerlinck18arxiv} reconstructed image, they use both events and the intensity image.
For (e)-(g), the shown frames are the chosen examples, where the length of the reconstructed video is based on the number of events. 
}
\vspace{-1mm}
\end{figure*}
\begin{figure*}[ht]
\begin{center}
\resizebox{\textwidth}{!}{
\begin{tabular}{ccccc}
\includegraphics[width=0.215\textwidth]{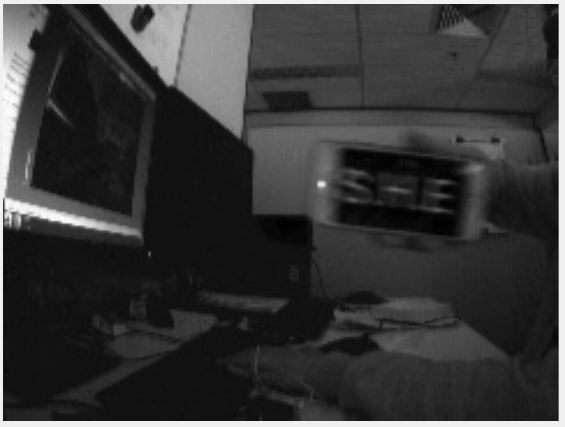}
&\includegraphics[width=0.215\textwidth]{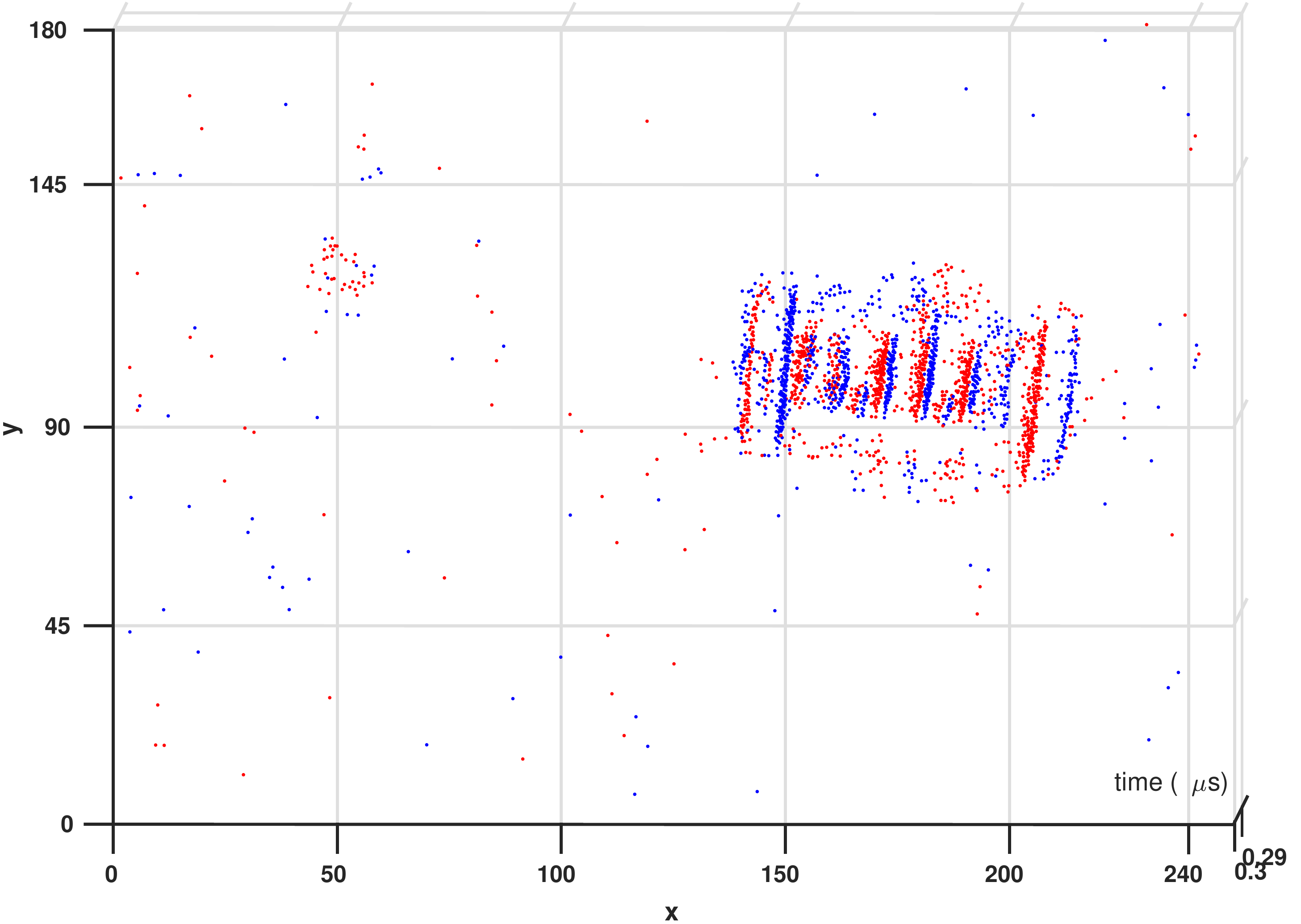}
&\includegraphics[width=0.215\textwidth]{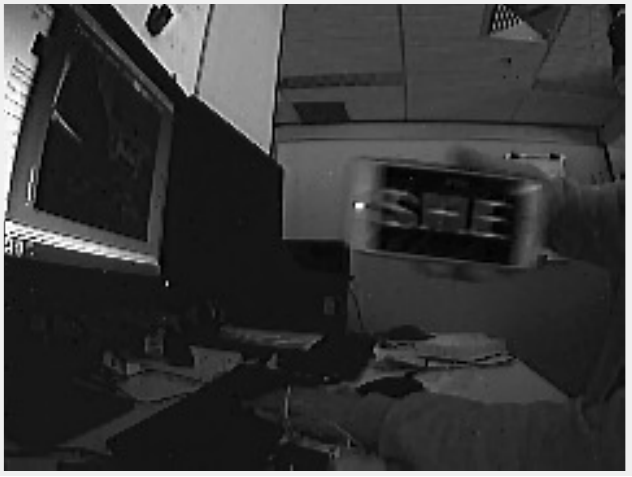}
&\includegraphics[width=0.215\textwidth]{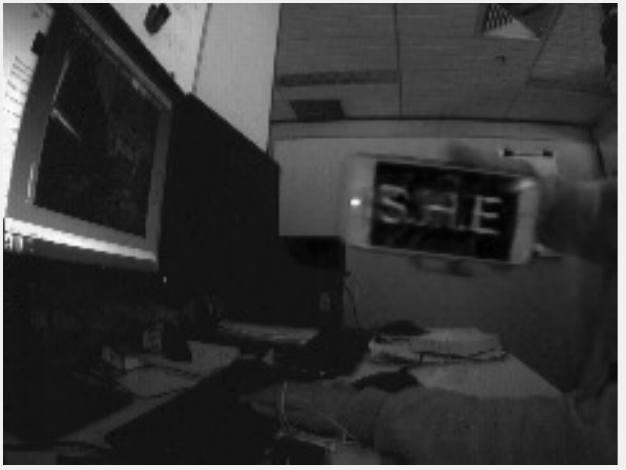}
&\includegraphics[width=0.215\textwidth]{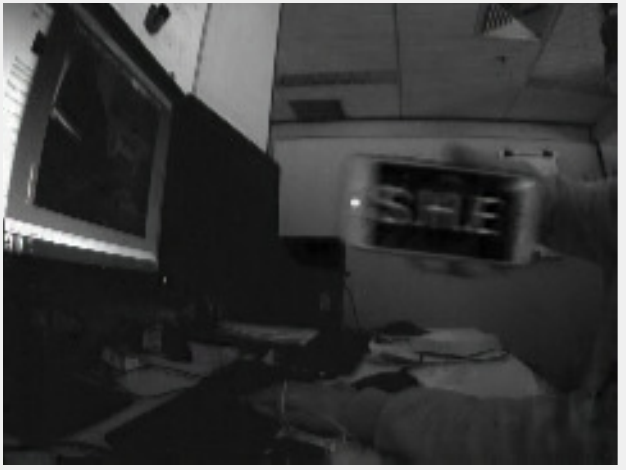}\\
\includegraphics[width=0.215\textwidth]{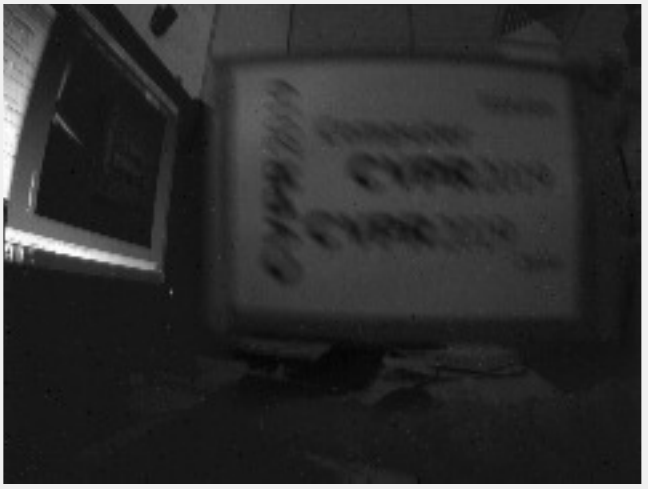}
&\includegraphics[width=0.215\textwidth]{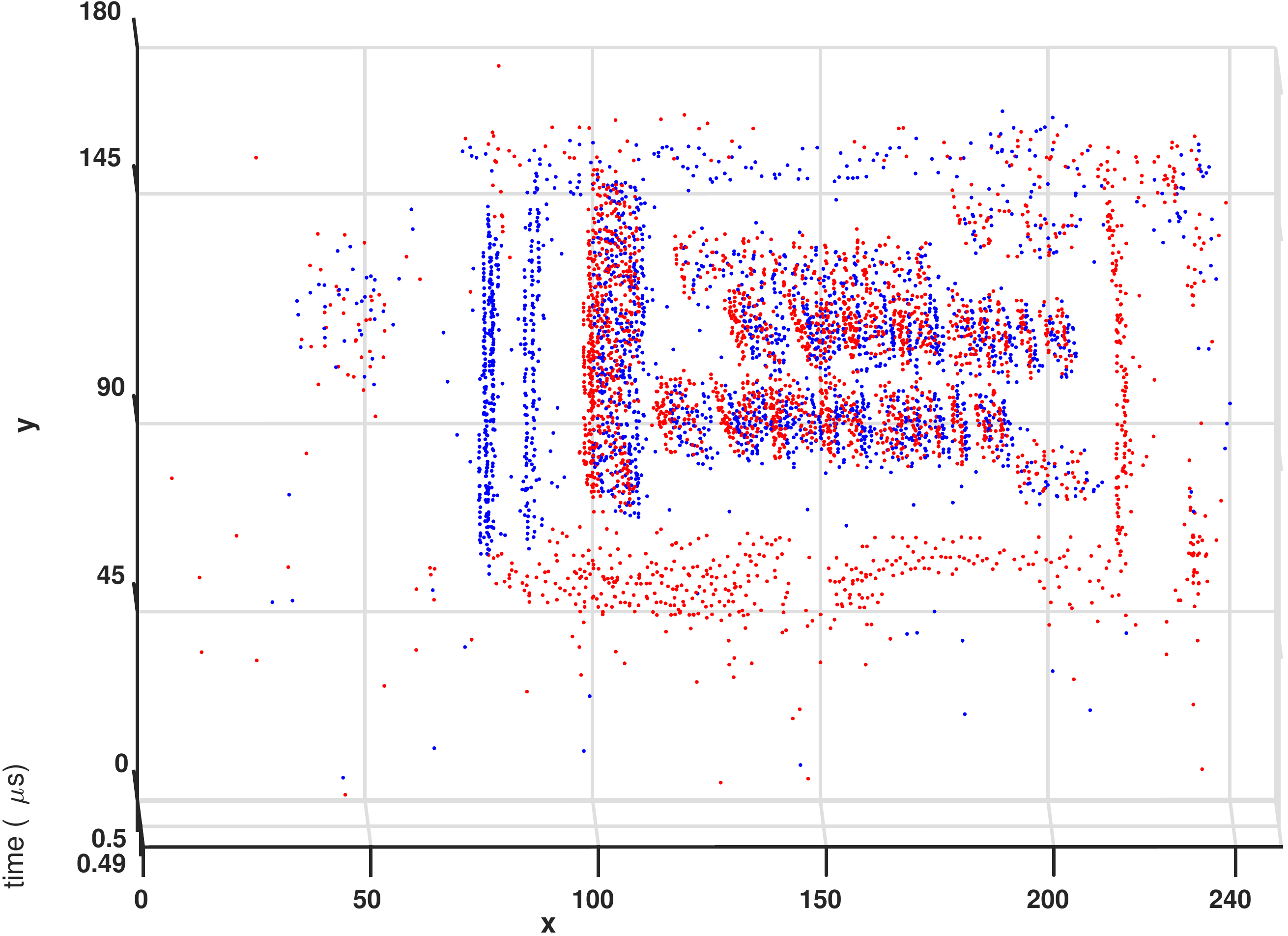}
&\includegraphics[width=0.215\textwidth]{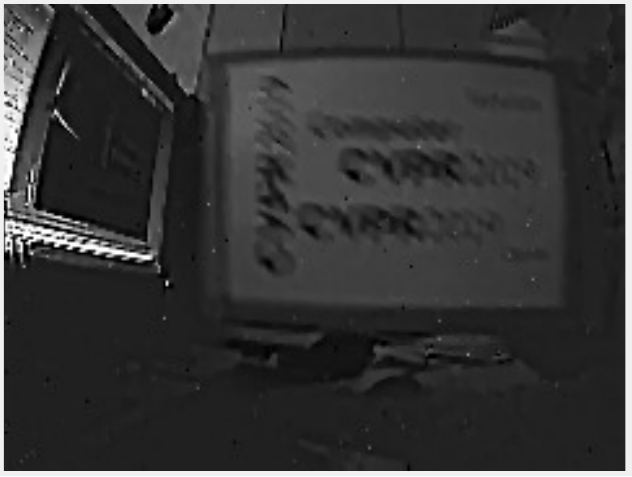}
&\includegraphics[width=0.215\textwidth]{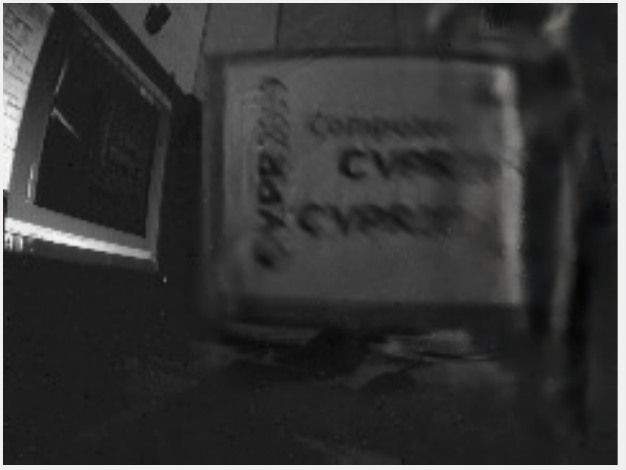}
&\includegraphics[width=0.215\textwidth]{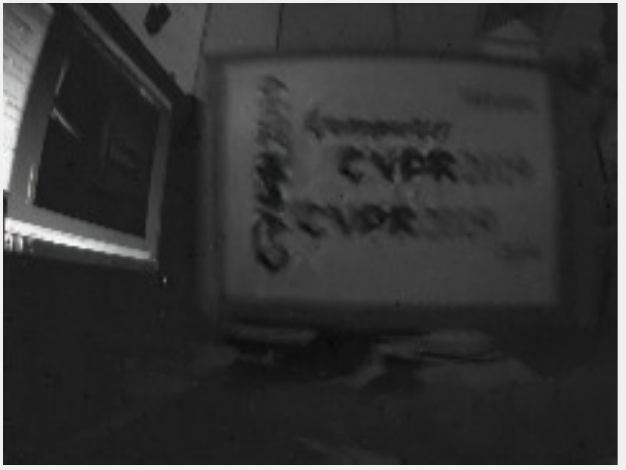}\\
(a) The Blurry Image  
&(b) The Event
&(c) Pan \etal \cite{pan2017deblurring}
&(d) Tao \etal \cite{Tao_2018_CVPR}
&(e) Nah \etal \cite{Nah_2017_CVPR}\\
\includegraphics[width=0.215\textwidth]{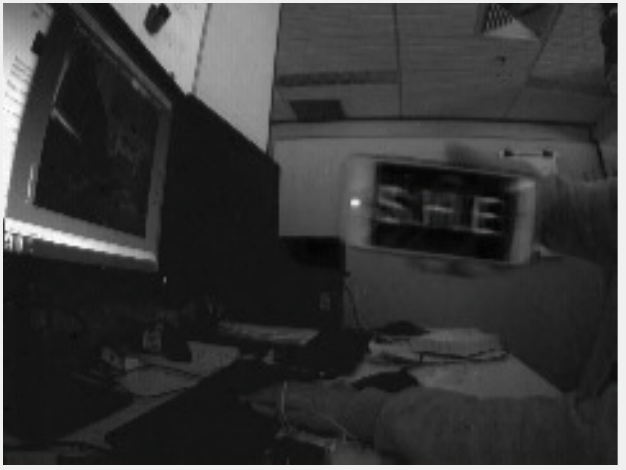}
&\includegraphics[width=0.215\textwidth]{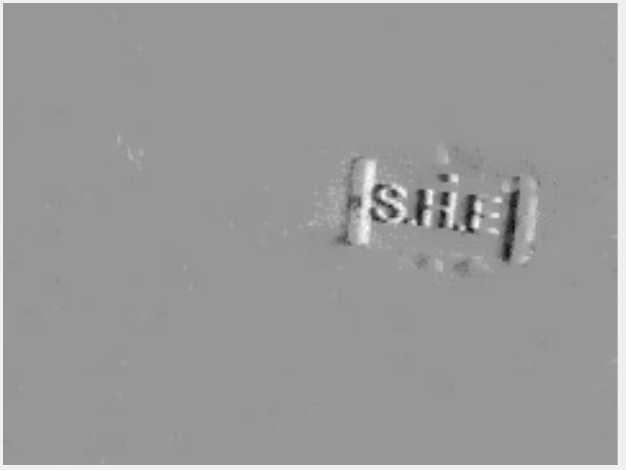}
&\includegraphics[width=0.215\textwidth]{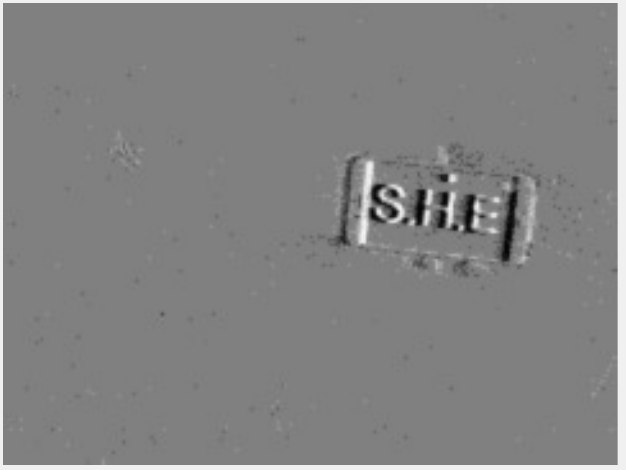}
&\includegraphics[width=0.215\textwidth]{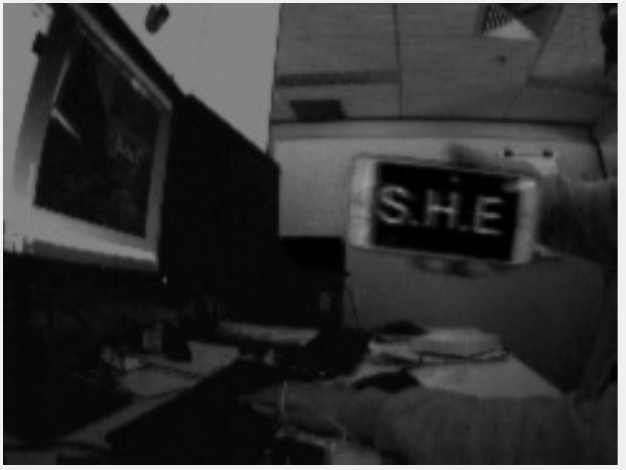}
&\includegraphics[width=0.215\textwidth]{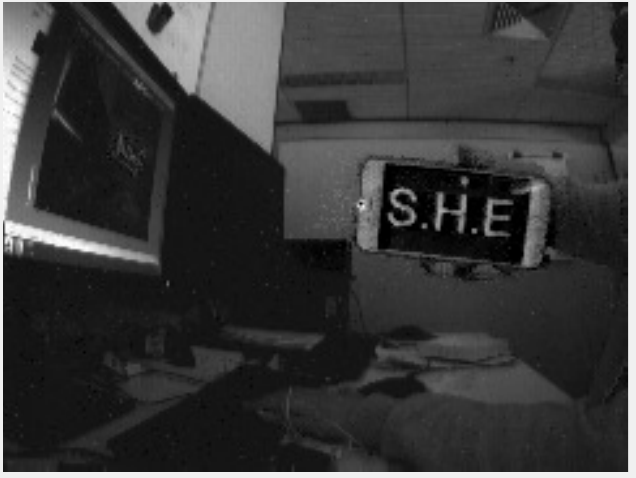}\\
\includegraphics[width=0.215\textwidth]{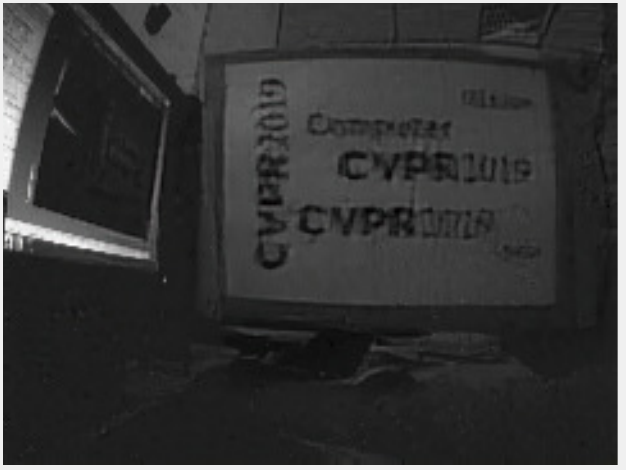}
&\includegraphics[width=0.215\textwidth]{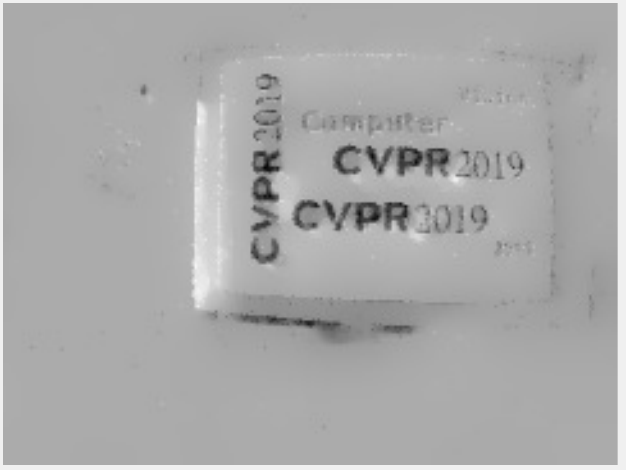}
&\includegraphics[width=0.215\textwidth]{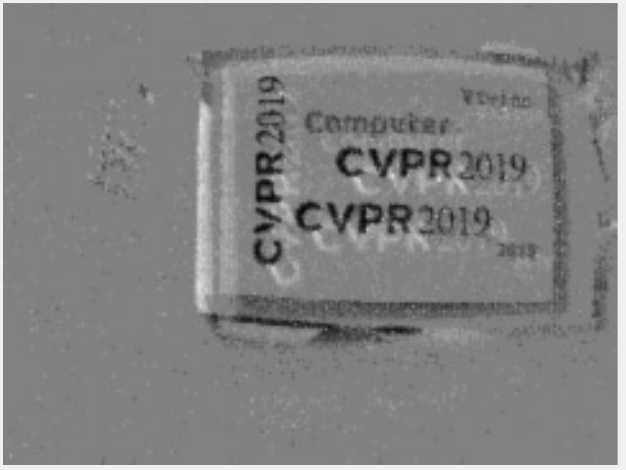}
&\includegraphics[width=0.215\textwidth]{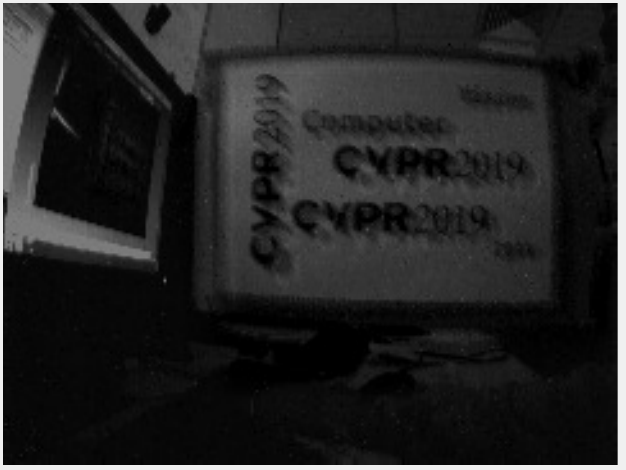}
&\includegraphics[width=0.215\textwidth]{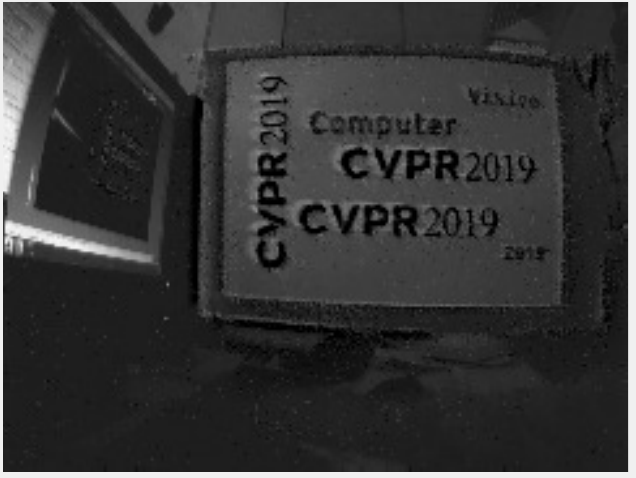}\\
(f) Jin \etal \cite{Jin_2018_CVPR} 
&(g) Reinbacher \etal \cite{Reinbacher16bmvc}  
&(h)  {\small\begin{tabular}[c]{@{}c@{}}Scheerlinck \etal \cite{Scheerlinck18arxiv}\\ (events only)\end{tabular}}
&(i) Scheerlinck \etal \cite{Scheerlinck18arxiv}
&(j) Ours\\
\end{tabular}
}
\end{center}
\vspace{-2.5 mm}
 \caption{\em \label{fig:Real} Examples of reconstruction result on our real {\it blurry event dataset} in low lighting and complex dynamic conditions
(a) Input blurry images. 
(b) The event information. 
(c) Deblurring results of \cite{pan2017deblurring}. 
(d) Deblurring results of \cite{Tao_2018_CVPR}. 
(e) Deblurring results of \cite{Nah_2017_CVPR}.  
(f) Deblurring results of \cite{Jin_2018_CVPR} and they use video as training data. 
(g) Reconstruction result of \cite{Reinbacher16bmvc} from only events. 
(h)-(i) Reconstruction results of \cite{Scheerlinck18arxiv}, (h) from only events, (i) from combining events and frames. 
(j) Our reconstruction result. 
Results in (c)-(f) show that real high dynamic settings and low light condition is still challenging in the deblurring area. Results in (g)-(h) show that while intensity information of a scene is still retained with an event camera recording, color, and delicate texture information cannot be recovered. 
}\vspace{-1mm}
\end{figure*}
\vspace{-2 mm}
\section{Experiment}

\subsection{Experimental Setup}

\vspace{-1mm}
\noindent{\bf{Synthetic dataset.}} 
In order to provide a quantitative comparison, we build a synthetic dataset based on the GoPro blurry dataset \cite{Nah_2017_CVPR}.
It supplies ground truth videos which are used to generate the blurry images. Similarly, we employ the ground-truth images to generate event data based on the methodology of \emph{event camera model}. 

\noindent{\bf{Real dataset.}} 
We evaluate our method on a public Event-Camera dataset \cite{mueggler2017event}, which provides a collection of sequences captured by the event camera for high-speed robotics.
Furthermore, we present our real {\it blurry event dataset}~\footnote{To be released with codes}, where each real sequence is captured with the DAVIS\cite{brandli2014240} under different conditions, such as indoor, outdoor scenery, low lighting conditions, and different motion patterns (\eg, camera shake, objects motion) that naturally introduce motion blur into the APS intensity images.

\noindent{\bf{Implementation details.}} 
For all our real experiments, we use the DAVIS that shares photosensor array to simultaneously output events (DVS) and intensity images (APS).
The framework is implemented by using MATLAB with C++ wrappers. It takes around 1.5 second to process one image on a single i7 core running at 3.6 GHz. 


\subsection{Experimental Results}
\vspace{-1mm}
We compare our proposed approach with state-of-the-art blind deblurring methods, including conventional deblurring methods \cite{pan2017deblurring,yan2017image}, deep based dynamic scene deblurring methods \cite{Nah_2017_CVPR,Jin_2018_CVPR,Tao_2018_CVPR,Zhang_2018_CVPR,sun2015learning}, and event-based image reconstruction methods \cite{Reinbacher16bmvc,Scheerlinck18arxiv}. Moreover, Jin \etal \cite{Jin_2018_CVPR} can restore a video from a single blurry image based on a deep network, where the middle frame in the restored odd-numbered sequence is the best.

In order to prove the effective of our \textbf{EDI} model, we show some baseline comparisons in Fig. \ref{fig:baseline} and Table \ref{all_all}. 
For baseline 1, we first apply a state-of-the-art deblurring method~\cite{Tao_2018_CVPR} to recover a sharp image, and then the recovered image as an input is then fed to a reconstruction method~\cite{Scheerlinck18arxiv}.
For baseline 2, we first use the video reconstruction method to construct a sequence of intensity images, and then apply the deblurring method to each frame.
As seen in Table~\ref{all_all}, our approach obtains higher PSNR and SSIM in comparison to both baseline 1 and baseline 2. This also implies that our approach better exploits the event data to not only recover sharp images but also reconstruct high frame-rate videos.

In Table~\ref{all_all}, we show the quantitative comparisons with the state-of-the-art image deblurring approaches \cite{sun2015learning, pan2017deblurring, gong2017motion, Jin_2018_CVPR, Tao_2018_CVPR, Zhang_2018_CVPR, Nah_2017_CVPR}, and the video reconstruction method \cite{Scheerlinck18arxiv} on our synthetic dataset, respectively.
As indicated in Table~\ref{all_all}, our approach achieves the best performance on SSIM and competitive result on PSNR compared to the state-of-the-art methods, and attains significant performance improvements on high-frame video reconstruction. 

In Fig.~\ref{fig:VideoGoPro}, we compare our generated video frames with the state-of-the-art deblurring methods \cite{ pan2017deblurring, Jin_2018_CVPR, Tao_2018_CVPR, Nah_2017_CVPR} qualitatively. Furthermore, image reconstruction methods \cite{Reinbacher16bmvc,Scheerlinck18arxiv} are also included for comparisons. Fig.~\ref{fig:VideoGoPro} shows that our method can generate more frames from a single blurry image and the recovered frames are much sharper.

We also report our reconstruction results on the real dataset, including text images and low-lighting images, in Fig.~\ref{fig:eventdeblur}, Fig.~\ref{fig:eventsample}, Fig.~\ref{fig:real23} and Fig.~\ref{fig:Real}. Compared with state-of-the-art deblurring methods, our method achieves superior results. 
In comparison to existing event-based image reconstructed methods \cite{Reinbacher16bmvc, Scheerlinck18arxiv}, our reconstructed images are not only more realistic but also contain richer details. 
More deblurring results and {\bf high-temporal resolution videos} are shown in the supplementary material.

\vspace{-1.5 mm}
\section{Conclusion}
\vspace{-1 mm}
In this paper, we propose a {\textbf{Event-based Double Integral (EDI)}} model to naturally connect intensity images and event data captured by the event camera, which also takes the blur generation process into account.
In this way, our model can be used to not only recover latent sharp images but also reconstruct intermediate frames at high frame-rate. We also propose a simple yet effective method to solve our EDI model.
Due to the simplicity of our optimization process, our method is efficient as well. Extensive experiments show that our method can generate high-quality high frame-rate videos efficiently under different conditions, such as low lighting and complex dynamic scenes. 

\bibliographystyle{ieee}
\bibliography{deblurring_bib,event_camera_bib}

\clearpage
\appendix

\section*{Appendix}
In this supplementary material, we provide more details about our datasets (Sec. \ref{sec::dataset}). Section \ref{sec::Video} show details in high frame-rate video generation. We also give an example of how to run our testing code (Sec. \ref{sec::Code}).

\begin{appendices}




\section{Datasets} \label{sec::dataset}

\subsection{Synthetic Dataset}\label{sec::Synthetic_dataset}

\begin{figure*}[ht]
\begin{center}
\resizebox{\textwidth}{!}{
\begin{tabular}{cccc}
\includegraphics[width=0.235\textwidth]{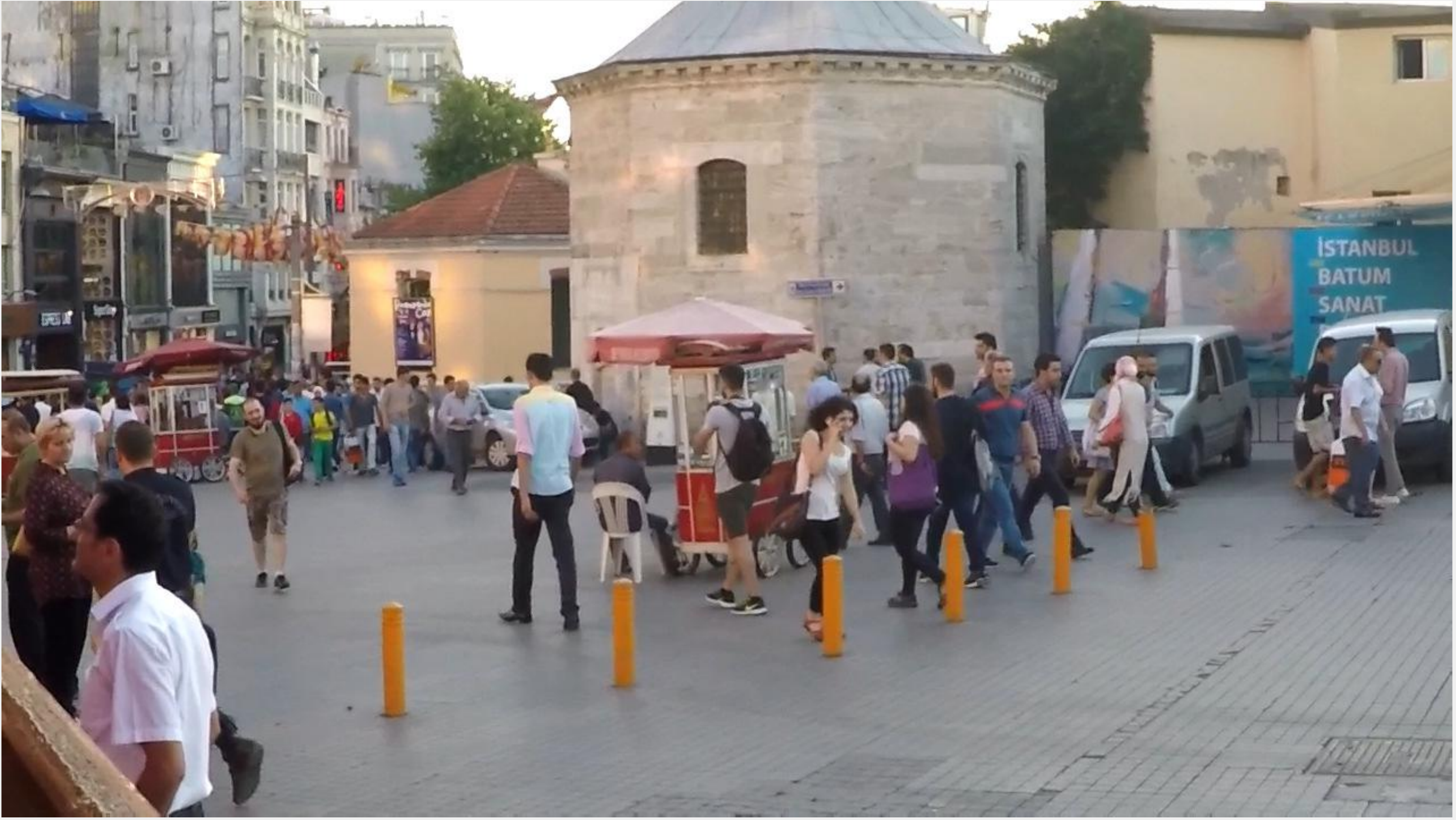}
&\includegraphics[width=0.235\textwidth]{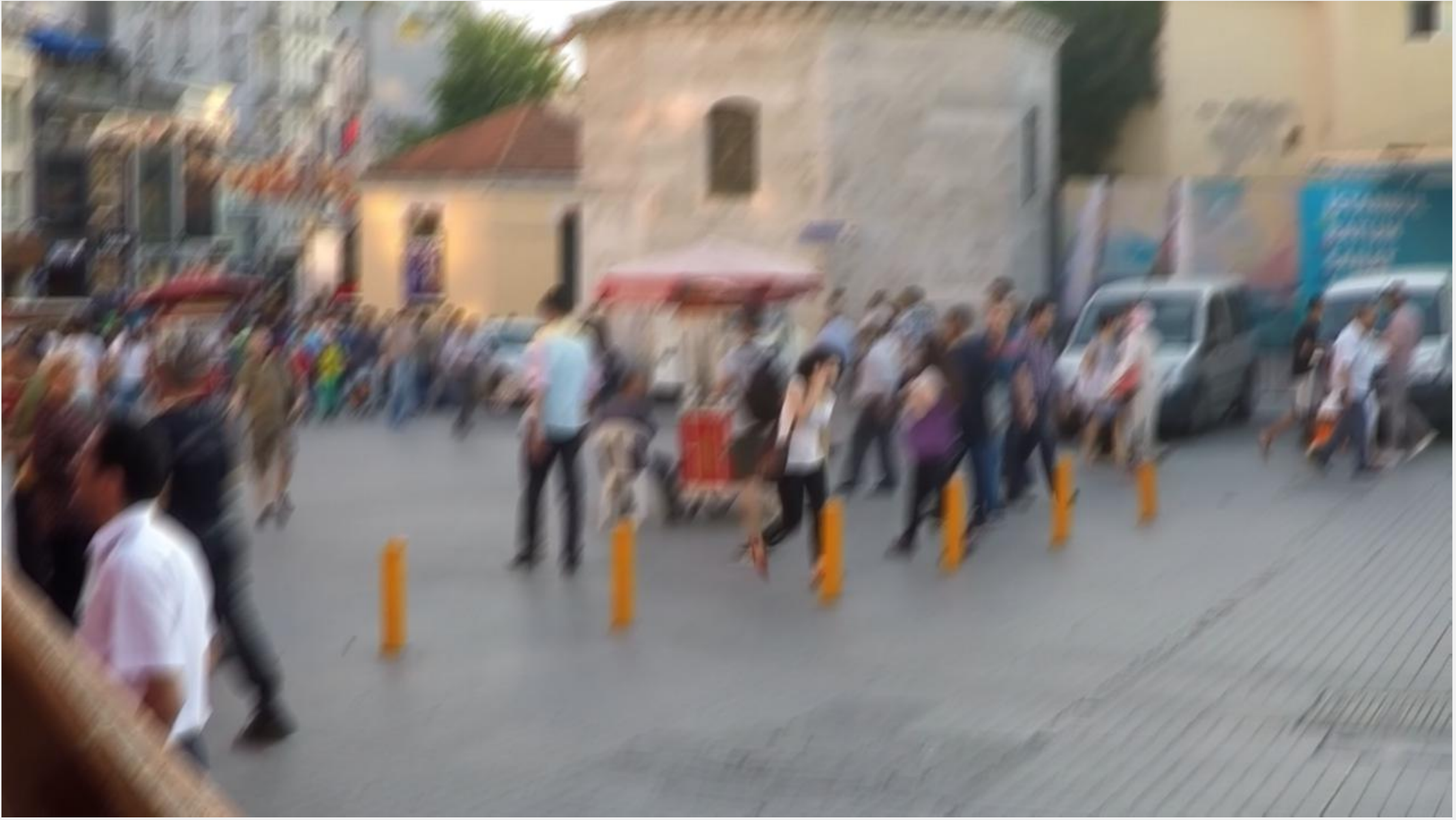}
&\includegraphics[width=0.235\textwidth]{./cGOPR/3_blur.pdf}
&\includegraphics[width=0.235\textwidth]{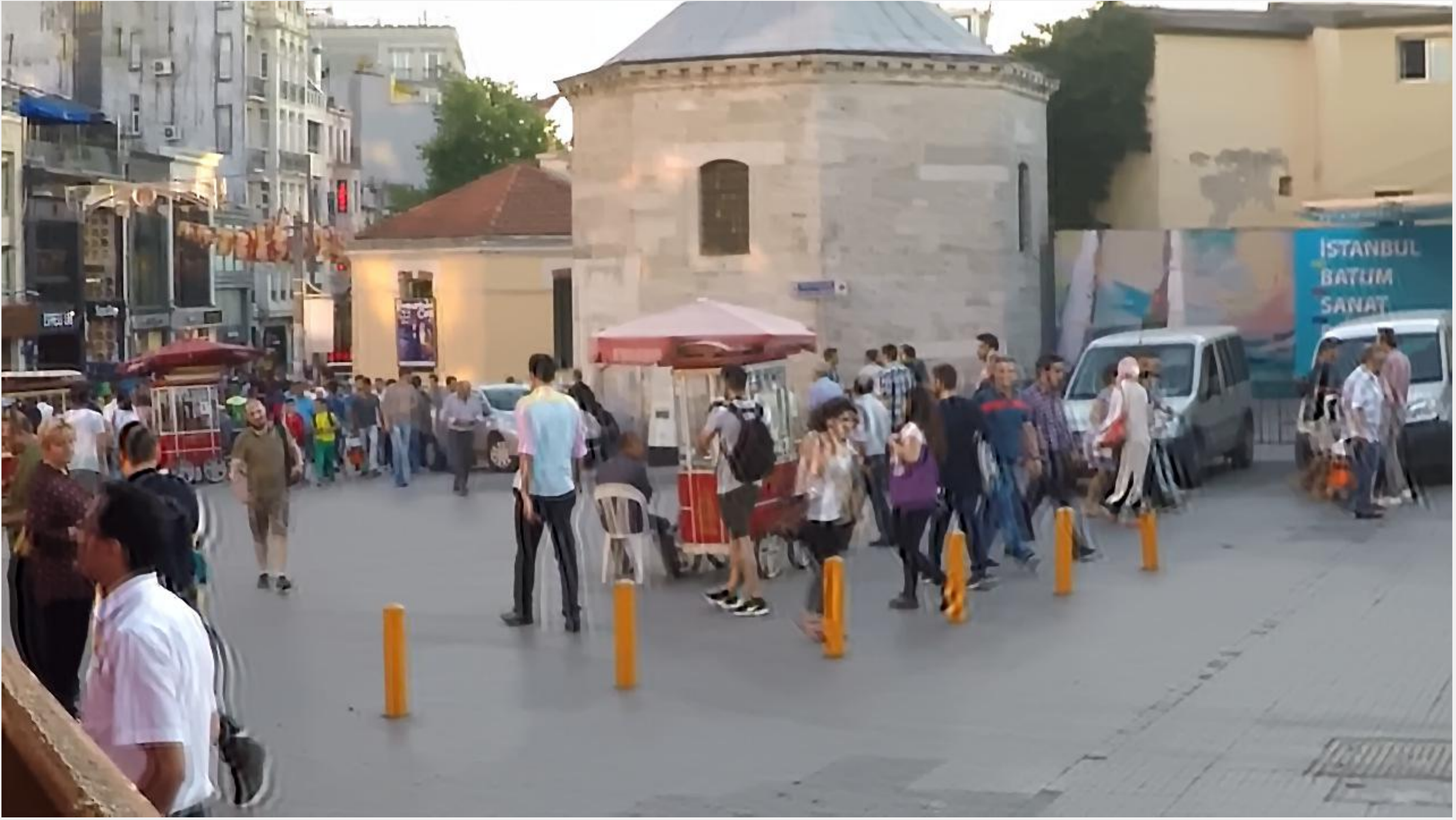}\\
\includegraphics[width=0.235\textwidth]{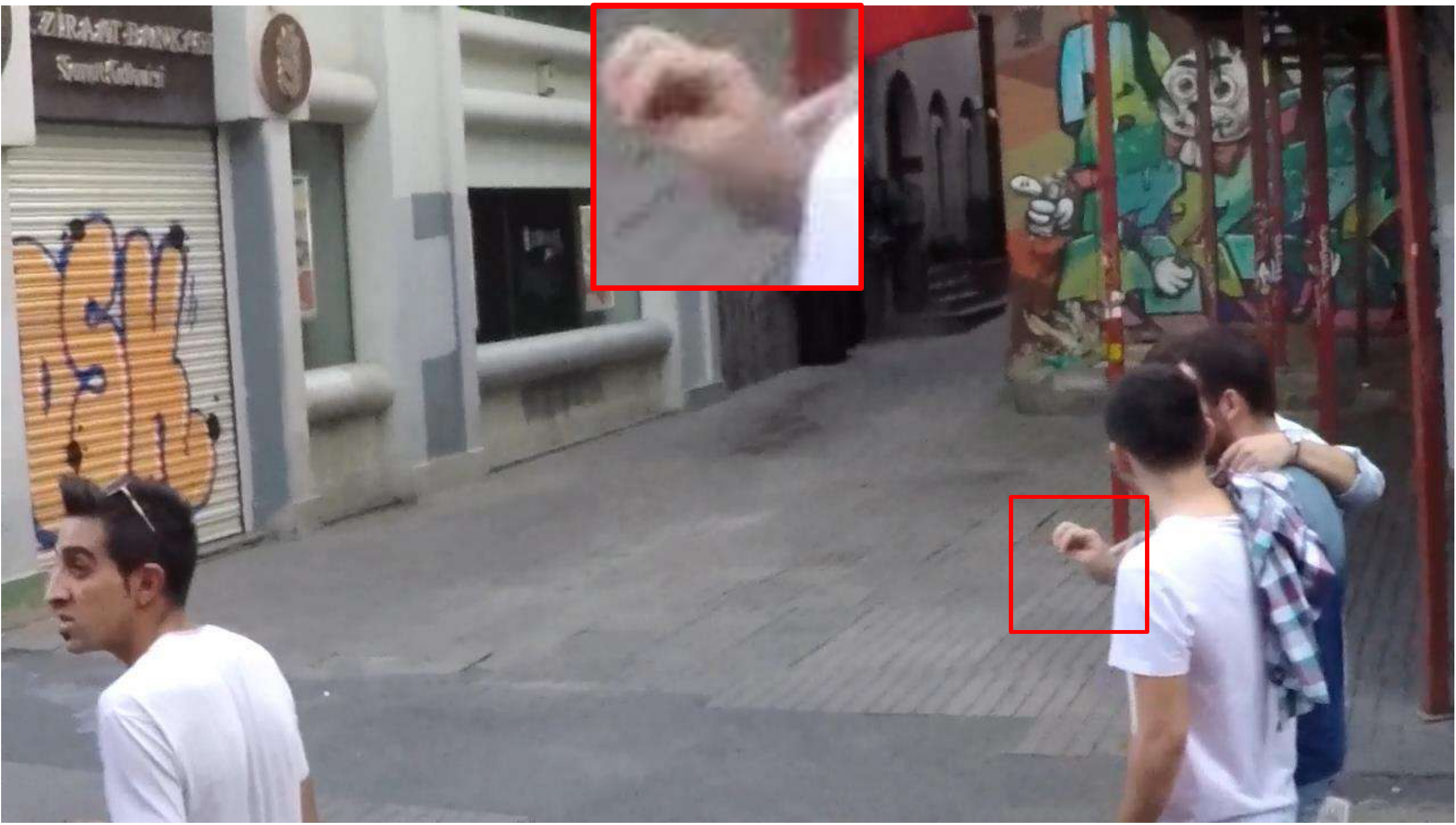}
&\includegraphics[width=0.235\textwidth]{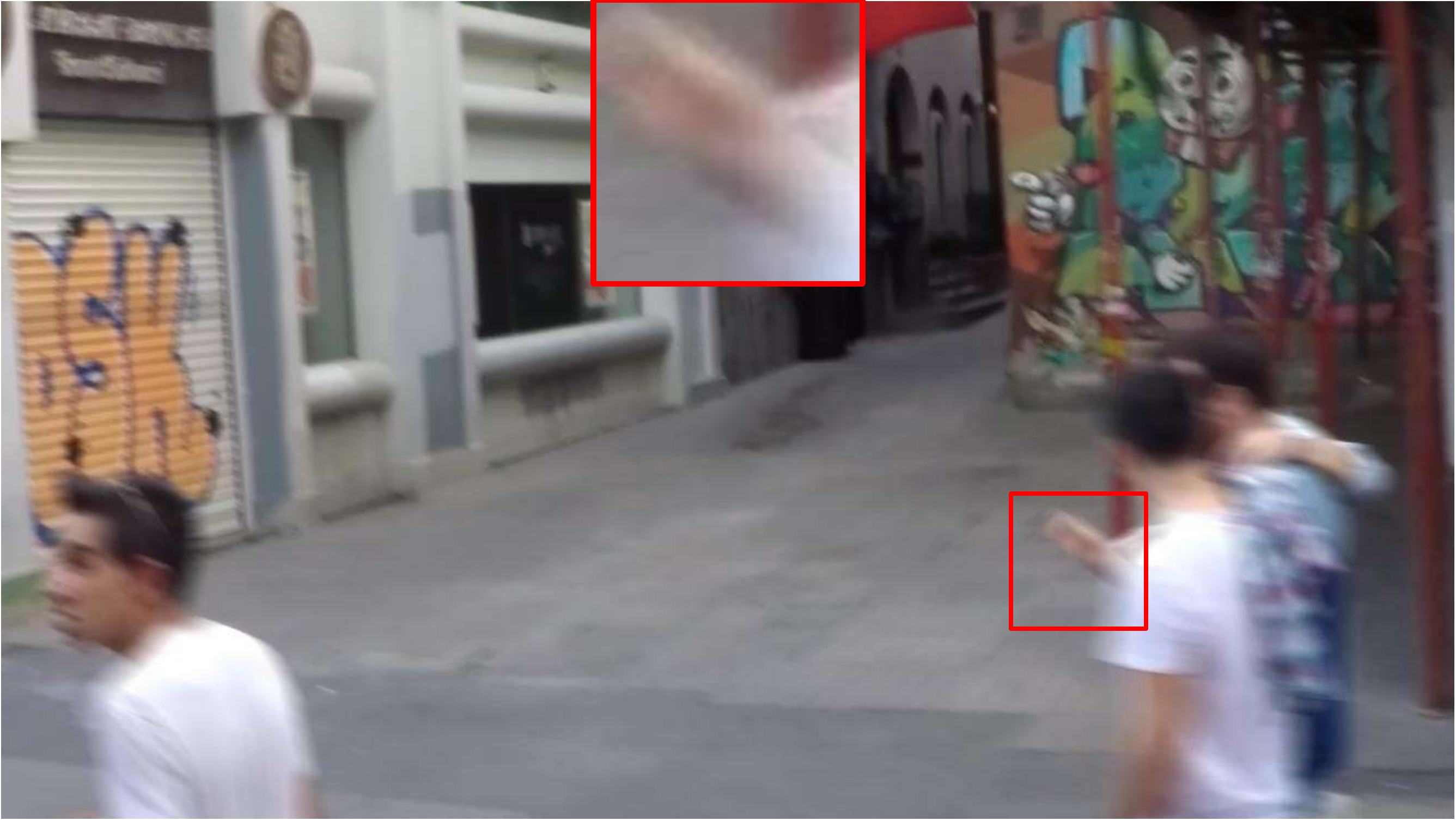}
&\includegraphics[width=0.235\textwidth]{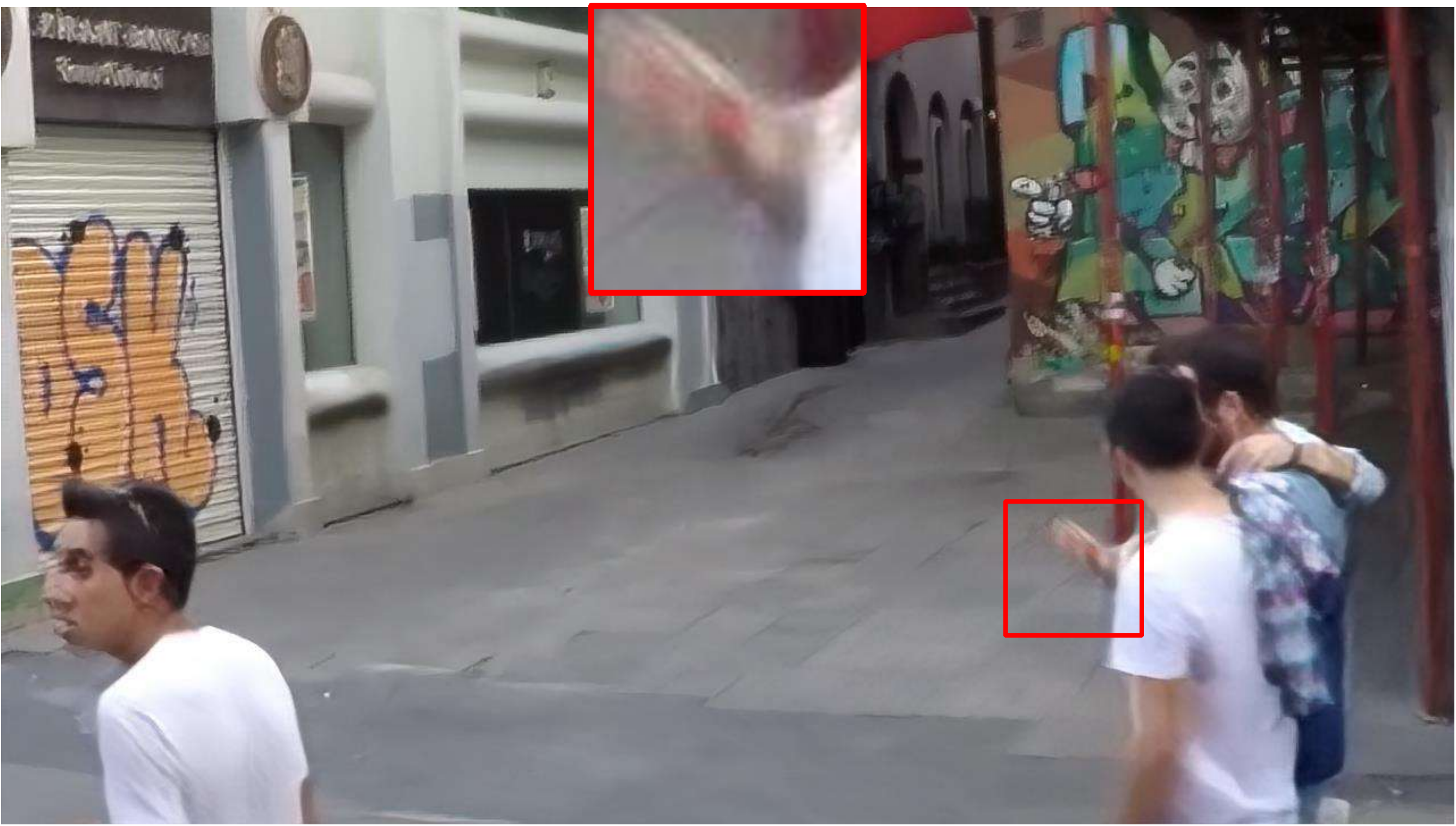}
&\includegraphics[width=0.235\textwidth]{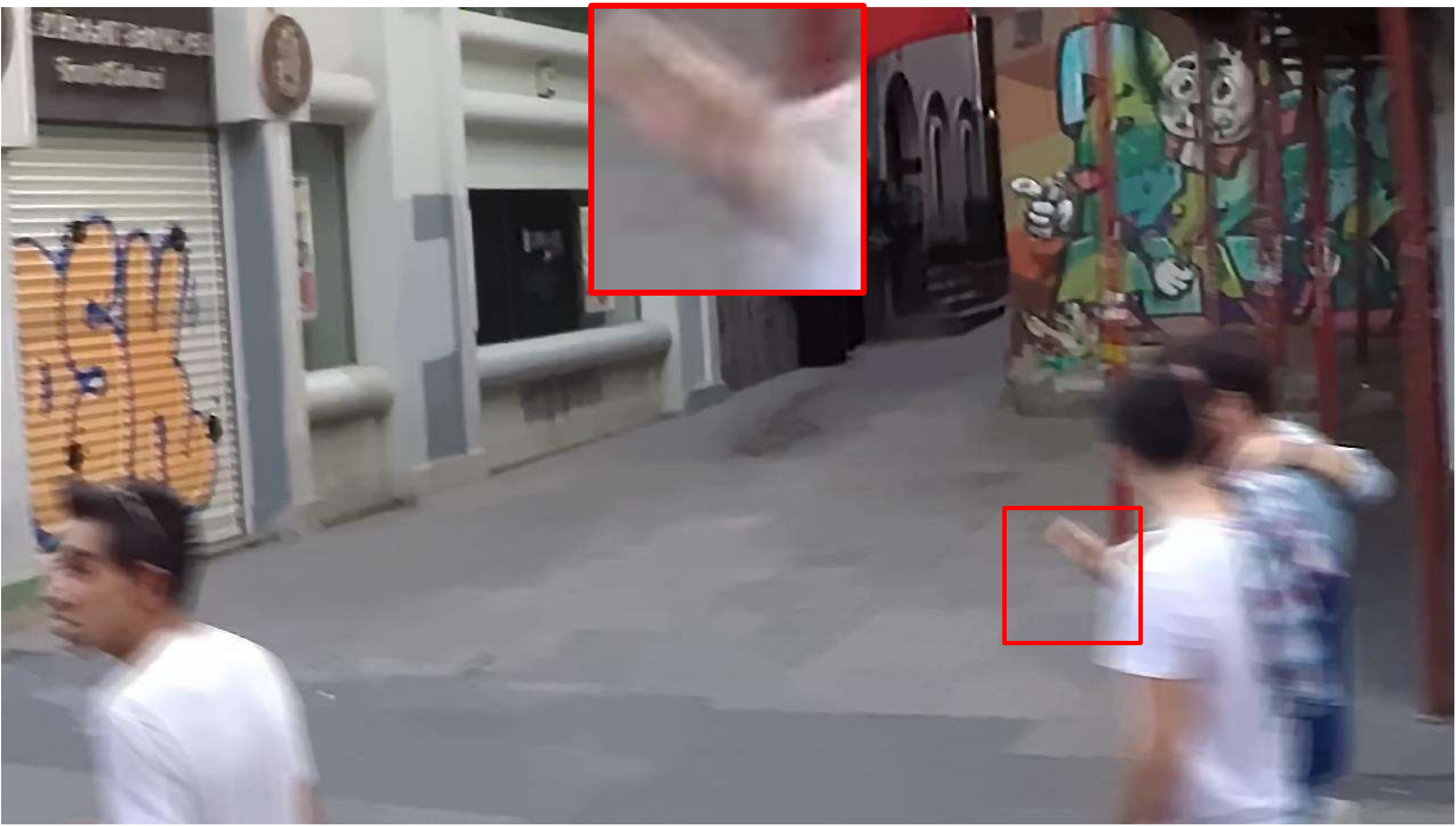}\\
\includegraphics[width=0.235\textwidth]{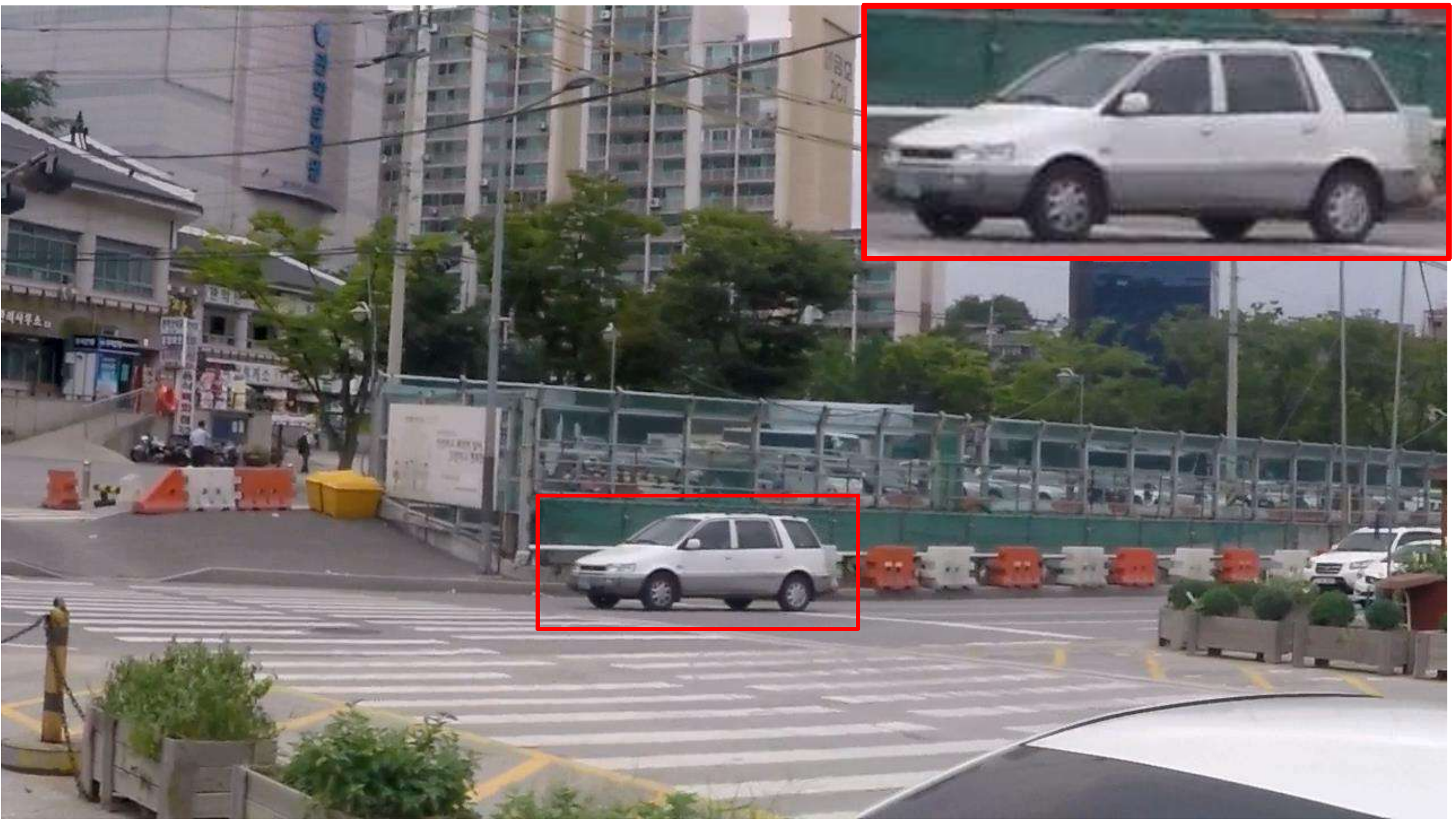}
&\includegraphics[width=0.235\textwidth]{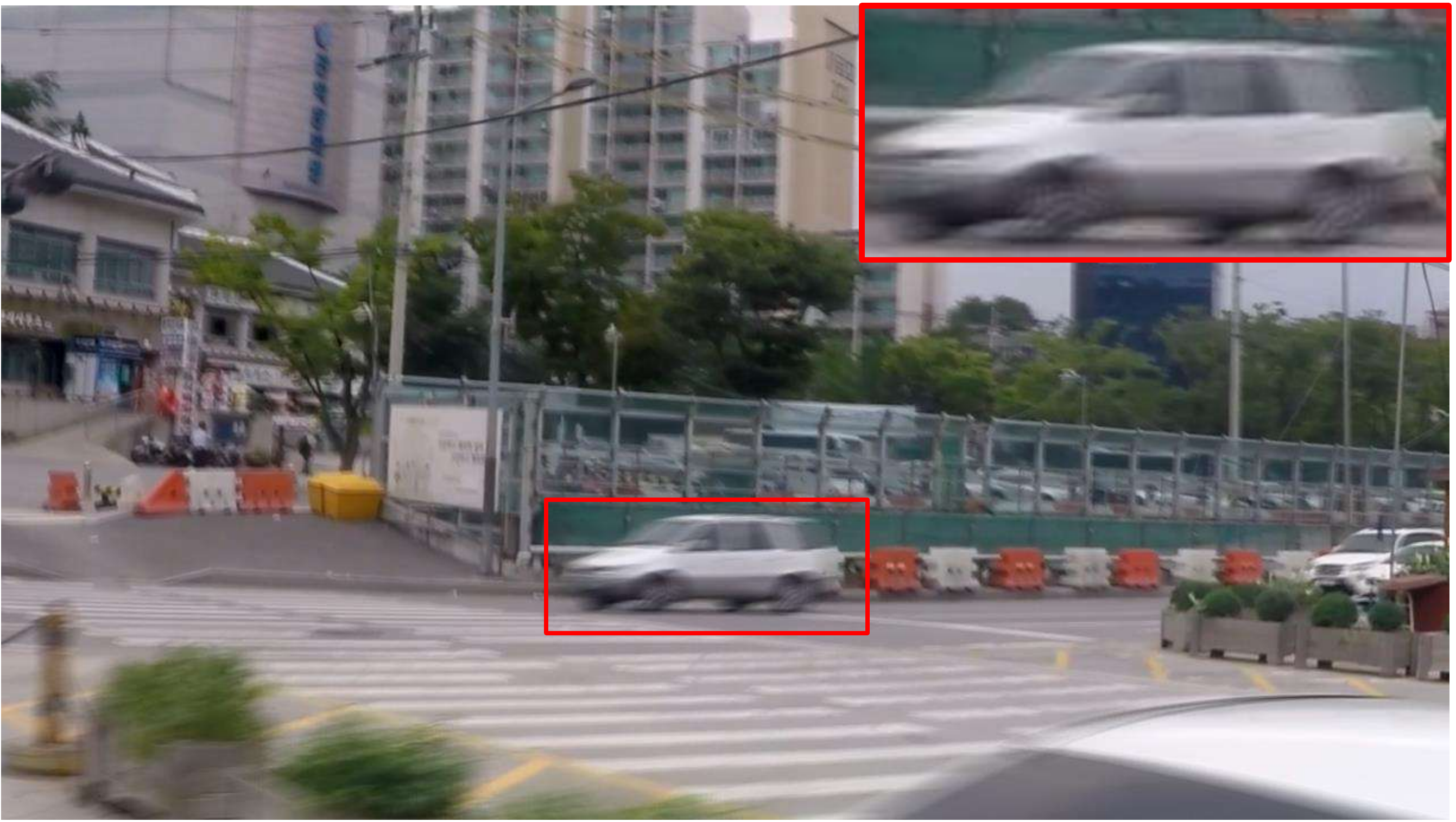}
&\includegraphics[width=0.235\textwidth]{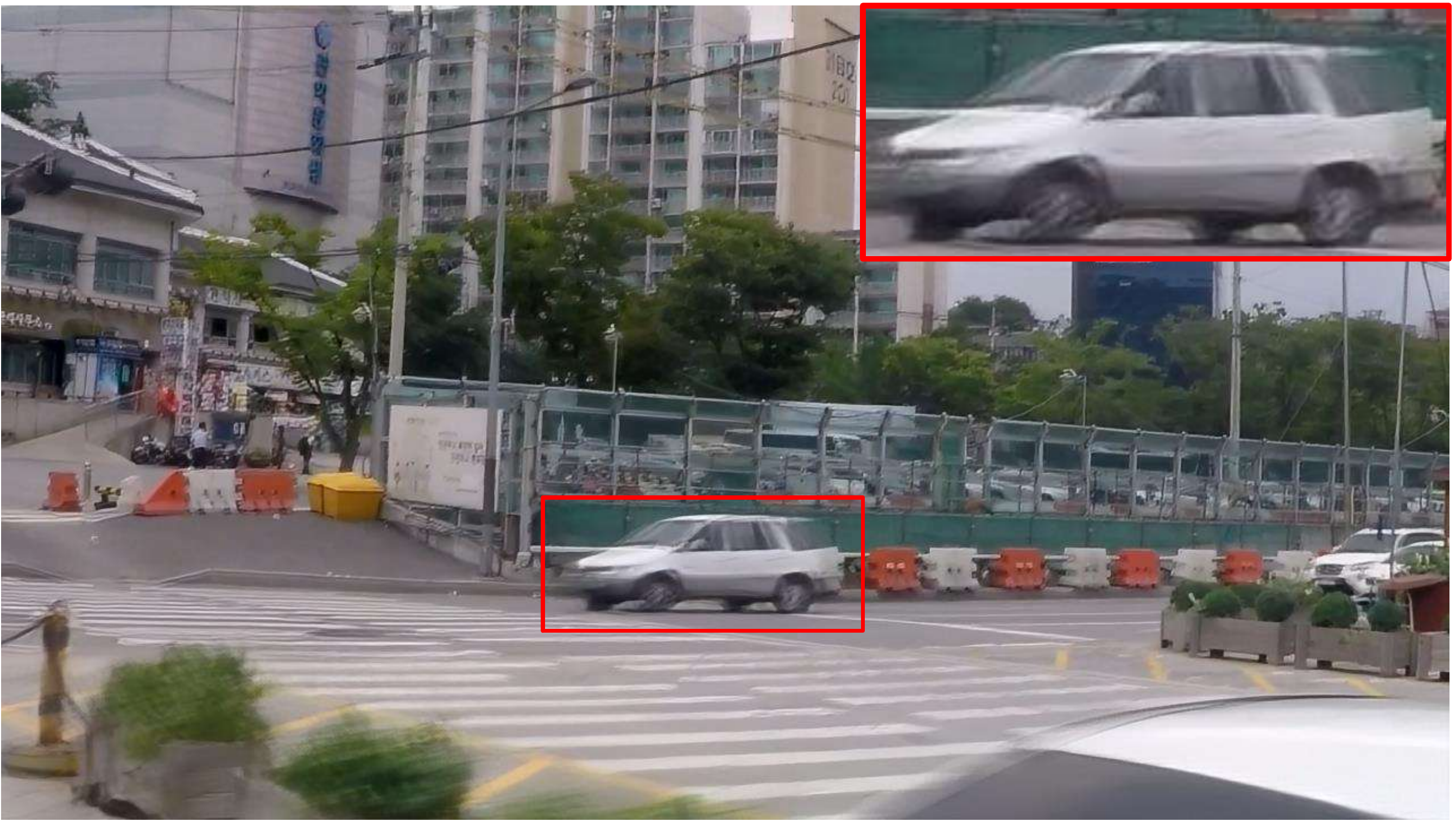}
&\includegraphics[width=0.235\textwidth]{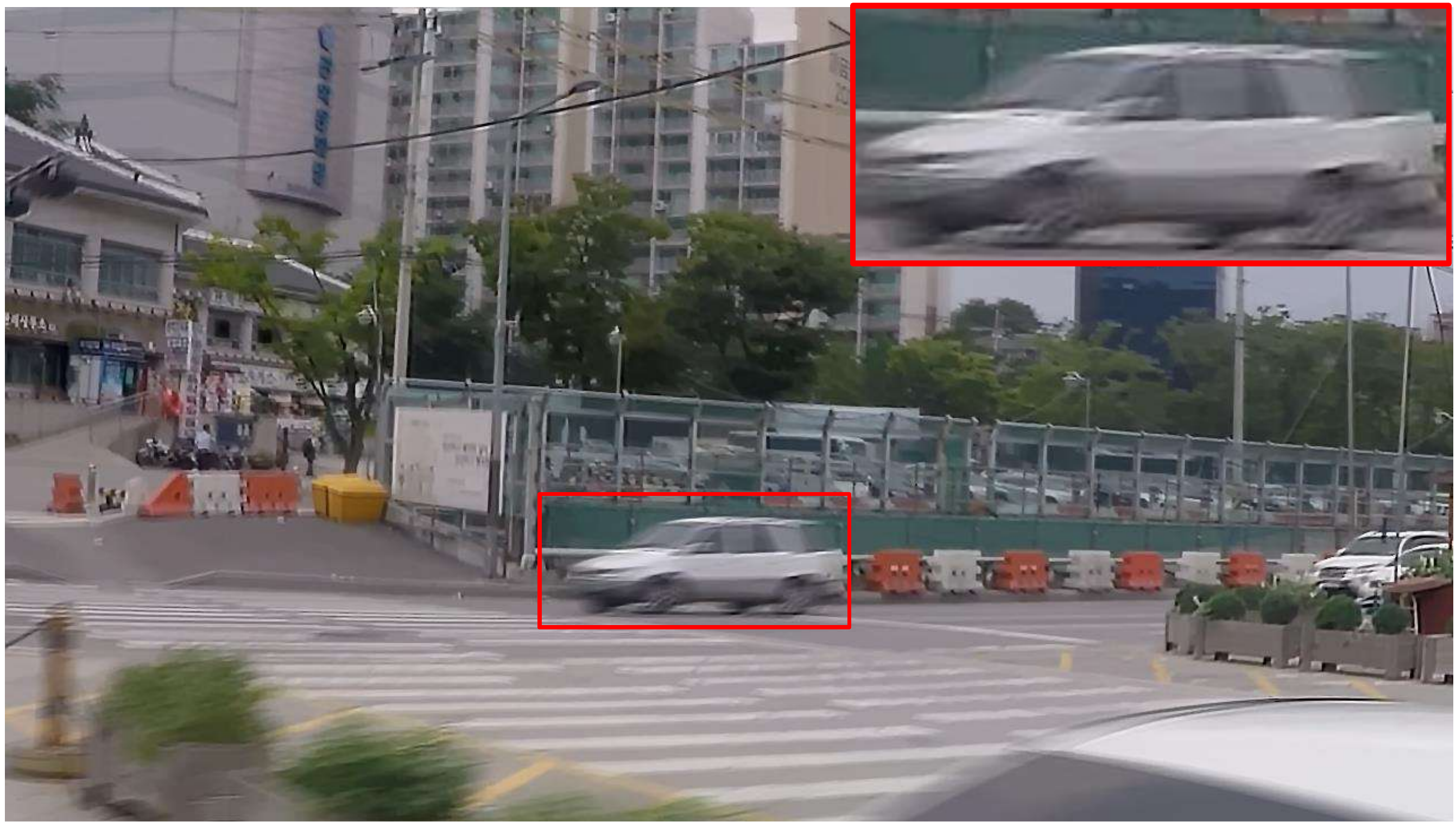}\\
\includegraphics[width=0.235\textwidth]{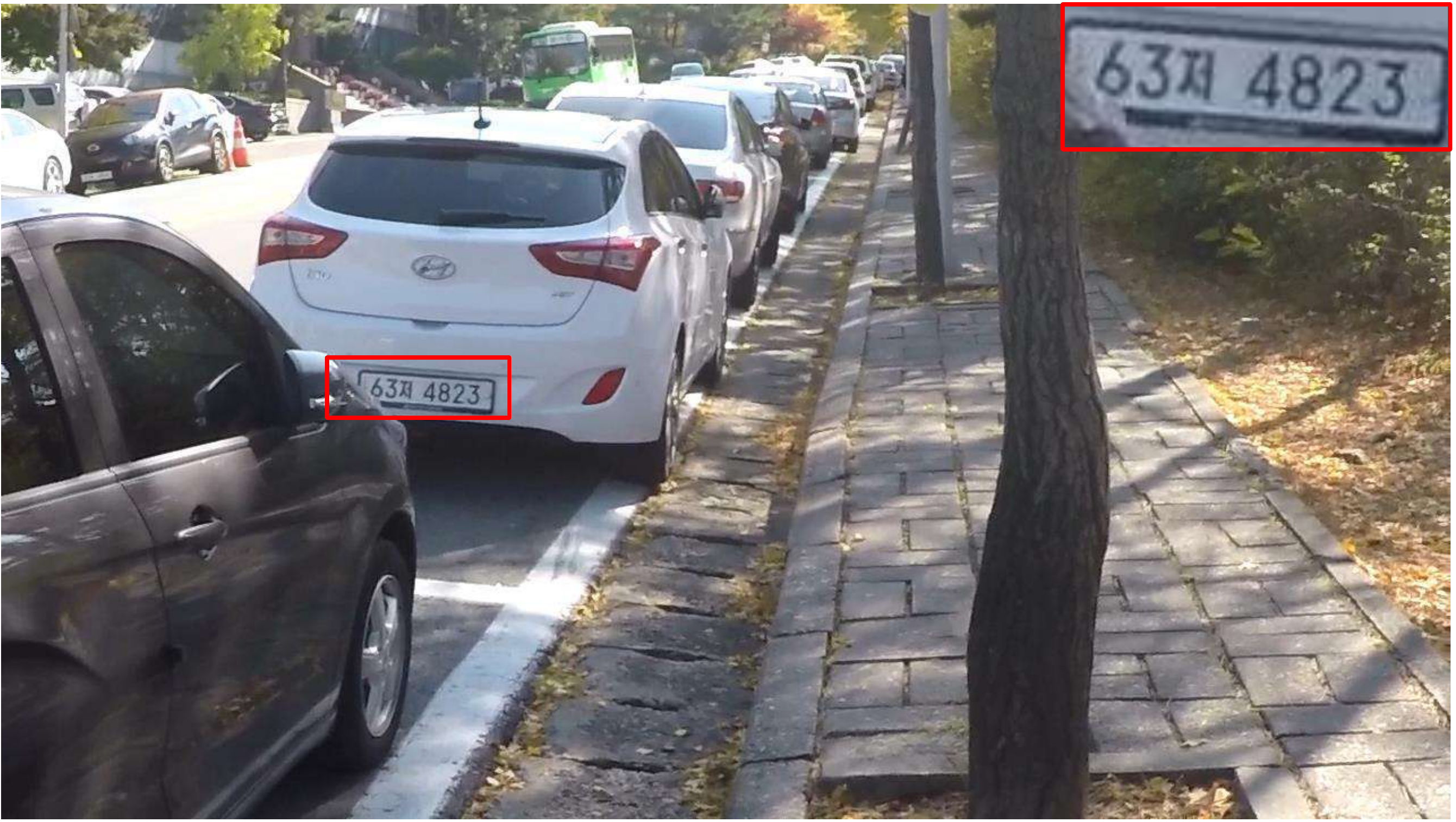}
&\includegraphics[width=0.235\textwidth]{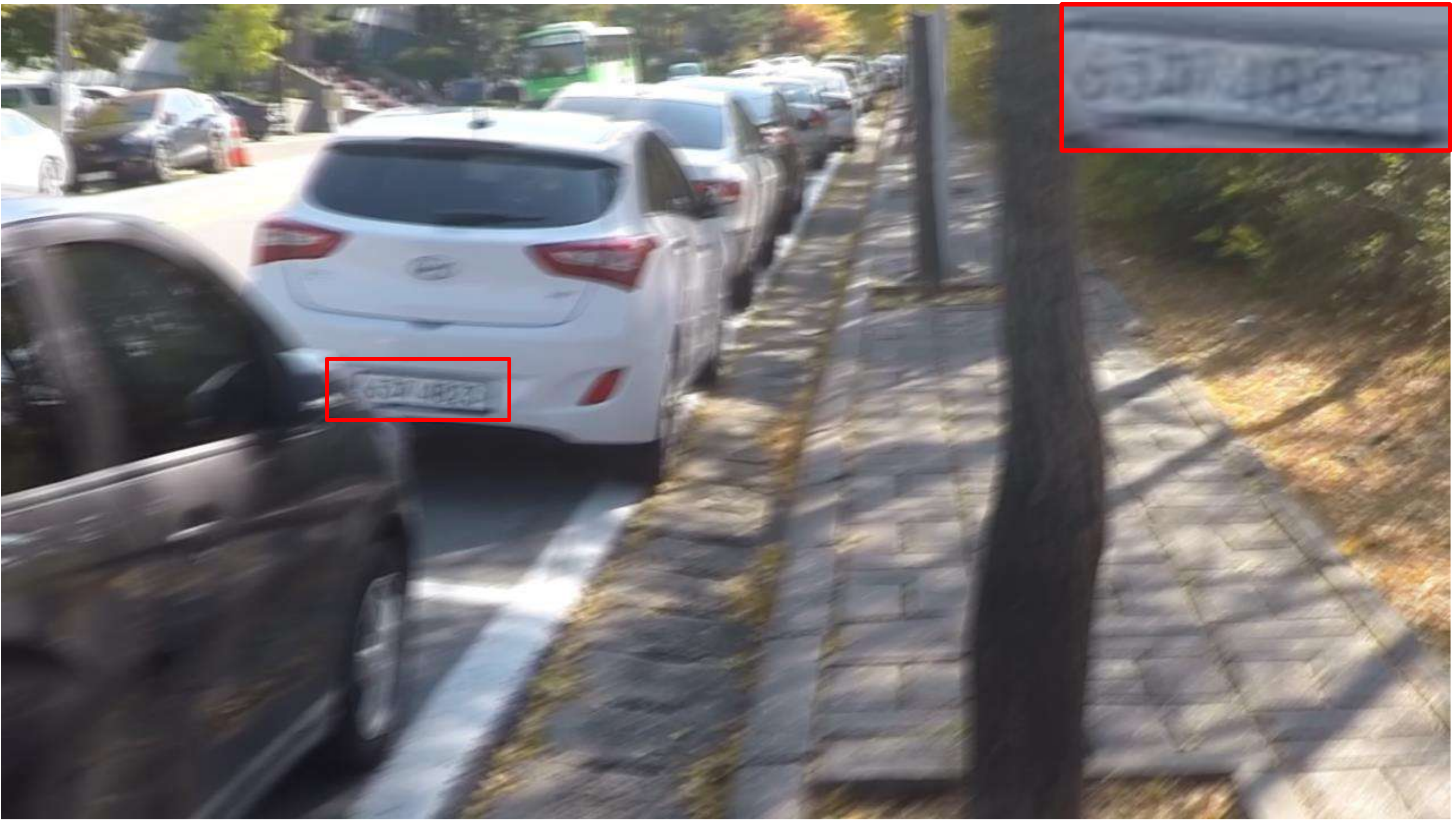}
&\includegraphics[width=0.235\textwidth]{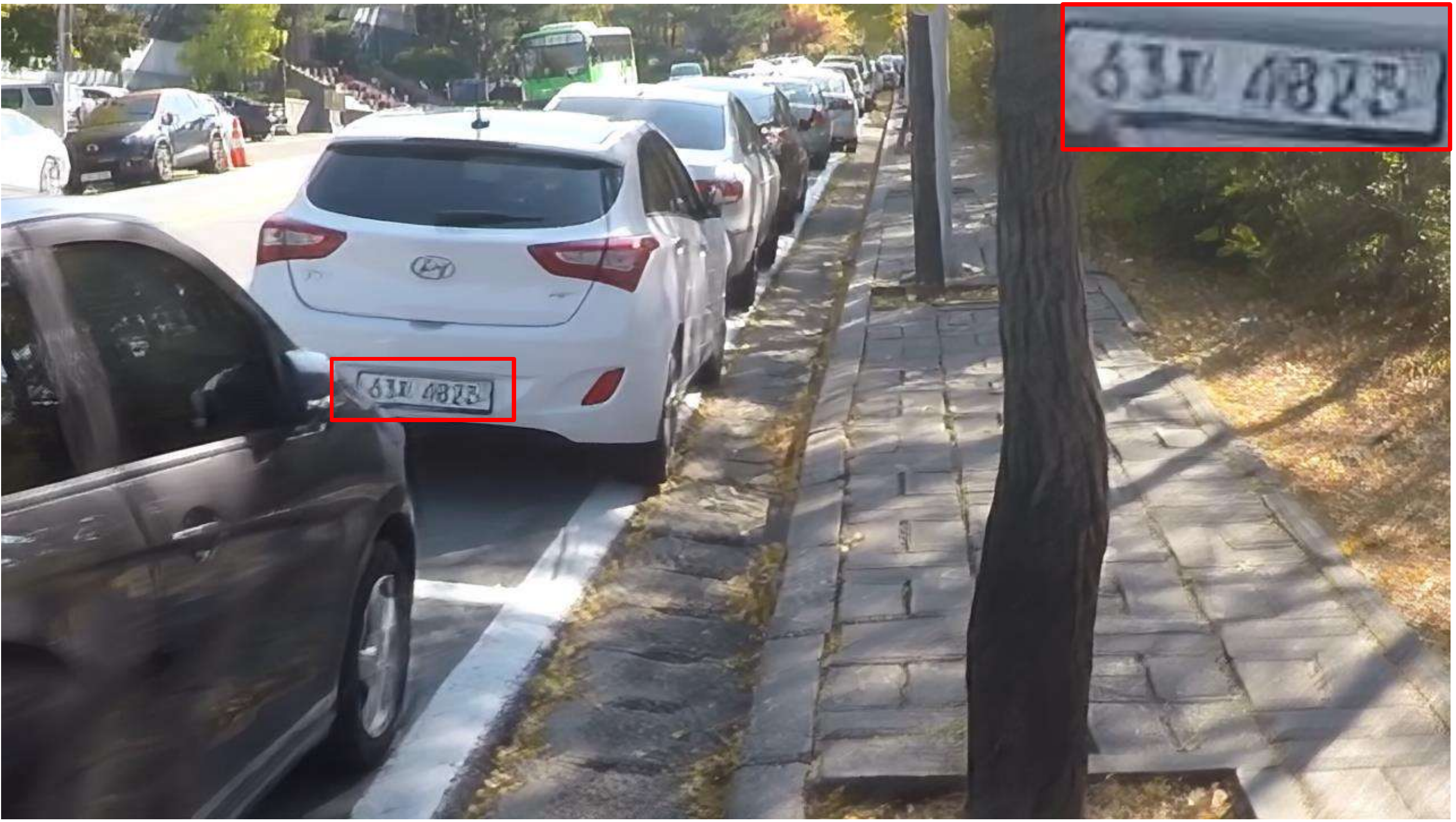}
&\includegraphics[width=0.235\textwidth]{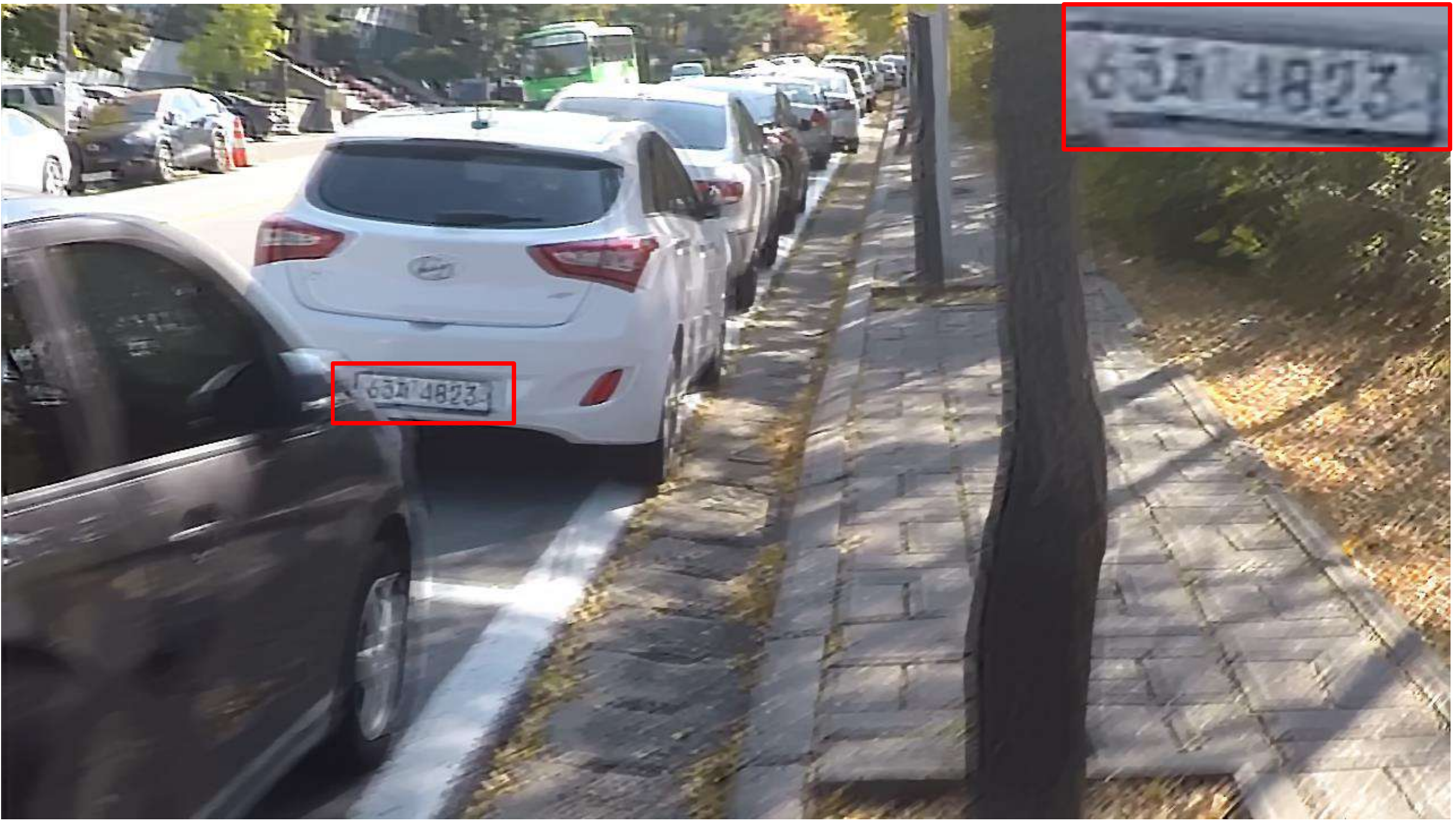}\\
(a) Sharp images  
&(b) Blurry images
&(c) Jin \etal \cite{Jin_2018_CVPR} 
&(d) Pan \etal \cite{pan2017deblurring} \\
\includegraphics[width=0.235\textwidth]{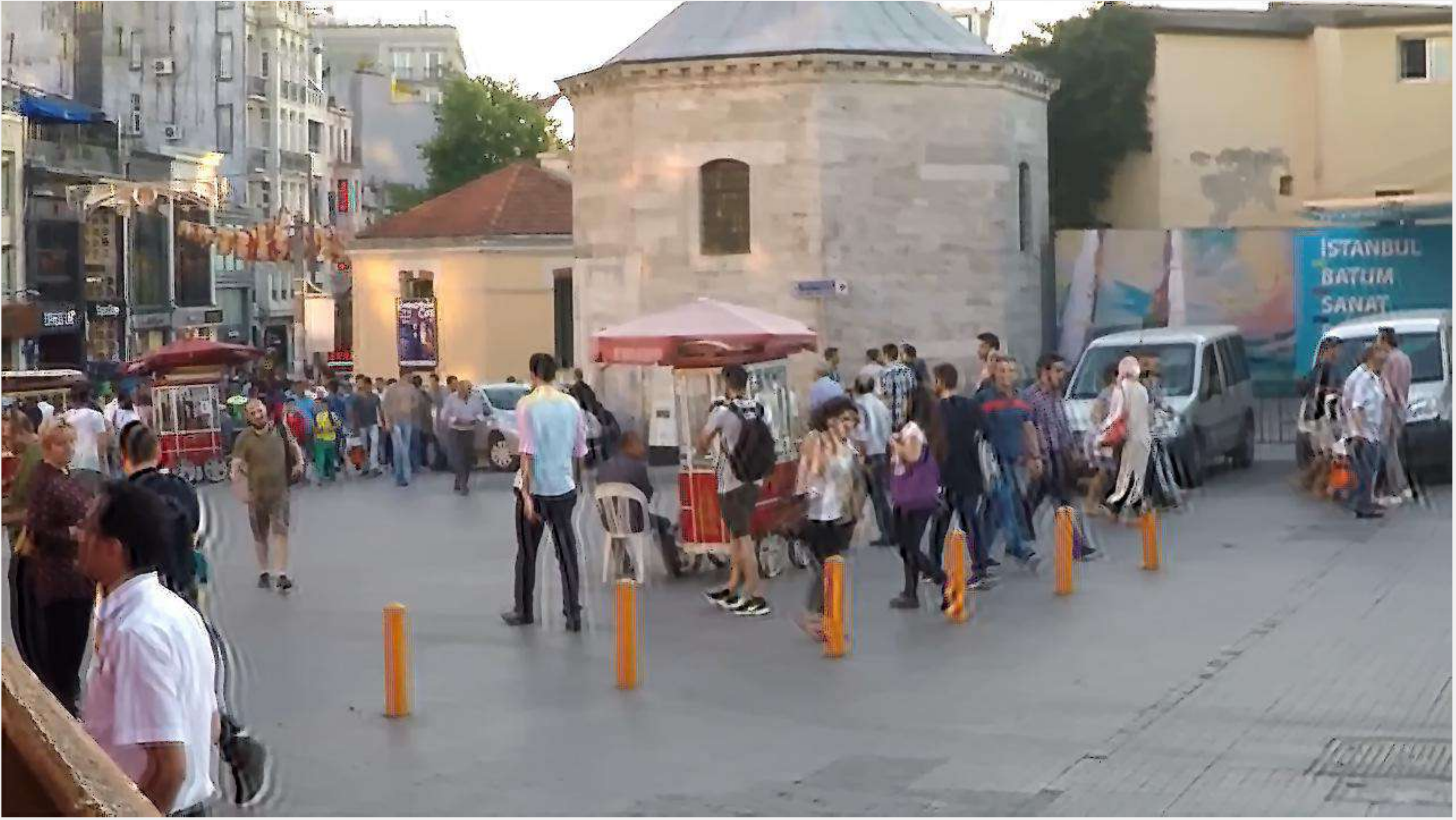}
&\includegraphics[width=0.235\textwidth]{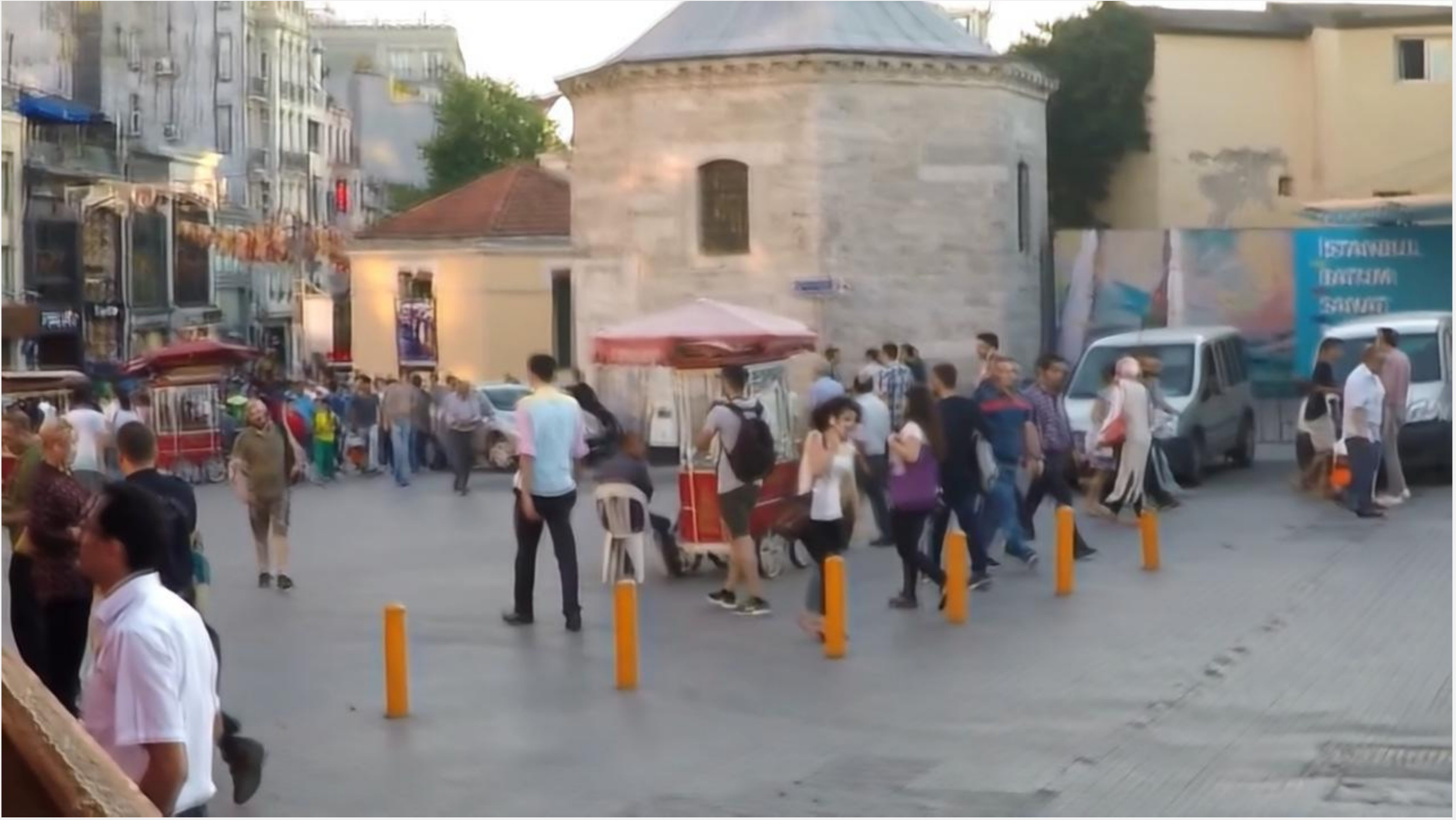}
&\includegraphics[width=0.235\textwidth]{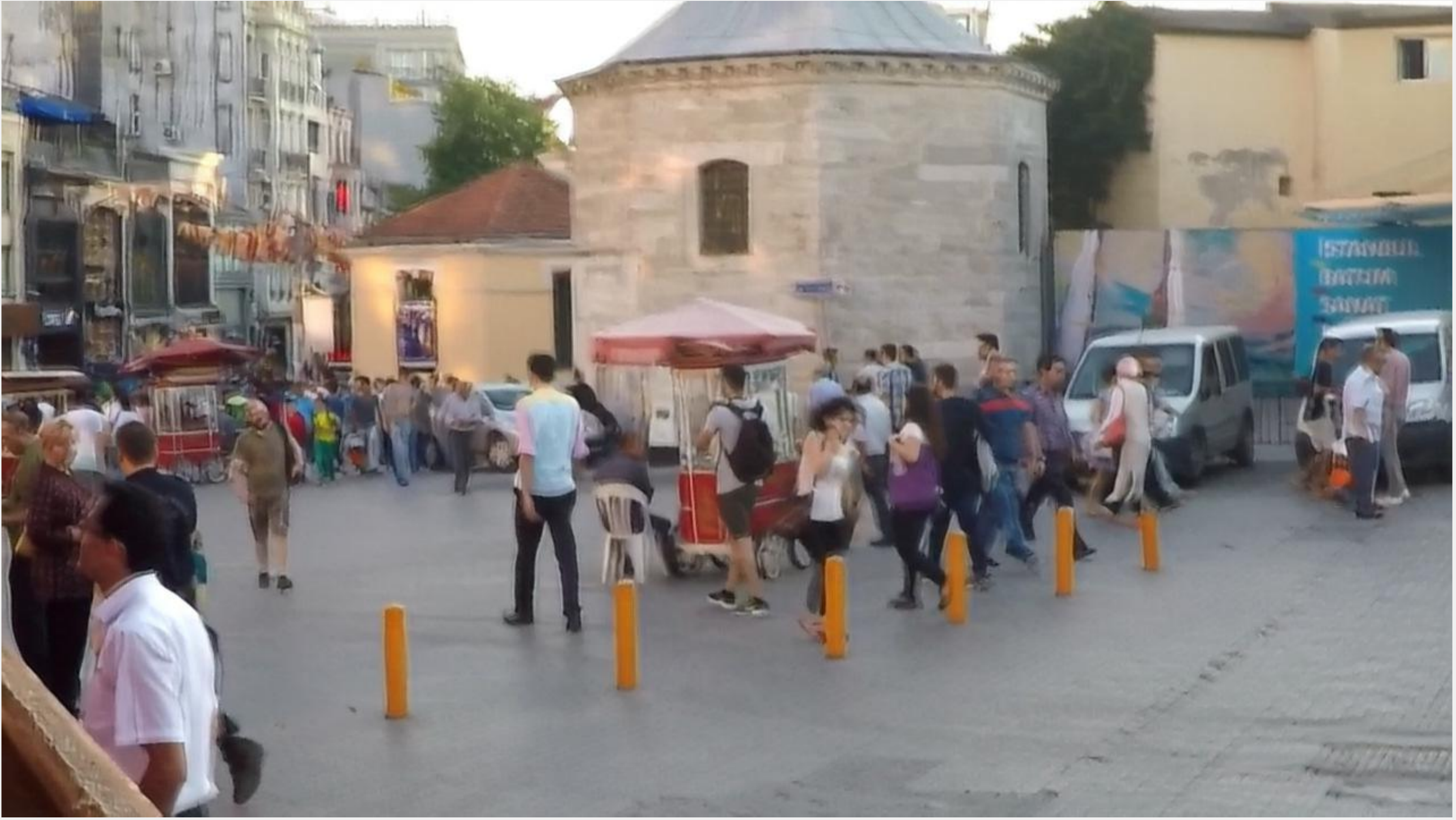}
&\includegraphics[width=0.235\textwidth]{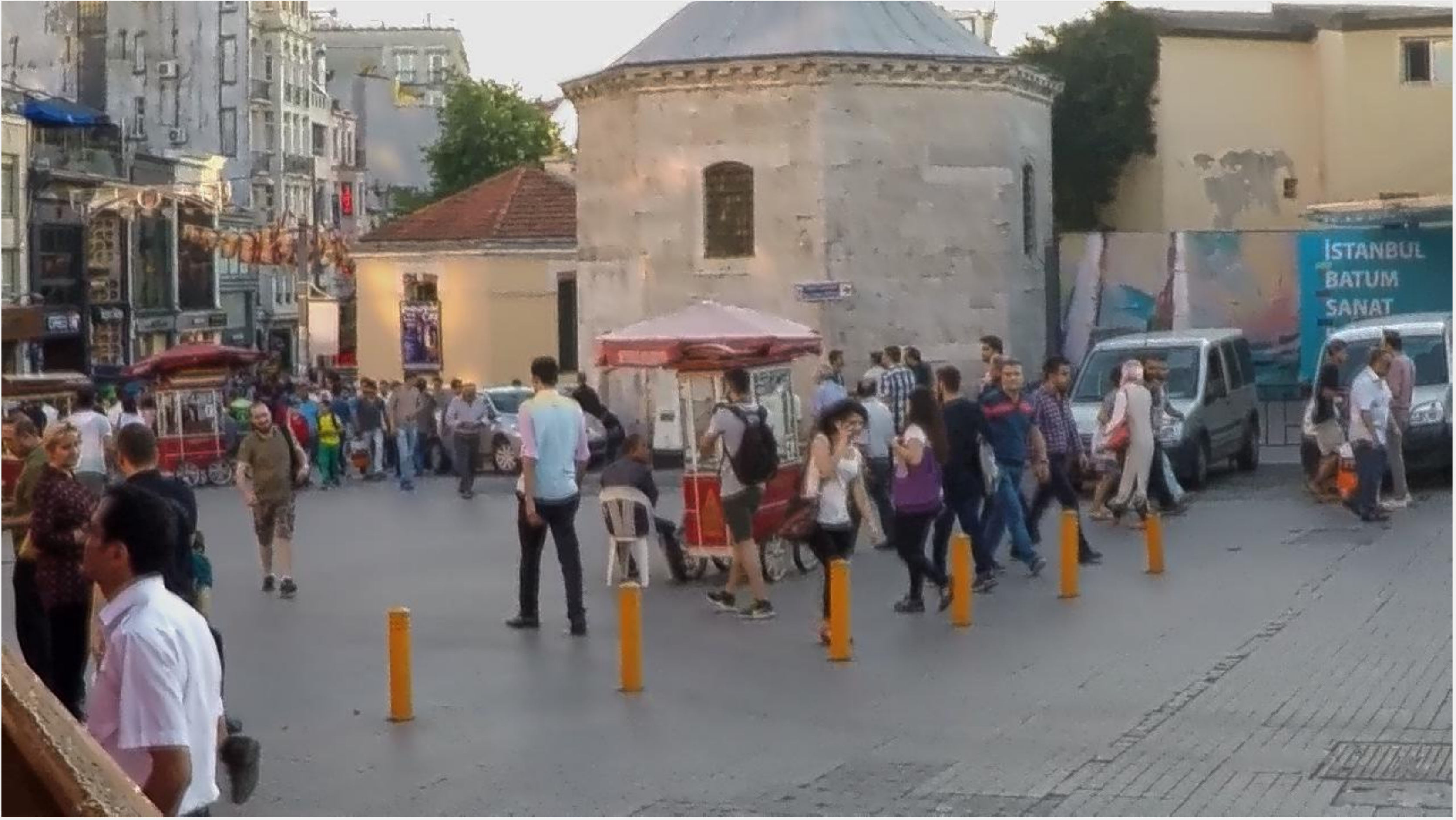}\\
\includegraphics[width=0.235\textwidth]{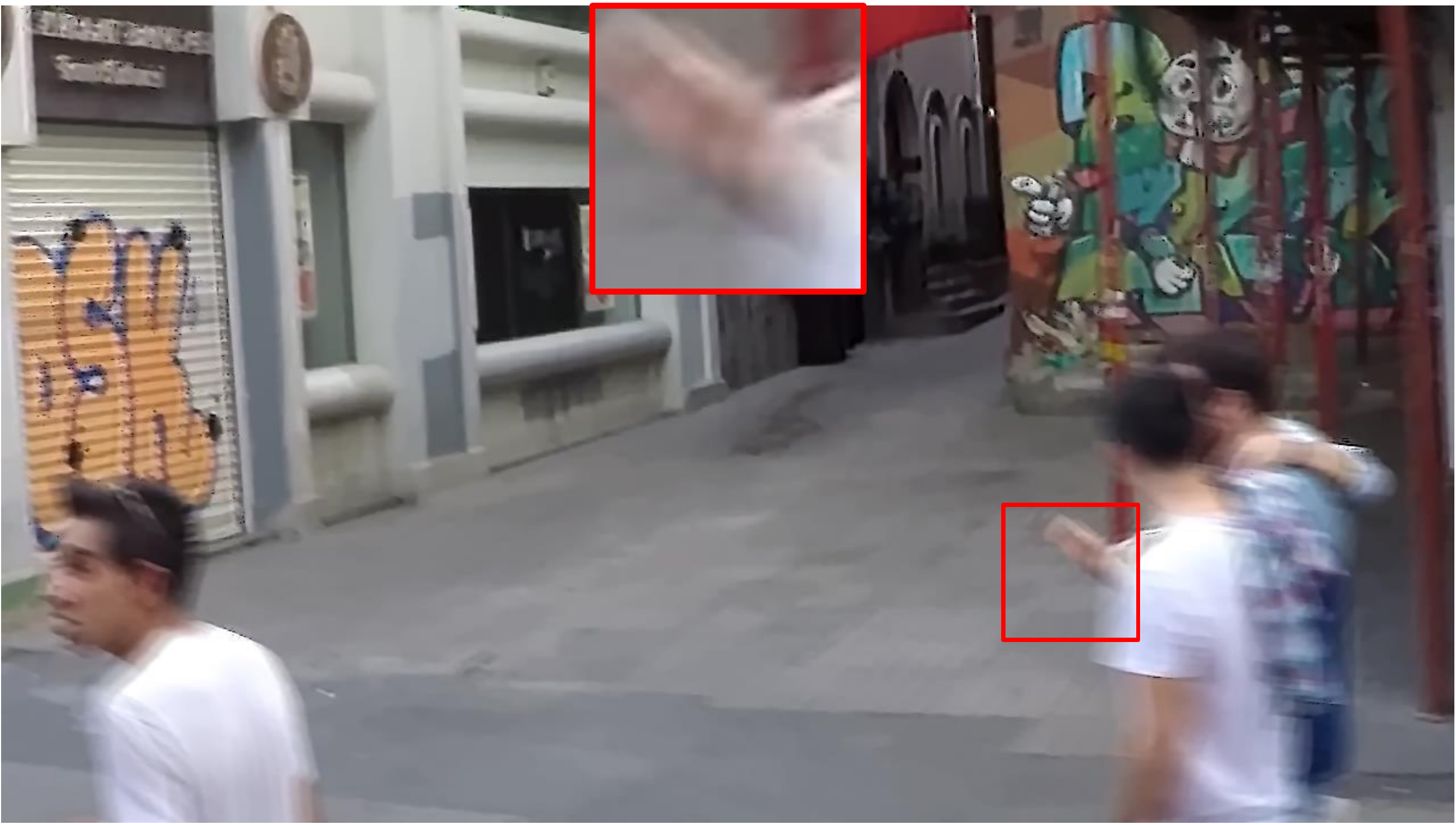}
&\includegraphics[width=0.235\textwidth]{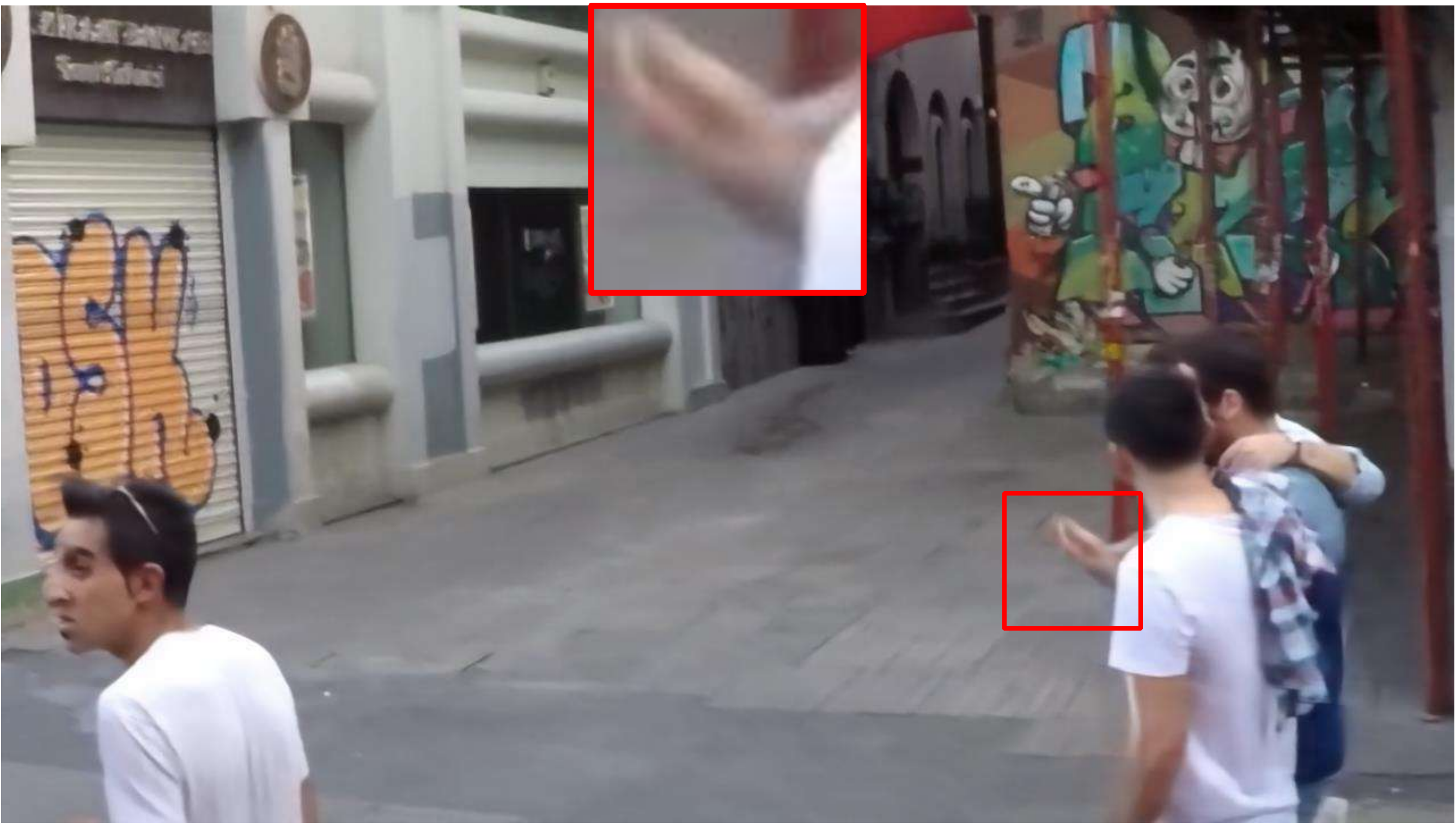}
&\includegraphics[width=0.235\textwidth]{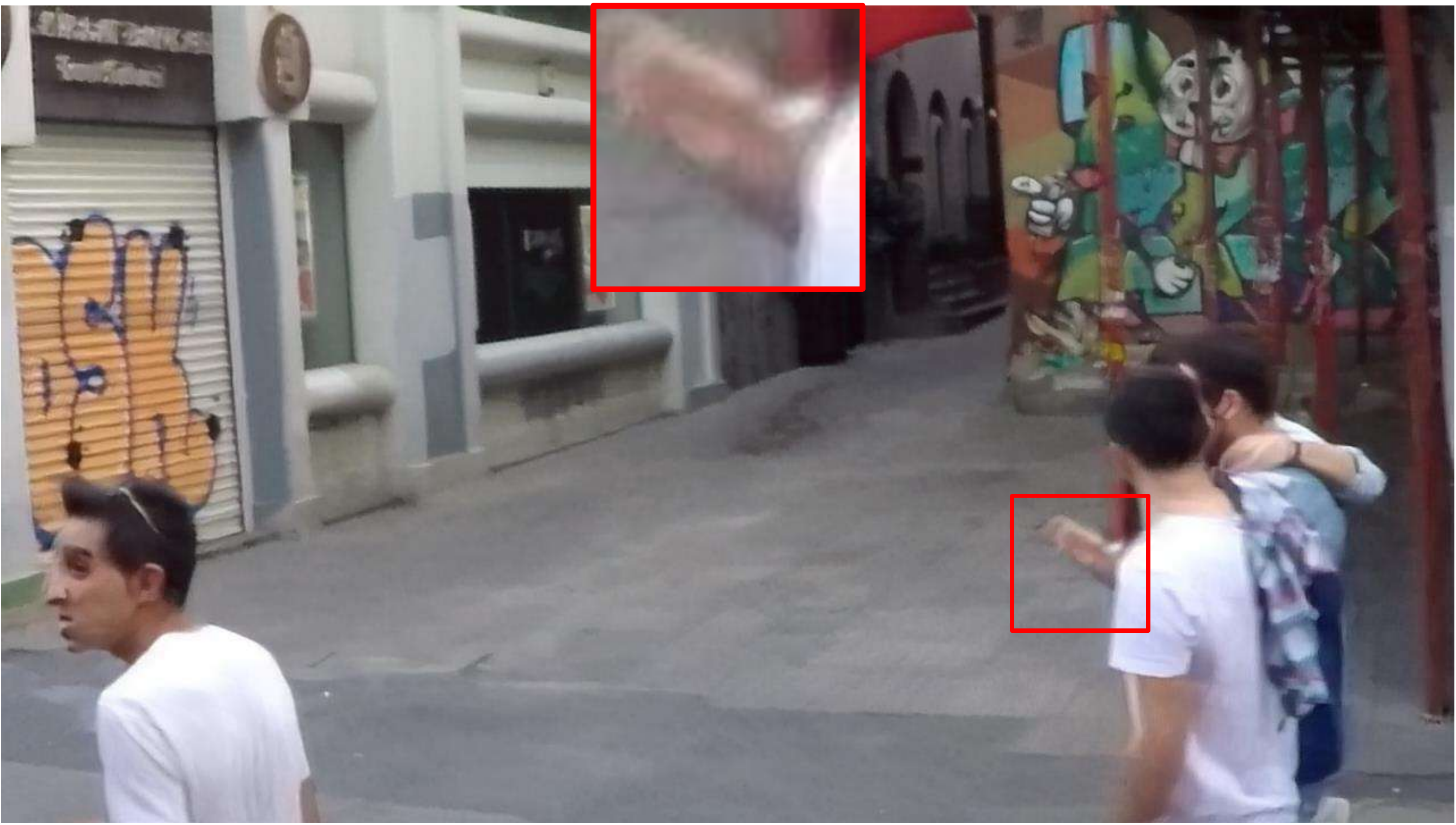}
&\includegraphics[width=0.235\textwidth]{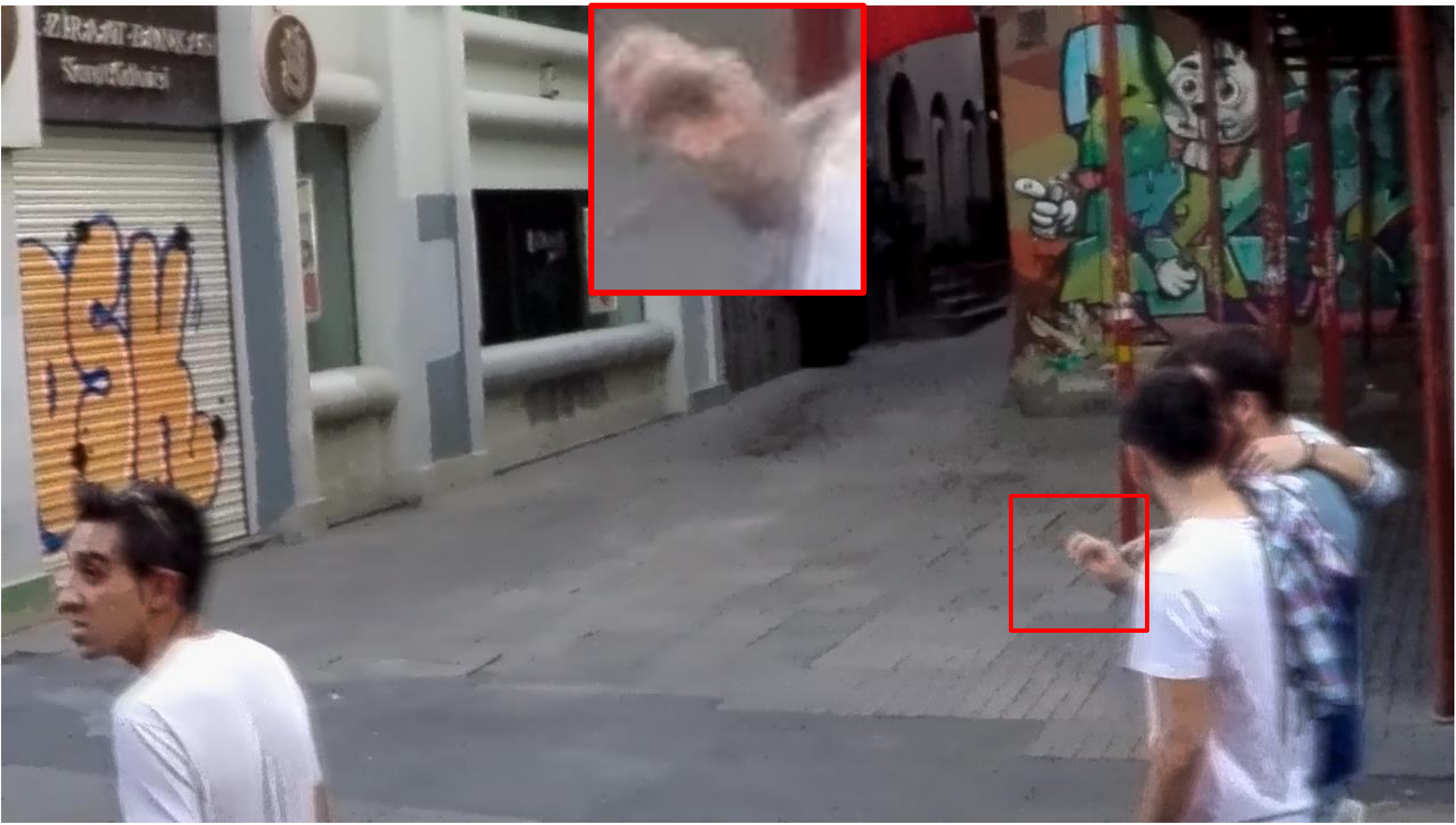}\\
\includegraphics[width=0.235\textwidth]{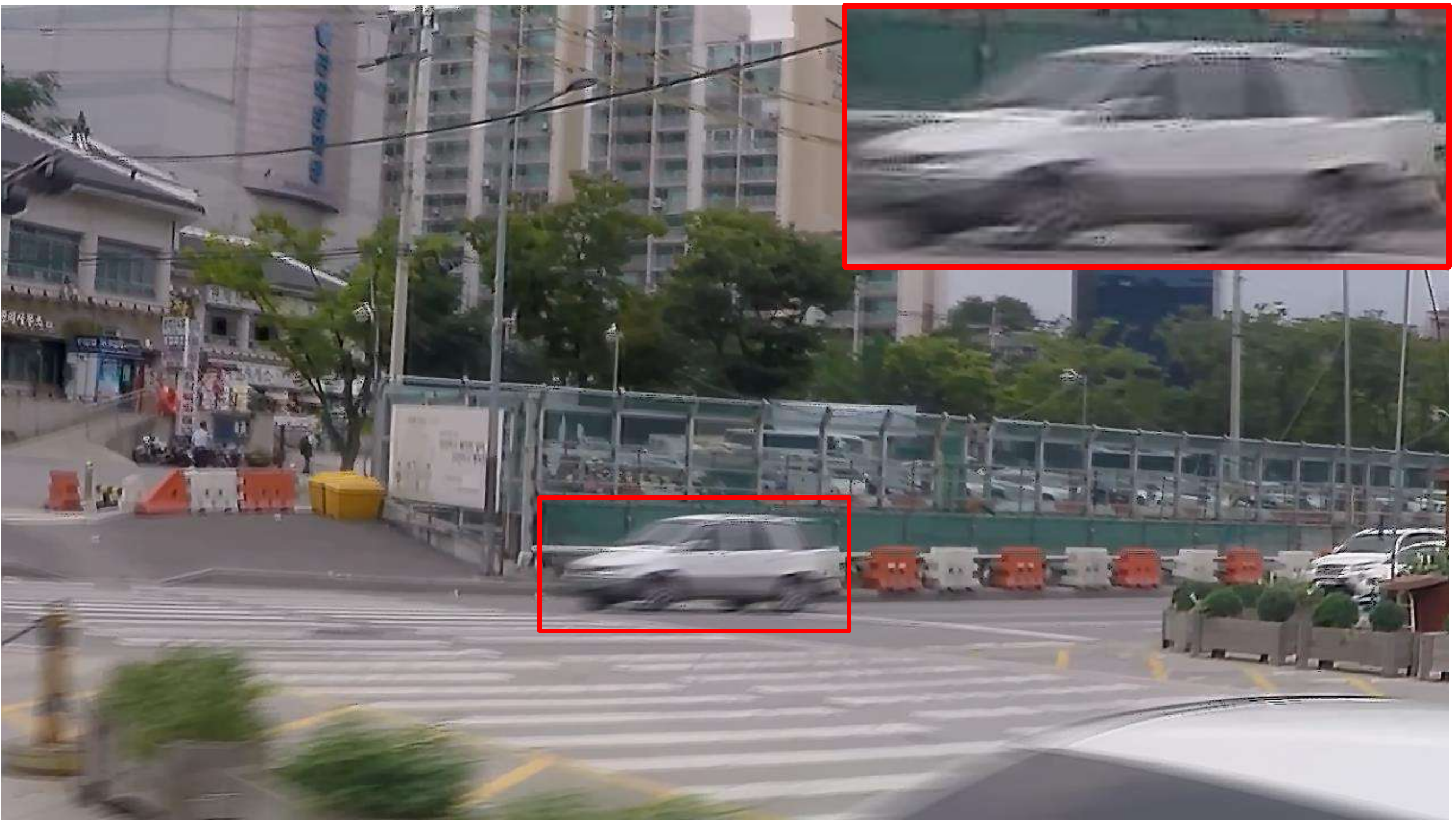}
&\includegraphics[width=0.235\textwidth]{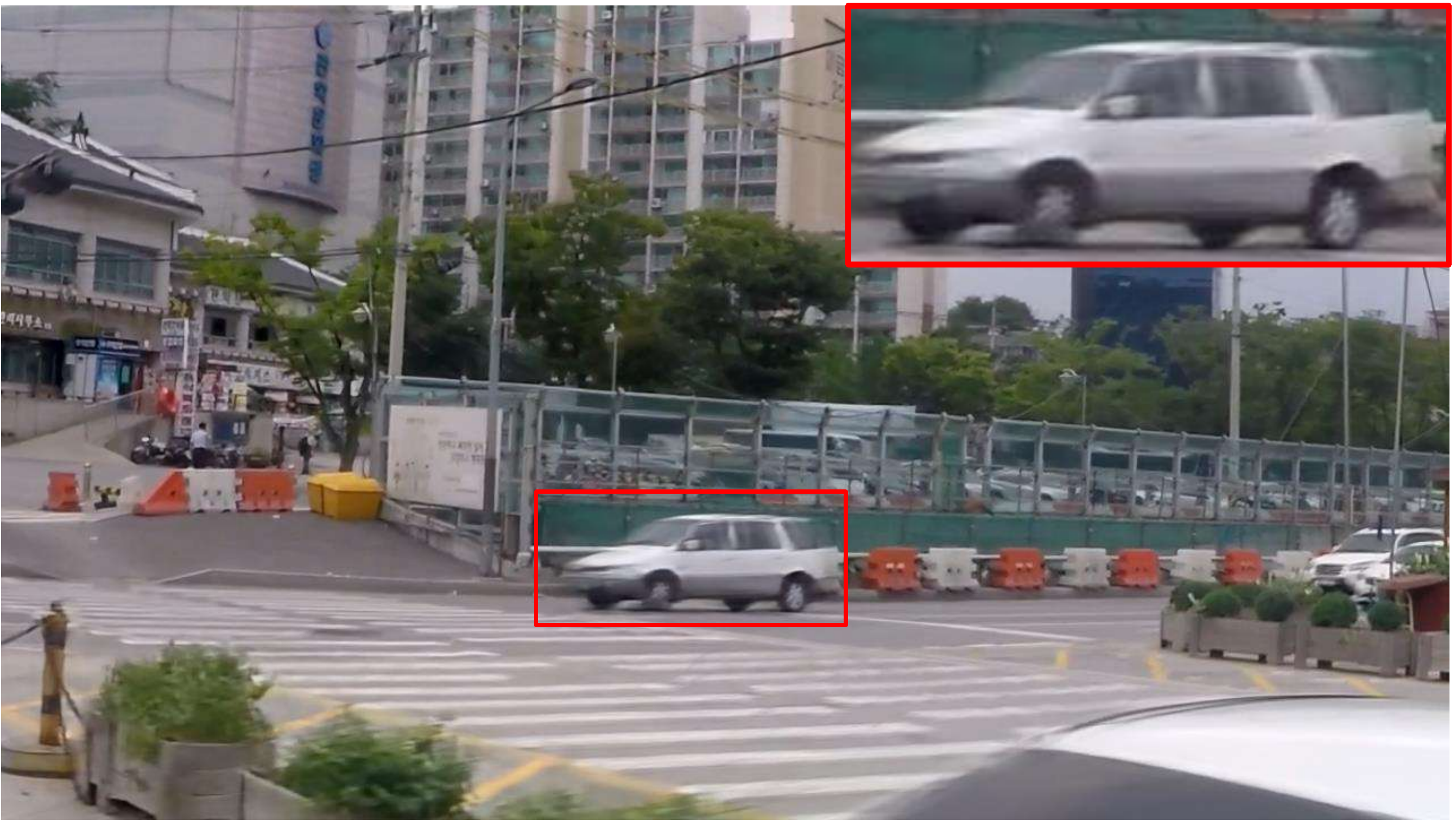}
&\includegraphics[width=0.235\textwidth]{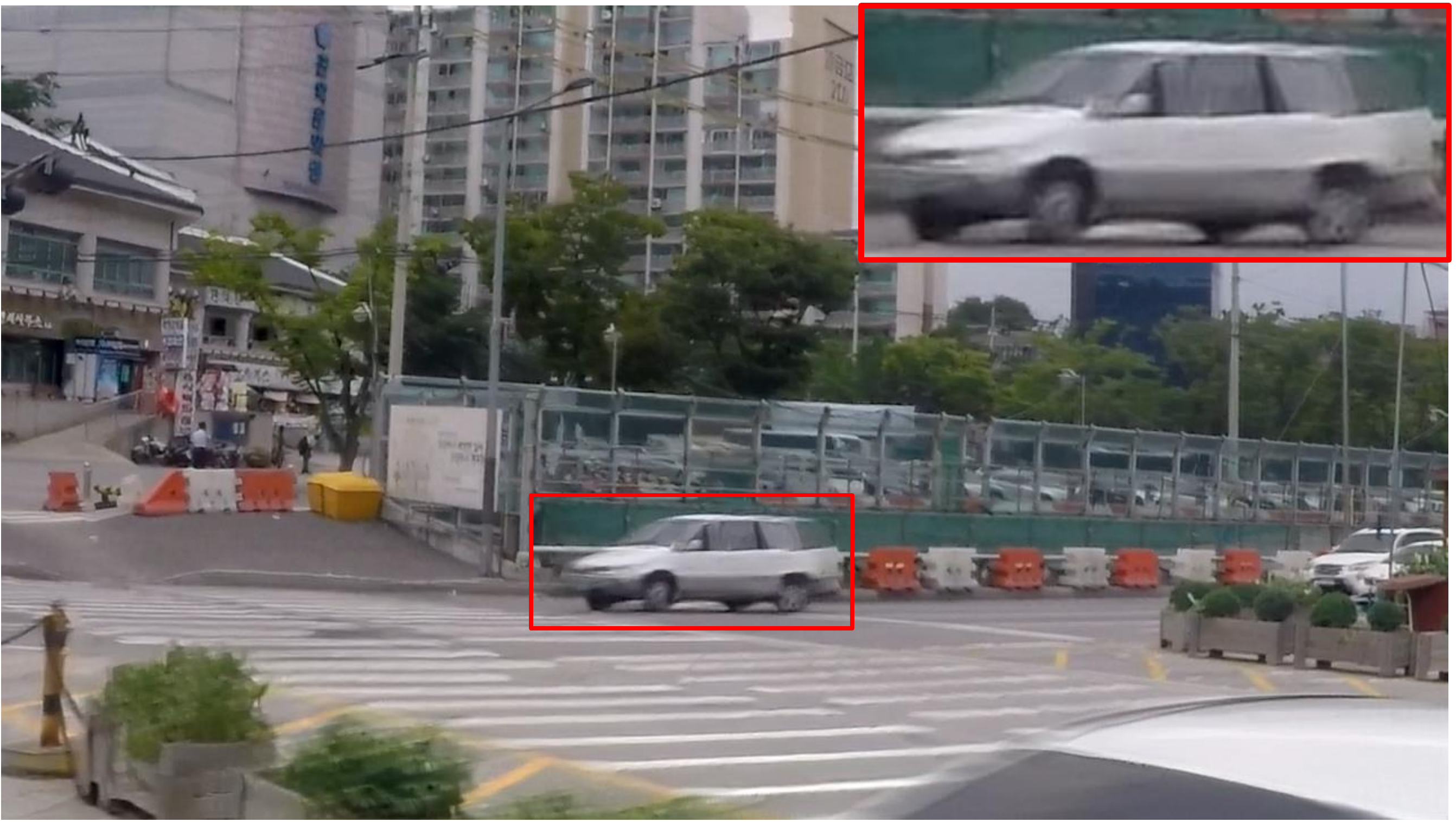}
&\includegraphics[width=0.235\textwidth]{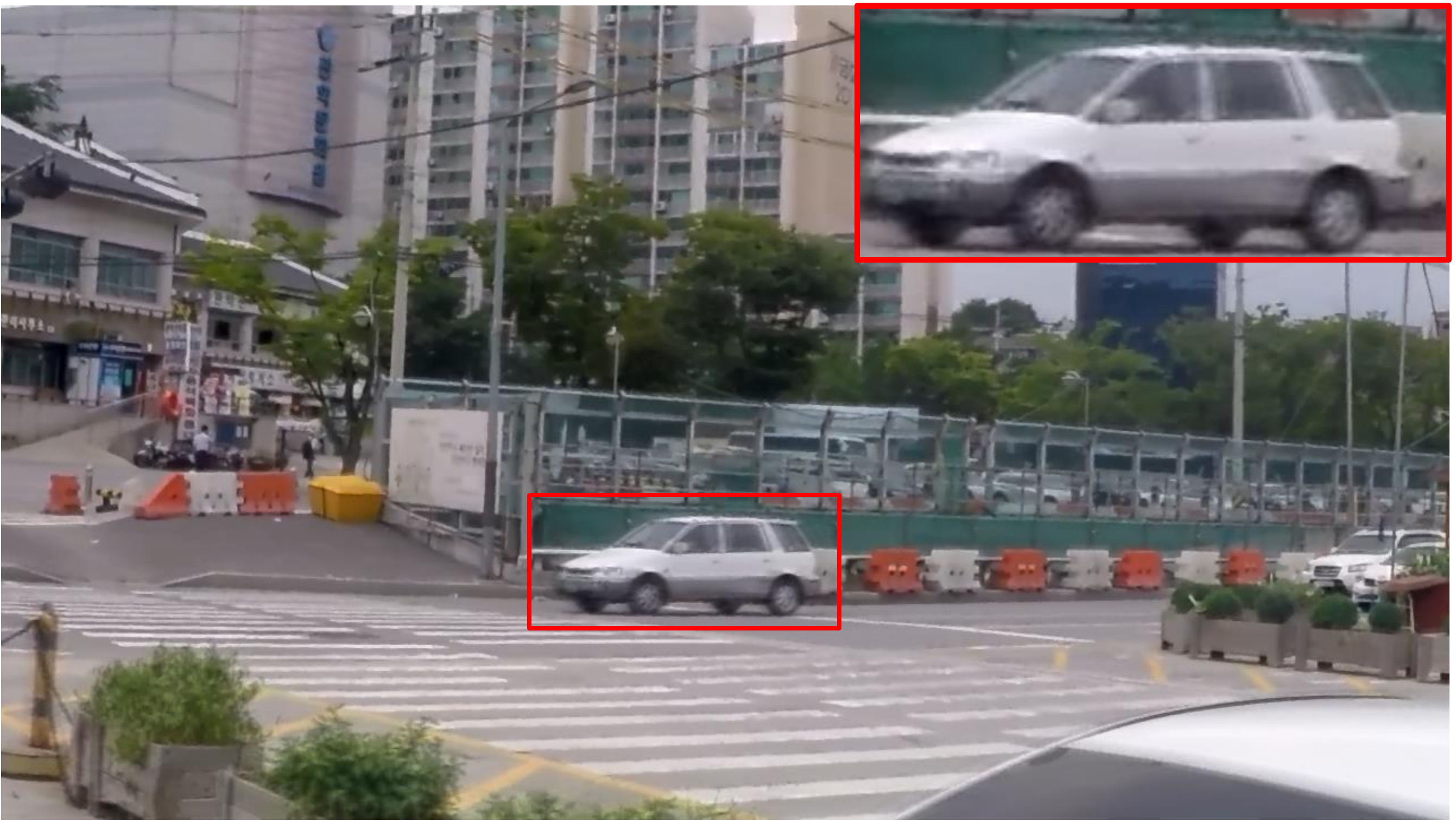}\\
\includegraphics[width=0.235\textwidth]{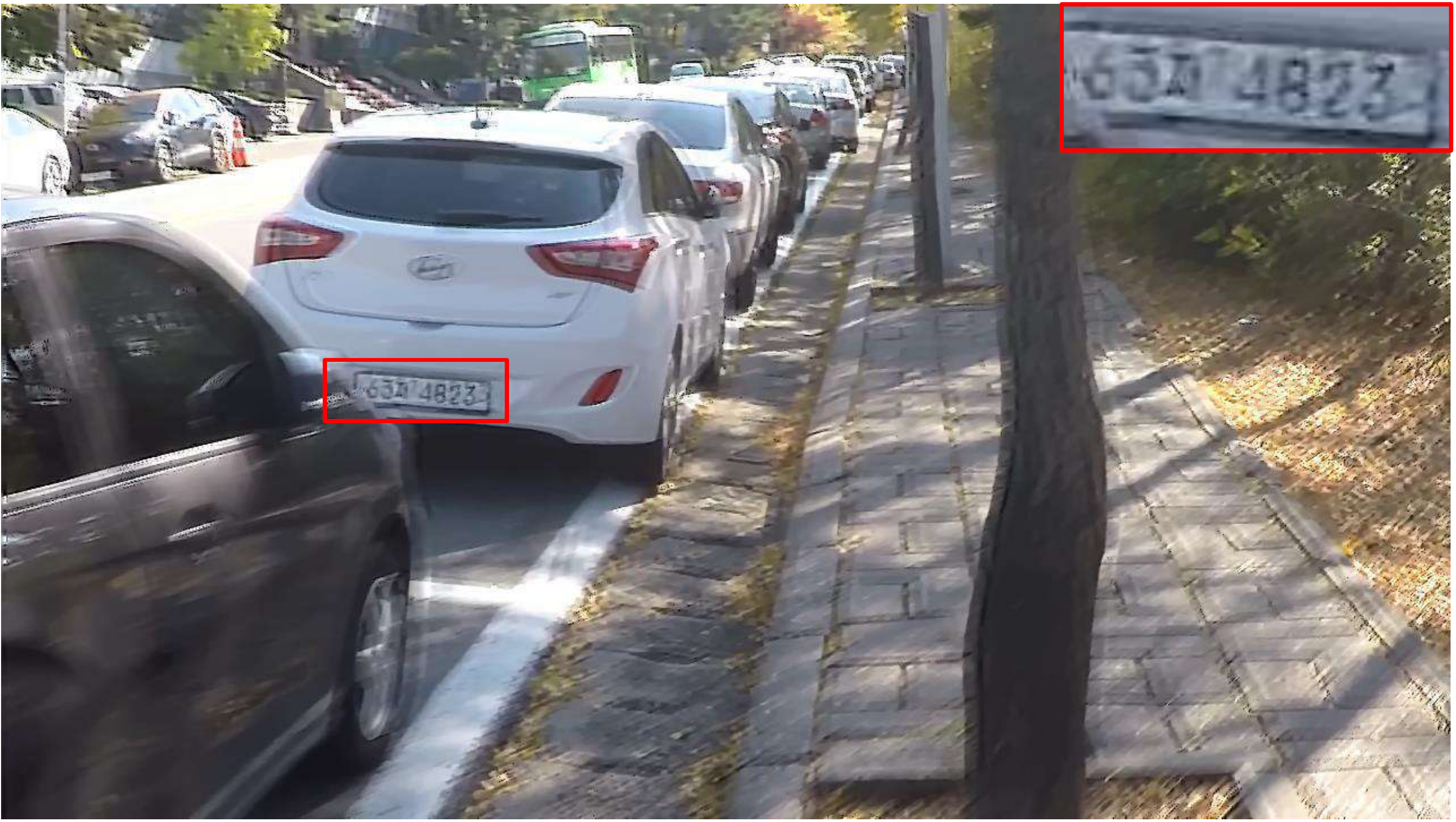}
&\includegraphics[width=0.235\textwidth]{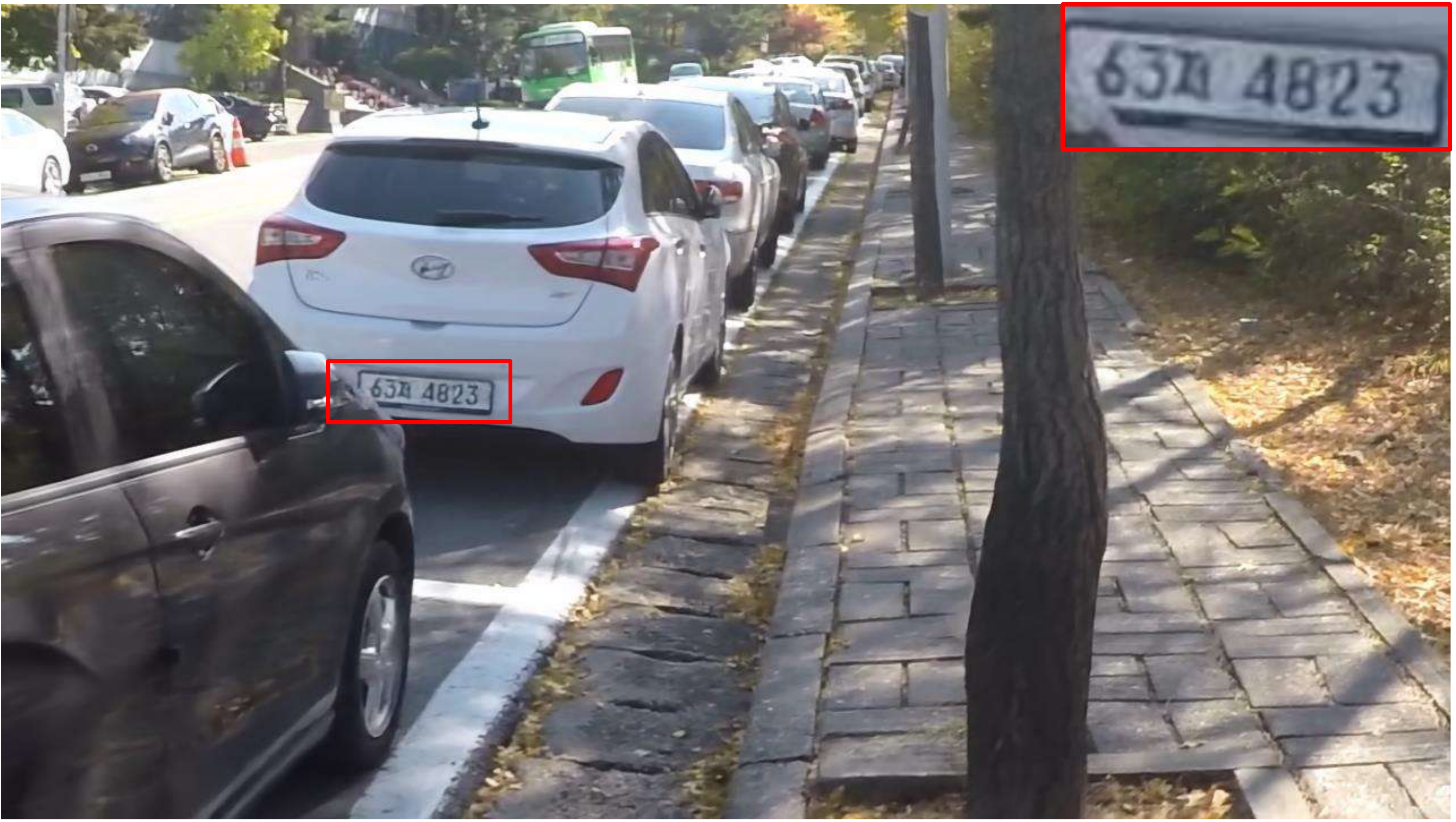}
&\includegraphics[width=0.235\textwidth]{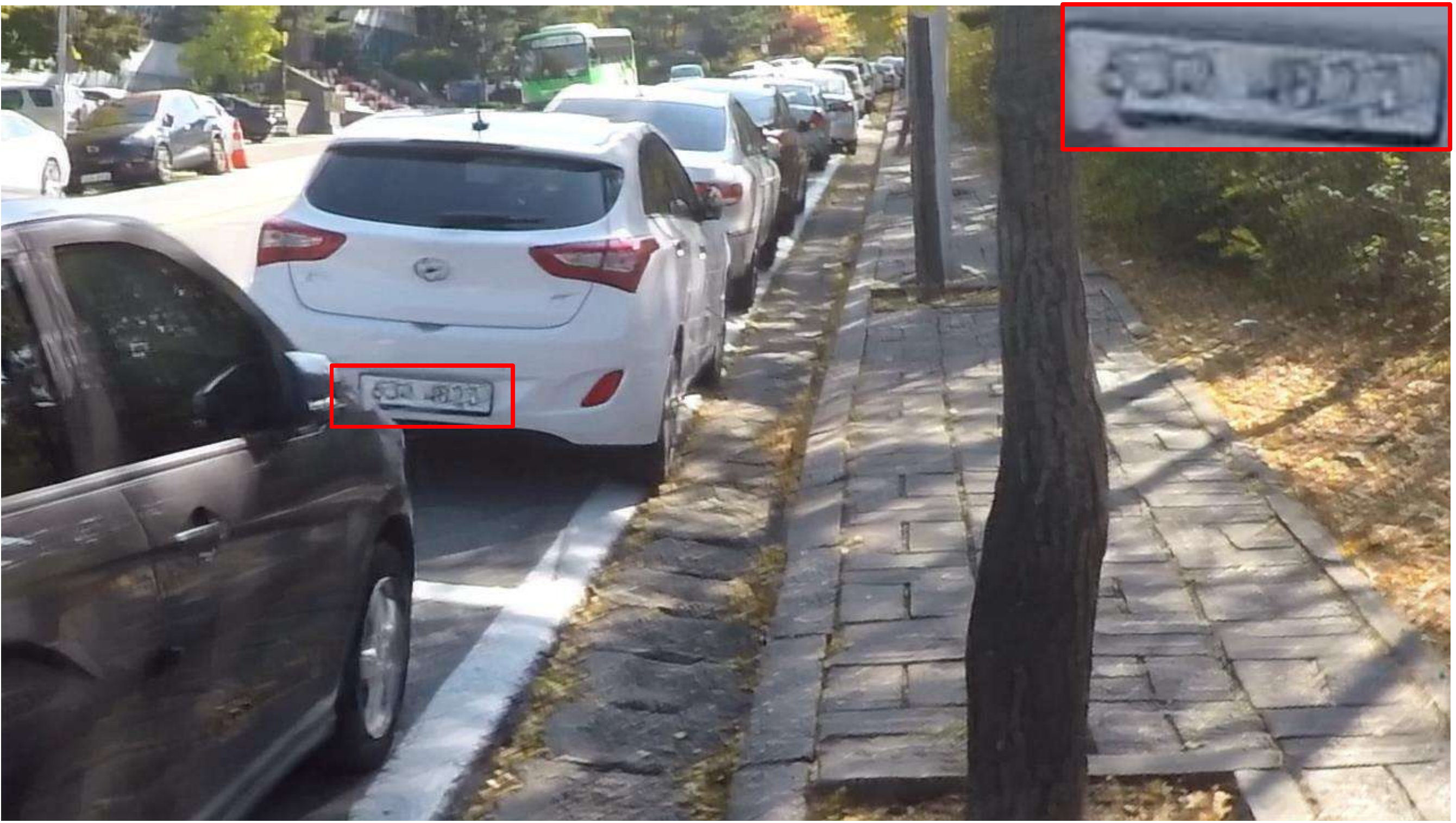}
&\includegraphics[width=0.235\textwidth]{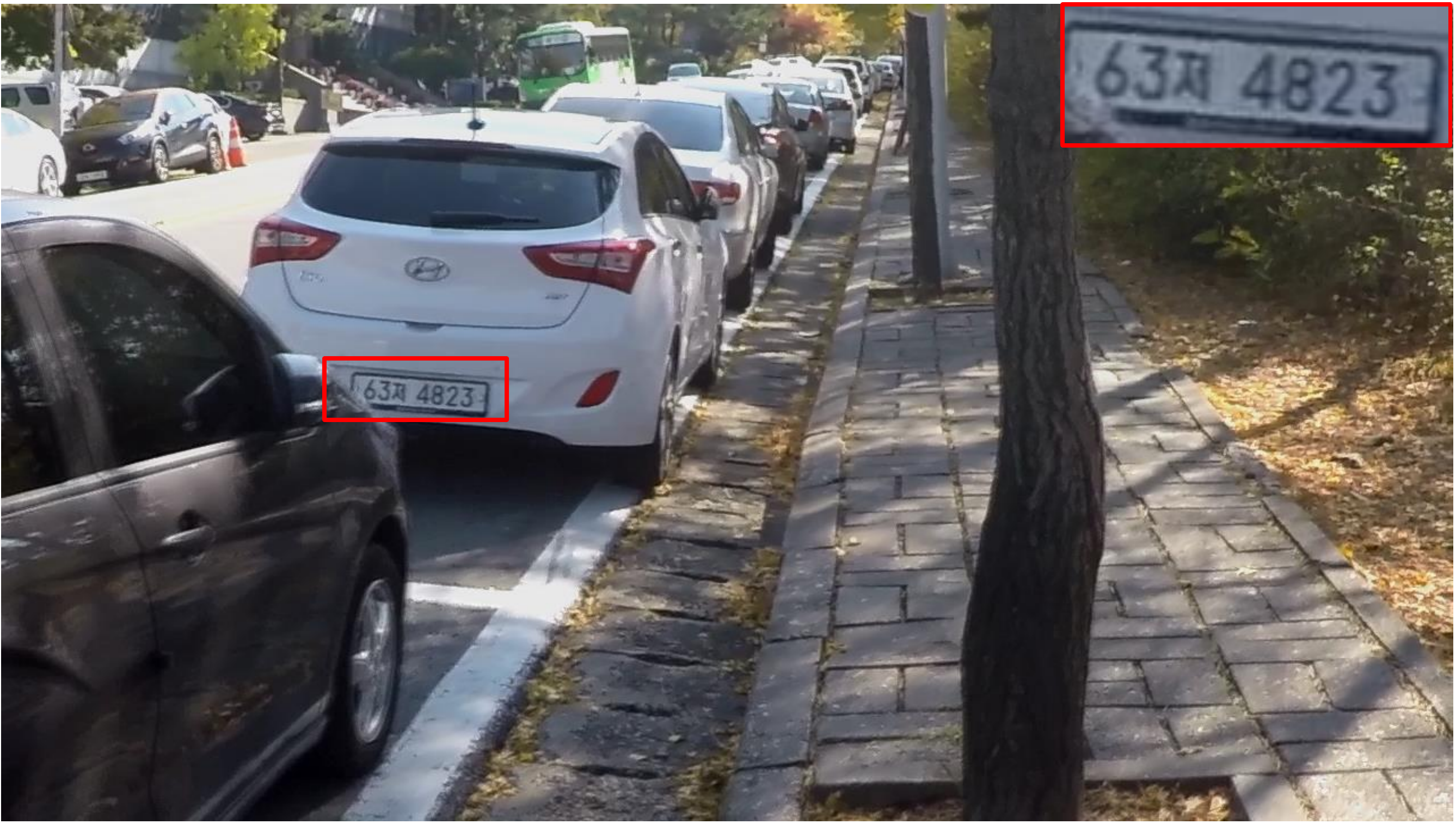}\\
(e) Yan \etal \cite{yan2017image} 
&(f) Tao \etal \cite{Tao_2018_CVPR}
&(g) Nah \etal \cite{Nah_2017_CVPR}
&(h) Ours\\
\end{tabular}
}
\end{center}
\caption{\label{fig:Gopr} Examples of deblurring results on our synthetic event dataset. (a) Sharp images. (b) Generated blurry images. (c) Deblurring results of \cite{Jin_2018_CVPR}. (d) Deblurring results of \cite{pan2017deblurring}. (e) Deblurring results of \cite{yan2017image}. (f) Deblurring results of \cite{Tao_2018_CVPR}. (g) Deblurring results of \cite{Nah_2017_CVPR}. (h) Our deblurring results. (Best view in color
on screen).}
\end{figure*}

In order to qualitatively comparing our method with state-of-the-art deblurring methods \cite{ pan2017deblurring,yan2017image, Jin_2018_CVPR, Tao_2018_CVPR, Nah_2017_CVPR}, we build a synthetic dataset based on the GoPro blurry dataset \cite{Nah_2017_CVPR}. We use the provided videos from \cite{Nah_2017_CVPR} to generate event data.
The comparing results are presented in Fig.~\ref{fig:Gopr}. It clearly shows that our method can generate more frames from a single blurry image and the recovered frames are much sharper.

\subsection{Real Dataset}\label{sec::real_dataset}

A Dynamic and Active-pixel Vision Sensor (DAVIS) \cite{brandli2014240} asynchronously measures the intensity changes at each pixel independently with microsecond temporal resolution.
It uses a shared photo-sensor array to simultaneously output events from its Dynamic Vision Sensor (DVS) and intensity images from its Active Pixel Sensor (APS).

In all our real experiments, we use a DAVIS to build our real {\it Blurry event dataset}. The sequences are taken under different conditions, such as indoor, outdoor, low-lighting, and different motion patterns (camera shake, moving object), \etc. 

Our intensity images are recorded at a frame rate from $5$ fps (frames per second) to 20 fps for different sequences. The resolution of the intensity images is $240\times180$. The maximum frame-rate of our recorded events is $200$ keps (thousands of events per second) since the events depend on the intensity changes of the scene. 

\section{High Frame-Rate Videos}\label{sec::Video}

The input of our method is a single image and its event data during the exposure time. Given a sharp frame $\vL(f)$, setting it as a starting point, we can reconstruct a high frame-rate video by using Eq.~\eqref{eq:L}.
\begin{equation} \label{eq:L}
\vL(t) = \vL(f)\exp \Big (c \int_{f}^{t} e(s) ds\Big).
\end{equation}

\begin{figure*}[ht]
\begin{center}
\resizebox{\textwidth}{!}{
\begin{tabular}{cccc}
\includegraphics[width=0.230\textwidth]{./cnewbaseline/rotate53blur.pdf}
&\includegraphics[width=0.230\textwidth]{./cnewbaseline/deblurjin.pdf}
&\includegraphics[width=0.230\textwidth]{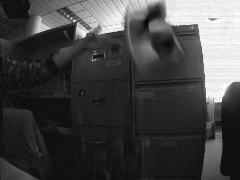}
&\includegraphics[width=0.230\textwidth]{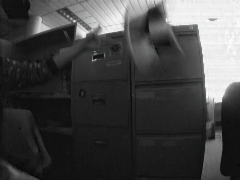}\\
(a) Input blurry image
&(b) Jin \etal \cite{Jin_2018_CVPR}  
&(c) Tao \etal \cite{Tao_2018_CVPR}  
&(d) Nah \etal \cite{Nah_2017_CVPR}  \\
\includegraphics[width=0.230\textwidth]{./cnewbaseline/video/board_2.pdf}
&\includegraphics[width=0.230\textwidth]{./cnewbaseline/video/board_50.pdf}
&\includegraphics[width=0.230\textwidth]{./cnewbaseline/video/board_90.pdf}
&\includegraphics[width=0.230\textwidth]{./cnewbaseline/video/board_130.pdf}\\
\multicolumn{4}{c}{(f) Samples of our reconstructed video}
\end{tabular}
}
\end{center}
 \caption{\em \label{fig:baseline1} Deblurring and reconstruction results on our real {\it blurry event dataset}. 
(a) Input blurry image. 
(b) Deblurring result of \cite{Jin_2018_CVPR}. 
(c) Deblurring result of \cite{Tao_2018_CVPR}. 
(d) Deblurring result of \cite{Nah_2017_CVPR}. 
(e) Samples of our reconstructed video from $\vL(0)$ to $\vL(150)$.
For baseline 1, we first apply a deblurring method to recover a sharp image, and the recovered image is then fed to a reconstruction method~\cite{Scheerlinck18arxiv}.
For baseline 2, we first use a video reconstruction method ~\cite{Scheerlinck18arxiv} to construct a sequence of intensity images, and then apply a deblurring method to each intensity image. 
On synthetic dataset, we use \cite{Tao_2018_CVPR} as the deblurring method for its state-of-the-art performance. 
On our real blurry event dataset, Jin \cite{Jin_2018_CVPR} achieves better results than \cite{Tao_2018_CVPR} (see Fig. 1 and 8 in our paper). Thus, we use Jin \cite{Jin_2018_CVPR} as the deblurring method.
}
\end{figure*}

When the input image is blurry, a trivial solution would be to first deblur the image with an existing deblurring method and then using Eq.~\eqref{eq:L} to reconstruct the video (see Fig.\ref{fig:baseline1} for details). 
However, in this way, the events between consecutive intensity images are not fully exploited, resulting in inferior performance.
We propose to exploit the inherent connection between event and blur to reconstruct the video,  and give the following model:
\begin{equation}\label{eq:logEDIM2}
\log (\vL(f)) = \log (\vB) - \log \left( \frac{1}{T} \int_{f-T/2}^{f+T/2} \exp(c \,\vE(t)) dt \right). 
\end{equation}


The right-hand side of Eq.~\eqref{eq:logEDIM2} is known, apart from perhaps
the value of the contrast threshold $c$, the first term from
the grey-scale image, the second term from the event sequence, 
it is possible to compute $\log (\vL(f))$, and hence $\vL(f)$ by
exponentiation.  Subsequently, from 
Eq.~\eqref{eq:L} the latent image $\vL(t)$ at any time may be computed.

When tackling a video, we process each frame in the video separately to generate our reconstructed video. 
To avoid accumulating errors, it is more suitable to construct each frame $\vL(t)$ using the closest blurred frame. 

Theoretically, we could generate a video with frame-rate as high as the DVS's eps (events per second). However, as each event carries little information and is subject to noise, several events must be processed together to yield a reasonable image. We generate a reconstructed frame every $50-100$ events, so for our experiments, the frame-rate of the reconstructed video is usually $200$ times greater than the input low frame-rate video. Our model (Eq. \eqref{eq:logEDIM2} ) can get a better result compared to the state-of-the-art deblurring methods. (see Fig.\ref{fig:baseline1} for an example.)
 

We put several results of our reconstructed video in the supplementary material and details are given in Table \ref{table:imlist}. For all our videos, the left part is the input image and the right part is our reconstructed video.

\begin{table*}[]
\begin{center}
\caption{\label{table:imlist} \em Details  of our real \emph{blurry event dataset}}
\begin{tabular}{|c|c|c|c|}
\hline
\textbf{Num.} & \textbf{Video name} & \textbf{Condition}  & \textbf{Motion type}                                                                                  \\ \hline
1             & Tea                 & indoor              & camera shake                                                                                          \\ \hline
2             & text\_SHE           & indoor \& low light & \begin{tabular}[c]{@{}c@{}}moving object \\ (The text shown in the phone is also moving)\end{tabular} \\ \hline
3             & chessboard1         & indoor              & moving object                                                                                         \\ \hline
4             & chessboard2         & indoor              & moving object                                                                                         \\ \hline
5             & jumping             & indoor              & moving object                                                                                         \\ \hline
6             & snowman             & indoor              & camera shake                                                                                          \\ \hline
7             & pillow              & indoor              & moving object                                                                                         \\ \hline
8             & outdoorrun               & outdoor \& dusk     & moving object and camera shake                                                                        \\ \hline
9             & runningman          & indoor              & moving object                                                                                         \\ \hline
10            & lego         & indoor              & moving objects                                                                                        \\ \hline
$11^*$            & nightrun         & outdoor \& low light             & moving object                                                                                        \\ \hline
\multicolumn{4}{l}{$*$:Data from \cite{Scheerlinck18arxiv}.}  \\ 
\end{tabular}
\end{center}
\end{table*}

\section{Code}\label{sec::Code}
\begin{figure*}[t]
 \begin{center}
 \includegraphics[width=0.9\textwidth]{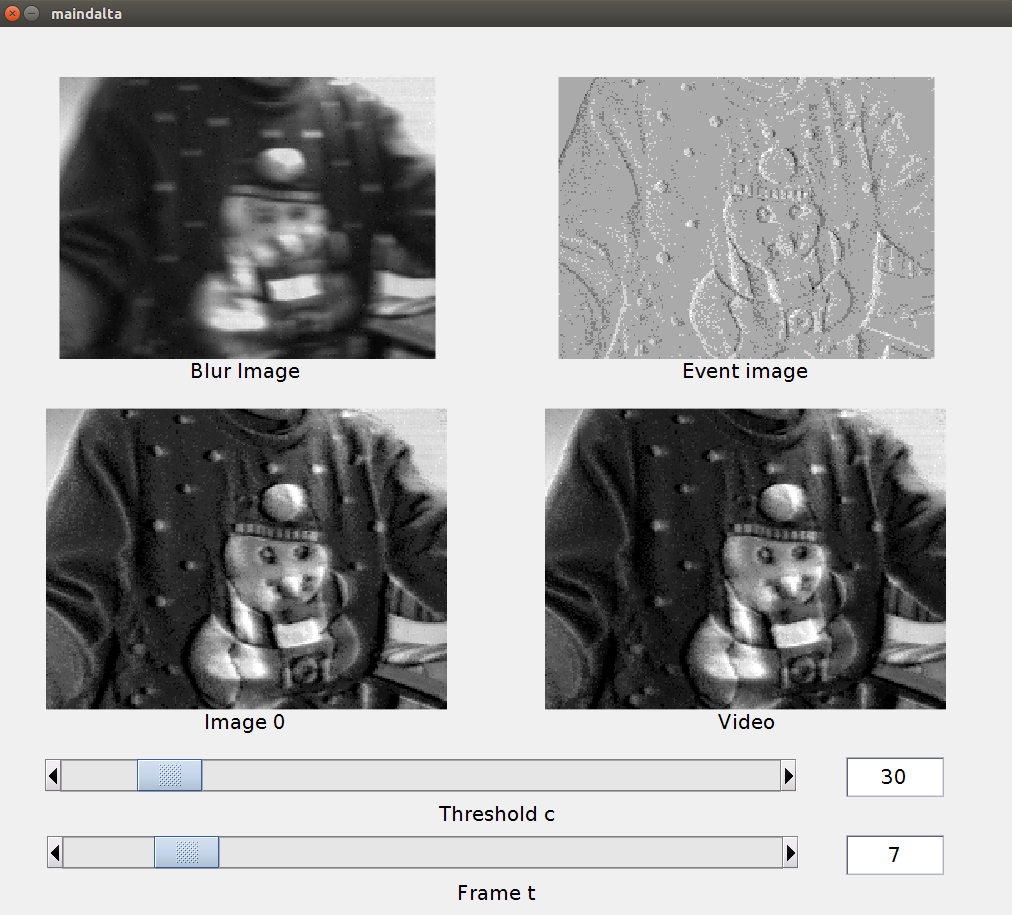}
 \end{center}
 \caption{\em \label{fig:deltaVsblur1} Our program interface.}
 \end{figure*}

All our codes will be released after publication. Here, we only provide a testing code with data `snowman' due to the size limitation of supplementary material ($100$MB). 
Please run `./snowman\_codefortest/maindalta.m', and manually change the two slide bars to visualize the reconstructed image. `Threshold c' is the threshold to trigger an event. `Frame t' is the reconstructed image at timestamp $t$. (see Fig. \ref{fig:deltaVsblur1}).

\end{appendices}

\end{document}